\pdfoutput=1
\documentclass[lettersize,journal]{IEEEtran}

\usepackage{graphicx}
\usepackage{comment}
\usepackage{amsmath,amssymb,amsfonts}
\usepackage{booktabs}
\usepackage{xcolor}

\usepackage{caption}
\usepackage{subcaption}

\usepackage{siunitx}
\usepackage{diagbox}

\usepackage{makecell}
\usepackage{adjustbox}
\usepackage{cite}

\newif\ifarxiv
\arxivtrue

\usepackage{url}
\usepackage{threeparttable} 

\def\MYTITLE{CMax-SLAM: Event-based Rotational-Motion Bundle Adjustment and SLAM System using Contrast Maximization}

\usepackage[pagebackref=false,breaklinks=true,colorlinks,bookmarks=true,bookmarksnumbered=true]{hyperref}
\hypersetup{
  pdftitle={\MYTITLE},
  pdfsubject={Robotics, Computer Vision, State Estimation},
  pdfauthor={Shuang Guo, Guillermo Gallego},
  pdfkeywords={Event Cameras, Motion Estimation, Bundle Adjustment, Asynchronous Sensor, High Dynamic Range, High Temporal Resolution}
}

\usepackage[capitalize]{cleveref}
\crefname{section}{Sec.}{Secs.}
\crefname{table}{Tab.}{Tabs.} 
\crefname{figure}{Fig.}{Figs.}
\Crefname{section}{Section}{Sections}
\Crefname{table}{Table}{Tables}
\Crefname{figure}{Figure}{Figures}

\usepackage{multirow}

\usepackage{url}
\pdfpxdimen=\dimexpr 1 in/192\relax  
\usepackage{calc}

\ifarxiv
\usepackage{orcidlink}
\usepackage[absolute]{textpos}
\fi

\def\Lum{L}
\def\pol{p} 

\def\cE{\mathcal{E}} 
\def\numEvents{N_e} 
\def\numPixels{N_p} 
\def\Warp{\mathbf{W}}

\def\bx{\mathbf{x}}
\def\bparams{\boldsymbol{\vartheta}}
\def\Rot{\mathtt{R}}

\def\velflow{\mathbf{v}}
\def\angvel{\boldsymbol{\omega}} 
\def\bg{\mathbf{g}} 

\def\imgpoint{\bx}

\def\mappointtk{\bp_{m}^{t_k}}

\def\cR{\mathcal{R}}

\def\mB{\mathtt{B}}
\def\bp{\mathbf{p}}
\def\bPhi{\boldsymbol{\Phi}} 
\def\mK{\mathtt{K}}
\def\bX{\mathbf{X}}

\newcommand{\novalue}{{\textendash}}

\definecolor{light-gray}{gray}{0.6}
\newcommand\gframe[1]{{\color{light-gray}\frame{#1}}}

\def\playroom{\emph{playroom}}
\def\bicycle{\emph{bicycle}}
\def\city{\emph{city}}
\def\street{\emph{street}}
\def\town{\emph{town}}
\def\bay{\emph{bay}}

\def\shapes{\emph{shapes}}
\def\poster{\emph{poster}}
\def\boxes{\emph{boxes}}
\def\dynamic{\emph{dynamic}}
\def\indoor{\emph{360$^\circ$ indoor}}

\def\fastmotion{\emph{fast motion}}

\def\river{\emph{river}}
\def\bridge{\emph{bridge}}
\def\crossroad{\emph{crossroad}}

\def\gate{\emph{Brandenburg Gate}}
\def\palace{\emph{Charlottenburg Palace}}
\def\column{\emph{Victory Column}}
\def\mainbuilding{\emph{TUB main building}}
\def\marbuilding{\emph{TUB MAR building}}
\def\platzcenter{\emph{square center}}
\def\platzside{\emph{square side}}

\def\smt{SMT}
\def\rtpt{RTPT}
\def\cmaxgae{CMax-GAE}
\def\cmaxw{CMax-$\angvel$}
\def\cmaxslam{CMax-SLAM}

\begin{document}

\title{\MYTITLE}

\ifarxiv
\definecolor{somegray}{gray}{0.5}
\newcommand{\darkgrayed}[1]{\textcolor{somegray}{#1}}
\begin{textblock}{11}(2.5, 0.4)  
\begin{center}
\darkgrayed{This paper has been accepted for publication at the IEEE Transactions on Robotics, 2024.
\copyright IEEE}
\end{center}
\end{textblock}
\fi 

\author{Shuang~Guo~and~Guillermo~Gallego~\ifarxiv\orcidlink{0000-0002-2672-9241}\fi
\thanks{The authors are with the Dept. of Electrical Engineering and Computer Science of TU Berlin, Berlin, Germany.
G. Gallego is also with the Einstein Center Digital Future (ECDF) and the Science of Intelligence (SCIoI) Excellence Cluster, Berlin, Germany.
Funded by the Deutsche Forschungsgemeinschaft (DFG, German Research Foundation) under Germany’s Excellence Strategy – EXC 2002/1 ``Science of Intelligence'' – project number 390523135.}
\ifarxiv
\thanks{Manuscript submitted Feb 10, 2023; accepted Feb 15, 2024.}
\fi
}

\maketitle

\begin{abstract}
Event cameras are bio-inspired visual sensors that capture pixel-wise intensity changes and output asynchronous event streams. 
They show great potential over conventional cameras to handle challenging scenarios in robotics and computer vision, such as high-speed and high dynamic range. 
This paper considers the problem of rotational motion estimation using event cameras.
Several event-based rotation estimation methods have been developed in the past decade, 
but their performance has not been evaluated and compared under unified criteria yet.
In addition, these prior works do not consider a global refinement step.
To this end, we conduct a systematic study of this problem with two objectives in mind: summarizing previous works and presenting our own solution.
First, we compare prior works both theoretically and experimentally.
Second, we propose the first event-based rotation-only bundle adjustment (BA) approach.
We formulate it leveraging the state-of-the-art Contrast Maximization (CMax) framework, 
which is principled and avoids the need to convert events into frames.
Third, we use the proposed BA to build \cmaxslam{}, the first event-based rotation-only SLAM system comprising a front-end and a back-end.
Our BA is able to run both offline (trajectory smoothing) and online (\cmaxslam{} back-end).
To demonstrate the performance and versatility of our method, we present comprehensive experiments on synthetic and real-world datasets, including indoor, outdoor and space scenarios. 
We discuss the pitfalls of real-world evaluation and propose a proxy for the reprojection error as the figure of merit to evaluate event-based rotation BA methods.
We release the source code and novel data sequences to benefit the community.
We hope this work leads to a better understanding and fosters further research on event-based ego-motion estimation.
\end{abstract}

\section*{Video, Source Code and Data}
Project page: \url{https://github.com/tub-rip/cmax_slam}

\section{Introduction}
\label{sec:intro}

Event cameras are novel bio-inspired visual sensors that measure per-pixel brightness changes \cite{Lichtsteiner08ssc,Finateu20isscc}.
In contrast to the images/frames produced by traditional cameras, the output of an event camera is a stream of asynchronous events, 
$e_k = (\imgpoint_k,t_k,p_k)$, comprising the timestamp $t_k$, pixel location $\imgpoint_k \doteq (x_k, y_k)^\top$, and polarity $p_k$ of the brightness changes. 
This unique working principle offers potential advantages over frame-based cameras: high temporal resolution (in the order of \si{\micro\second}), high dynamic range (HDR) (\SI{140}{\decibel} vs. \SI{60}{\decibel} of standard cameras), temporal redundancy suppression and low power consumption (\SI{20}{\mW} vs. \SI{1.5}{\W} of standard cameras) \cite{Gallego20pami}. 
They are beneficial for challenging tasks in robotics and computer vision, such as ego-motion estimation \cite{Weikersdorfer12robio,Gallego17pami,Gallego17ral,Bryner19icra,Zhu19cvpr,Chamorro20bmvc,Jiao21cvprw} and Simultaneous Localization And Mapping (SLAM) \cite{Weikersdorfer13icvs,Kim14bmvc,Kim16eccv,Rebecq17ral,Reinbacher17iccp,Zhu17cvpr,Mueggler18tro,Rosinol18ral,Zhou20tro,Weipeng22iros,Chamorro22ral} in high-speed, extreme lighting, or HDR conditions. 

Rotational motion estimation is an essential problem in vision and robotics, and it acts as a foundation for higher order motion formulations (e.g., six degrees-of-freedom, 6-DOF). 
Despite decades of research, reliably estimating the motion of a purely rotating camera is still 
challenging, especially using frame-based cameras since motion blur, under/over-exposure and large inter-image displacements are likely to occur and ruin data association.
Conversely, event cameras do not suffer from blur, low dynamic range or large displacements between data bits \cite{Gallego20pami}, hence they can take on high-speed and/or HDR rotational motion estimation. 
The challenge consists of developing novel algorithms to process the unconventional, 
motion-dependent data produced by event cameras 
to reliably establish data association and thus enable robust motion estimation.

\begin{figure*}[t]
     \centering
     \includegraphics[width=.97\linewidth]{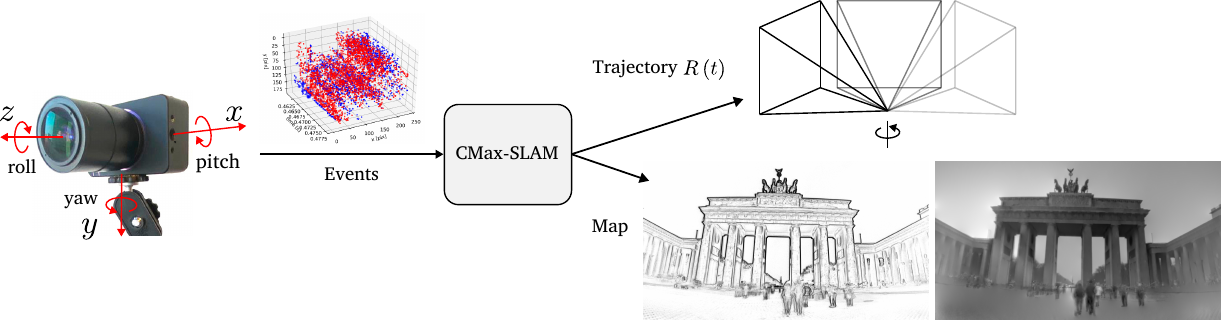}
\caption{The proposed \cmaxslam{} system takes as input the asynchronous event stream to estimate the rotational ego-motion of the camera while producing a sharp panoramic map (IWE) as by-product. 
We can further reconstruct a grayscale map by feeding the events and the estimated trajectory to the mapping module of \smt{} \cite{Kim14bmvc}.}
\label{fig:eyecatcher}
\end{figure*}

Several works have demonstrated the capabilities of event cameras to estimate rotational motion in difficult scenarios (high speed, HDR) \cite{Kim14bmvc,Gallego17ral,Reinbacher17iccp,Kim21ral}. 
Early work \cite{Kim14bmvc} proposed a 3-DOF simultaneous mosaicing and tracking method consisting of two Bayesian filters operating in parallel to estimate the camera motion and scene map. 
Also working in parallel, but using non-linear least squares, a real-time panoramic tracking and probabilistic mapping was presented in \cite{Reinbacher17iccp}.
Recently, \cite{Kim21ral} has extended the Contrast Maximization (CMax) framework in \cite{Gallego17ral} to simultaneously estimate angular velocity and absolute orientation.
However, these methods have not been compared or evaluated under unified criteria yet. 
Carrying out such a benchmark is challenging and useful to identify best practices that foster further improvements.
Besides, all these methods are short-term (i.e., they estimate the rotation for the current set of events), corresponding to the front-ends of SLAM systems \cite{Cadena16tro}, and therefore do not consider a long-term (i.e., global) refinement step.
Such a step, implemented in the back-end of modern SLAM systems, is a desirable feature because it improves accuracy and robustness.

\subsection*{Contributions}
This work aims at filling these gaps. 
Hence we conduct a systematic comparative study on rotational motion estimation using event cameras, summarizing previous works and presenting our own solution to the problem. 
We propose a novel rotation-only bundle adjustment (BA) approach for event cameras, which acts as a back-end. 
In contrast to the reprojection or photometric errors used in frame-based BA \cite{Szeliski10book}, we seek to exploit the state-of-the-art CMax framework \cite{Gallego18cvpr,Gallego19cvpr,Stoffregen19iccv,Liu20cvpr,Nunes20eccv,Peng20eccv,Gu21iccv,Nunes21pami,Zhou21tnnls,Peng21pami,Shiba22eccv,Ghosh22aisy,McLeod22eccvw,Shiba22aisy} for the task of BA. 
We explicitly model the camera trajectory continuously in time and use it to warp the events onto a global (panoramic) map. 
We cast the problem of event-based BA as that of finding the trajectory of the rotating camera that achieves the sharpest (i.e., motion-compensated) scene map. 
Hence, the map of the scene is internally obtained as a by-product of the optimization (see \cref{fig:eyecatcher}). 
Furthermore, the effectiveness of our BA is demonstrated for the refinement of the trajectories produced by the above-mentioned event-based rotation estimation front-ends.

We pair the proposed BA (back-end) with an event-based angular velocity estimator as front-end \cite{Gallego17ral} to implement an event-based rotational SLAM system, called \cmaxslam{}. 
We show that our system is able to work reliably on challenging synthetic and real-world datasets, including indoor and outdoor scenarios, while outperforming previous works.
Furthermore, we also show how our method can handle very different scenes, including the data produced by an event camera attached to a telescope, in a star tracking task.

In summary, our contributions are:
\begin{enumerate}
    \item We theoretically compare and experimentally benchmark several event-based rotational motion estimation methods under unified criteria, in terms of accuracy and efficiency (\cref{sec:compared_methods,sec:experiments}). 
    \item We propose the first event-based rotation-only BA method to refine the continuous-time trajectory of an event camera while reconstructing a sharp panoramic map of scene edges (\cref{sec:method:ideal_model}).
    \item We propose an event-based rotation-only SLAM system called \cmaxslam{}, comprising both a front-end and a back-end for the first time  (\cref{sec:method:pipeline}). 
    \item We demonstrate the method on a variety of scenarios: indoor, outdoor and Space applications, which shows the versatility of our approach  (\cref{sec:experiments,sec:space_application}).
    \item We highlight the potential pitfalls of evaluating rotation-only methods on non-strictly rotational data and propose a sensible figure of merit for event-based rotational BA in real-world scenarios (\cref{sec:experiments}).
    \item We release the source code and novel data sequences with high (VGA) spatial resolution.
\end{enumerate}

We hope our work leads to a better understanding of the field of event-based ego-motion estimation and helps taking on related motion-estimation tasks.
\section{Event-based rotation estimation methods}
\label{sec:compared_methods}

Let us review the event-based rotational motion estimation approaches involved in this benchmark. 
The primary features of these works are summarized and compared in \cref{tab:features}. 
For more details, we refer to \cite{Kim14bmvc,Gallego17ral,Reinbacher17iccp,Kim21ral}. 

\begin{table*}[ht]
\caption{Comparison of the main characteristics of the methods considered and the new one (last column).\label{tab:features}}
\adjustbox{max width=\textwidth}{
\begin{centering}
\begin{tabular}{llllll}
\toprule 
  & Kim et al. \cite{Kim14bmvc} (\smt{}) & Gallego et al. \cite{Gallego17ral} (\cmaxw{}) & Reinbacher et al. \cite{Reinbacher17iccp} (\rtpt{}) & Kim et al. \cite{Kim21ral} (\cmaxgae{}) & Ours (Back-end)\\

\midrule
Problem solved & Absolute poses & Angular velocity & Absolute poses & Angular velocity and absolute poses & Trajectory refinement\\

Tracking & PF/EKF & CMax (Local only) & Minimize photometric error \eqref{eq:rtpt:tracker} & CMax (Local and global) & CMax (Global)\\
 
Mapping & Grayscale panoramic map & Local IWE & Probabilistic panoramic map & 3D sphere of points & Panoramic (global) IWE\\

Is EGM used? & Yes \eqref{eq:smt:egm:tracker}, \eqref{eq:smt:egm:mapper} & No & Yes \eqref{eq:RTPT:probevent}, No, using \cite{Weikersdorfer13icvs} 
& No & No \\

Is polarity required? & Yes & Optional & No & Yes (for $\angvel$) /No (for $\Delta\Rot$) & No \\

Event processing & By event & By batch & By batch & By batch & By event (by batch for speed-up) \\

Bootstrapping & Random & N/A & Random & Run \cite{Gallego17ral} & N/A \\

\bottomrule
\end{tabular}
\par\end{centering}
}
\end{table*}

\begin{figure*}
\def\figHeight{2.84cm} 
     \centering
     \begin{subfigure}[b]{0.265\linewidth}
         \centering
         \gframe{\includegraphics[trim={40px 0 60px 0},clip,height=\figHeight]{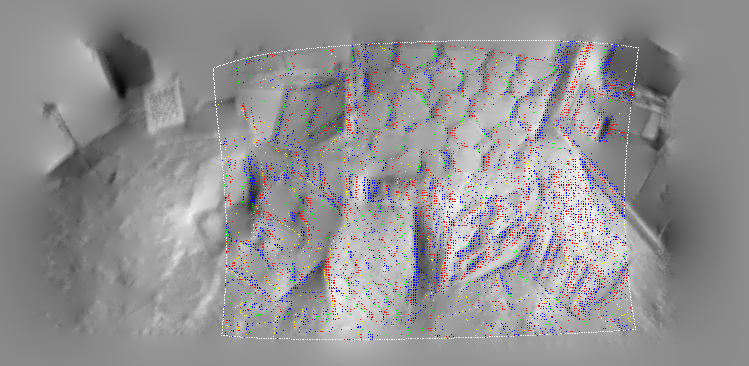}}
         \caption{\smt{} \cite{Kim14bmvc}}
         \label{fig:map_compare:SMT}
     \end{subfigure}
     \;
     \begin{subfigure}[b]{0.195\linewidth}
         \centering
         \gframe{\includegraphics[height=\figHeight]{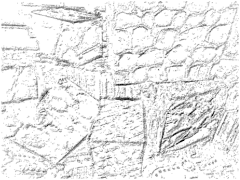}}
         \caption{\cmaxw{} \cite{Gallego17ral}}
         \label{fig:map_compare:RAL17}
     \end{subfigure}
     \;
     \begin{subfigure}[b]{0.22\linewidth}
         \centering
         \gframe{\includegraphics[trim={40px 20px 40px 20px},clip,height=\figHeight]{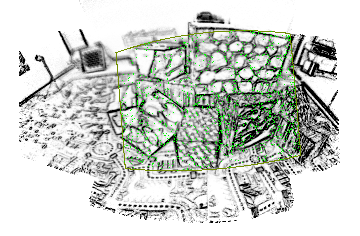}}
         \caption{\rtpt{} \cite{Reinbacher17iccp}}
         \label{fig:map_compare:RTPT}
     \end{subfigure}
     \;
     \begin{subfigure}[b]{0.25\linewidth}
         \centering
         \gframe{\includegraphics[height=\figHeight]{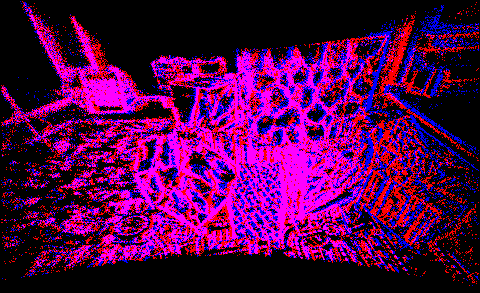}}
         \caption{\cmaxgae{} \cite{Kim21ral}}
         \label{fig:map_compare:CMAX-GAE}
     \end{subfigure}
    \caption{Maps produced by the rotation estimators involved in the benchmark.
    (a) Grayscale panoramic map by \cite{Kim14bmvc} (red and blue dots indicate positive and negative inlier events while yellow and green dots represent positive and negative outlier events in the current FOV respectively).
    (b) Local IWE after CMax by \cite{Gallego17ral}, where the darkness represents the amount of accumulated events. 
    (c) Probabilistic map by \cite{Reinbacher17iccp}, where dark means $\approx 1$ (edge), and white means $\approx 0$ (no edge). 
    The green dots indicate the events in the current FOV.
    (d) Global map containing projected visible events by \cite{Kim21ral} (blue and red dots have the same meaning as in \cref{fig:map_compare:SMT}, and pink dots indicate pixels with both positive and negative events).
    }
    \label{fig:map_compare}
\end{figure*}

\subsection{Event-based Simultaneous Mosaicing and Tracking (\smt{})}
\label{sec:compared_methods:bmcv14}

\smt{} \cite{Kim14bmvc} consists of two modules operating in a parallel tracking-and-mapping (PTAM) fashion \cite{Klein07ismar}. 
In the tracking module, a Particle Filter (PF) estimates the rotation of the camera given a map of the scene. 
The mapping module builds a grayscale panoramic map of the scene in two steps: 
first, Extended Kalman Filters (EKFs) incrementally update the grayscale gradient at each pixel of the panoramic map; 
second, Poisson integration recovers the absolute intensity (\cref{fig:map_compare:SMT}). 
Both modules collaborate by assuming that the output from each other is accurate. 
However, this assumption often does not hold, which leads to drift accumulation and tracking failure.

The event generation model (EGM) \cite{Gallego20pami} in its original form (for the PF) or in its linearized version (for the EKFs) is used in the measurement models of the Bayesian filters.
Specifically, the orientation tracker sets the likelihood of each event $p(e_k | \Rot)$ by computing the brightness increment 
\begin{equation}
\label{eq:smt:egm:tracker}
z = M( \mappointtk ) - M( \bp_{m}^{t_k-\Delta t_k} )
\end{equation}
and using it to look up the value of a Gaussian-shaped curve centered at $C$.
Here $M$ is the logarithm of the grayscale panoramic map 
and $\mappointtk \equiv \bp(\Rot(t_k), \imgpoint_k)$ is the map point obtained by transferring event coordinates $\imgpoint_k=(x_k,y_k)^\top$ according to the rotation at time $t_k$, $\Rot(t_k)$. 
In the mapping module, each map point's EKF sets the likelihood of each event $p(e_k | M)$ using the following measurement model:
\begin{equation}
\label{eq:smt:egm:mapper}
z = \frac{1}{\Delta t_k}, \qquad 
h(\bg) = \frac{-\bg \cdot \velflow}{\pol_k C},
\end{equation}
where the state is $\bg \doteq \nabla M( \mappointtk )$, i.e., 
the spatial gradient of the intensity at the map point corresponding to the current event $e_k$ and rotation $\Rot(t_k)$.
The velocity is given by $\velflow \approx \Delta \mappointtk / \Delta t_k$, 
where $\Delta \mappointtk \doteq \mappointtk - \bp_{m}^{t_k-\Delta t_k}$ approximates the displacement traveled by the spatial gradient during $\Delta t_k$.

\subsection{Event-based Angular Velocity Estimation}
\label{sec:compared_methods:ral17}

Gallego et al.~\cite{Gallego17ral} propose estimating the angular velocity $\angvel$ of a rotating camera by aligning events corresponding to the same scene edge.
Events $\cE=\{e_k\}_{k=1}^{\numEvents}$ are processed in packets and warped according to the candidate motion, 
$e_k \mapsto e'_k \doteq \Warp(e_k; \angvel)$, yielding an image of warped events (IWE) $I(\angvel, \cE)$.
Then the problem is formulated as finding the angular velocity that achieves the sharpest IWE (\cref{fig:map_compare:RAL17}). 
Several alignment objective functions are available \cite{Gallego19cvpr,Stoffregen19cvpr}. 
The variance of the IWE (i.e., image contrast) is adopted in \cite{Gallego17ral} to quantify event alignment, and the non-linear conjugate gradient method of Fletcher and Reeves (CG-FR) is used as optimizer:
\begin{equation}
    \angvel^\ast = \mathop{\arg\max}_{\angvel} {\rm Var} \left(I(\angvel,\cE)\right),
\label{eq:ral17}
\end{equation}
where $\cE$ is the current packet of events.
 
The sharp IWE acts as a local map of the scene. 
Since \cite{Gallego17ral} focuses on estimating angular velocity rather than absolute orientation, it does not need a global (panoramic) map.
The method assumes events are caused by moving edges, and thus the sharp IWE is the spatial gradient entangled with the motion \cite{Gallego18cvpr,Zhang22pami}: 
$|I(\angvel^\ast,\cE)| \propto |\nabla \Lum \cdot \velflow|$ (not simply $|\nabla L|$), where $\Lum$ is the log-brightness on the sensor's image plane.

\subsection{Real-Time Panoramic Tracking for Event Cameras (\rtpt{})}
\label{sec:compared_methods:iccp17}

Similarly to \smt{}, \rtpt{} also has two parallel modules for tracking and mapping. 
The tracking part uses direct alignment techniques \cite{Engel17pami} to minimize the error between the current events (``local map'') and the global map built from past events:
\begin{equation}
\label{eq:rtpt:tracker}
\arg\min_{\bparams} \frac1{2} \sum_{k=1}^{\numEvents} \bigl(1-M(\pi(\imgpoint_k; \bparams))\bigr)^2,
\end{equation}
where $\pi(\imgpoint_k; \bparams) \equiv \mappointtk$ is the map point corresponding to the current event and rotation parameters $\bparams$ (e.g., exponential coordinates), 
and $M$ is the global (panoramic) map.
Hence the tracking problem is formulated as a non-linear least squares (NLLS) problem.

The mapping module builds a probabilistic 2D map of scene edges $M(\bp)$ (or ``spatial event rate''), where higher values indicate events are more likely to be produced when sensor pixels cross that map point (\cref{fig:map_compare:RTPT}).
The values are interpreted as probabilities, hence they are bounded, $M \in [0,1]$, 
and they are designed so that the residual \eqref{eq:rtpt:tracker} vanishes if the rotated events $\mappointtk$ (i.e., local map) align with high values of the edge map $M$.
The map is theoretically justified using the linearized EGM, deriving a formula for the probability of an event being triggered by a map point $\bp$ (factoring out the camera motion by integrating over a uniform prior) \cite[Eq.(7)]{Reinbacher17iccp}:
\begin{equation}
\label{eq:RTPT:probevent}
P(e | \bp) = 1 - \frac{2}{\pi} \arcsin \left(\frac{C}{\|\nabla \log I(\bp)\|}\right),
\end{equation}
where $\log I(\bp)$ is the logarithm of the intensity at $\bp$.
This probability curve model grows with a concave shape from 0 at $\|\nabla \log I(\bp)\|/C = 1$ to 1 for $\|\nabla \log I(\bp)\| \gg 1$.
In practice, the map is built using a frequentist approach from prior work~\cite{Weikersdorfer13icvs}:
$M(\bp) =  O(\bp) / N(\bp)$, 
where $O(\bp)$ and $N(\bp)$ count the number of triggered events and the number of possible event occurrences at every map point, respectively.

Actually, the event stream is fed to the tracking module in packets to obtain a pose/rotation update using a Gauss-Newton--type scheme.
Then the mapping module updates the map using the whole event packet. 
In this benchmark, \rtpt{} is the only one whose code relies on GPU for acceleration.

\subsection{CMax over Globally Aligned Events (\cmaxgae{})}
\label{sec:compared_methods:ral21}

Kim et al. \cite{Kim21ral} extend the method in \cite{Gallego17ral} to jointly estimate angular velocity and absolute rotation. 
That is, besides aligning all events in the current packet, the local IWE is also aligned to a global IWE by solving the problem
\begin{equation}
    \mathop{\arg\max}_{\boldsymbol{\omega}, \Delta \Rot}\; \lVert I_L\left(\boldsymbol{\omega}, \Delta \Rot\right) + I_G\left(\Rot\right) \rVert^{2},
\label{eq:ral21_obj}
\end{equation}
where $\boldsymbol{\omega}$ and $\Delta \Rot$ are the angular velocity 
and incremental pose change for the \emph{current} event packet,
$\Rot$ is the camera pose maintained during the running sequence,
$I_L$ is the local IWE rotated by $\Delta \Rot$ so that it is aligned to the global map, 
and $I_G$ is the global map centered at $I_L$ (twice as high and wide as $I_L$). 
$I_G$ is obtained by projecting all visible past events (stored on the unit sphere, which acts as the scene map) onto the image plane (\cref{fig:map_compare:CMAX-GAE}). 

Once the optimal incremental rotation $\Delta \Rot^\ast$ has been obtained for the current packet by solving \eqref{eq:ral21_obj}, the camera pose is updated in preparation for the next packet of events:
\begin{equation}
    \Rot \gets \Rot \; \Delta \Rot^\ast.
\label{eq:ral21_update}
\end{equation}

Note that in \cite{Kim21ral} the Mean Square (MS) metric is adopted to compute the contrast of the IWE, and the RMSProp optimizer is employed to maximize the objective function (instead of the variance and CG-FR method used in \cite{Gallego17ral} and in \cref{sec:compared_methods:ral17}).

\subsection{Discussion of the State-of-the-Art Motion Estimators}

Let us highlight some of the pros and cons of each approach.
\smt{} \cite{Kim14bmvc} recovers the richest form of scene map (absolute brightness panorama), in which the EGM is used and the contrast threshold $C$ is required. 
However, for real event cameras, it is difficult to accurately know the value of $C$ and it varies greatly even within a single dataset \cite{Stoffregen20eccv}. 
In addition, the error propagation between tracking and mapping parts is inevitable, which is especially noticeable when the camera revisits past regions of the scene.

\rtpt{} emphasizes the idea that event polarity is not needed for panoramic tracking; only the space-time coordinates of the events are needed.
Hence the resulting map (an edge map) is significantly different from the grayscale map in \smt{}.
Formulating the tracking problem as NLLS enables the use Gauss-Newton's method and acceleration to converge faster to the motion parameters.
Thanks to the probabilistic paradigm, \rtpt{} is, to some extent, robust to scenes with moving objects. 
However, the implementation of this approach stores all events throughout the whole running process, which is memory-consuming and intractable for long-term tracking. 
Both PF-\smt{} \cite{Kim14bmvc} and \rtpt{} require a GPU to run at moderate speeds. 

\begin{figure*}[t]
\centering
\includegraphics[width=.99\linewidth]{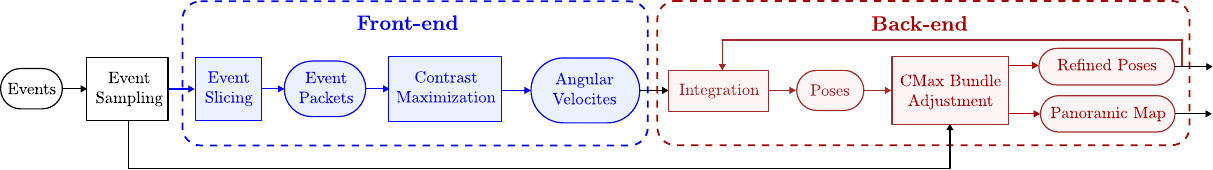}
\caption{
Overview of the proposed rotational event-based SLAM system. 
}
\label{fig:pipeline}
\end{figure*}

The methods based on CMax \cite{Gallego17ral,Kim21ral} do not require the specification of the contrast threshold $C$.
These methods, while not being formulated as NLLS (to exploit fast convergence of Gauss-Newton), run at moderate speeds on standard CPUs, i.e., without requiring a GPU.
\cmaxw{} \cite{Gallego17ral} continuously provides accurate angular velocity estimations 
and does not suffer from error propagation since only a local IWE is needed. 
However, it does not consider absolute pose estimation, which may give rise to accumulated drift if the velocities were integrated into poses.
\cmaxgae{} \cite{Kim21ral} partially reduces drift by additionally introducing a global alignment term in the CMax objective function. 
However, it maintains a 3D sphere consisting of all past events as well as their visibility in the current FOV, which takes up a lot of memory and, therefore, does not support long-term tracking.

Our method (last column of \cref{tab:features}) builds on the strengths of previous works:
specifying $C$ is not needed,
building an edge map is sufficient for ego-motion estimation,
a global map helps reduce drift,
storing all events is avoided because it is expensive (time- and memory-wise).
We also observe that an explicit continuous trajectory model is currently missing in the event-based rotation-only estimation literature, hence we incorporate it to facilitate the formulation of the solution.

\section{Methodology}
\label{sec:method}

This section first describes the theoretical formulation of the CMax-based BA (\cref{sec:method:principle,sec:method:ideal_model}). 
Then it introduces the pipeline of \cmaxslam{} (\cref{sec:method:pipeline}), including both front-end (\cref{sec:method:pipeline:frontend}) and back-end (\cref{sec:method:pipeline:backend}), with a focus on how to adapt the proposed BA to an online system. 
Finally, the characteristics of our method are briefly discussed (\cref{sec:method:discussion}). 

\subsection{The Principle of CMax for Rotational Motion Estimation}
\label{sec:method:principle}
Our method builds on the idea of event alignment put forward by the CMax framework \cite{Gallego17ral,Gallego18cvpr}.
Assuming constant illumination, when the event camera rotates relative to the scene, 
edge patterns on the image plane move along point trajectories defined by the camera motion, producing events.
Simply summing events pixel-wise or along arbitrary point trajectories yields a blurred image (or histogram) of warped events (IWE). 
Instead, summing events along the point trajectories defined by the camera motion yields a sharp image of the (motion-compensated) edges that caused the events. 
This insight is leveraged in \cite{Gallego17ral}: 
during the time span of a small packet of events, the point trajectories can be parametrized by a constant angular velocity $\angvel$ model 
and CMax recovers the camera's velocity by searching for the $\angvel$ whose point trajectories best align with the events, producing the sharpest IWE. 
Shiba et al. \cite{Shiba22sensors,Shiba22aisy} shows that the point trajectories defined by this angular velocity warp are well-behaved, not suffering from ``event collapse'' (i.e., the IWEs cannot be ``over-sharp'').
As we show, the above idea can be extended to longer time intervals, and consequently, 
more complex ways to parametrize the camera motion $\Rot(t)$ (than simply constant $\angvel$) are needed.
A continuous-time model $\Rot(t)$ is the most generic one we can think of. 
While the CMax objective function seems to be ill-posed in the general (\mbox{6-DOF}) camera motion case \cite{Shiba22sensors,Shiba22aisy}, 
our method does not suffer from event collapse. 
The proof is given in Appendix~\ref{app:nocollapse}.

\subsection{Event-based Bundle Adjustment}
\label{sec:method:ideal_model}
Bundle Adjustment (BA) is a mature technique in frame-based visual SLAM \cite{Triggs00,Cadena16tro},
where it is usually formulated as the minimization of the reprojection error or the photometric error, so as to refine camera poses, scene structure and even calibration parameters as well. 
In contrast, the asynchronous event stream contains neither trackable motion-invariant features nor intensity information, so frame-based BA algorithms are not directly applicable to event cameras.

Inspired by \cite{Gallego17ral,Gallego18cvpr}, which recover local motion parameters by aligning events in a short-time IWE, we leverage CMax to formulate event-based BA as follows. 
The camera trajectory is described by a continuous-time model $\Rot(t)$. 
Each event $e_k$ is warped according to its corresponding rotation $\Rot(t_k)$, projected and accumulated into a global IWE. 
Then we define the rotational BA problem as finding the camera trajectory $\Rot(t)$ that maximizes event alignment (e.g., sharpness) of the global map, as measured by some objective function $f$ \cite{Gallego19cvpr}:
\begin{equation}
    \mathop{\arg \max}_{\Rot(t)} f\bigl( I(\Rot(t); \cE)\bigr).
    \label{eq:ideal_model}
\end{equation}

Solving \eqref{eq:ideal_model} yields the refined camera trajectory $\Rot^\ast(t)$ and its associated sharp map $M \equiv I^\ast$ (i.e., motion-compensated). 
Since the map is fully determined by the events and the camera trajectory, the search only needs to be carried out over the space of camera trajectories, as opposed to searching over a larger space of camera trajectories and maps.

\subsubsection{Trajectory Parameterization}
As presented in \cite{Mueggler18tro}, due to the high temporal resolution and asynchronous nature of the event camera it is suboptimal to estimate or refine its trajectory as a set of discrete-time poses. 
To this end, we represent the continuous-time camera trajectory in \eqref{eq:ideal_model} using linear or cubic B-splines, which are parameterized by a set of temporally equispaced control poses in $\mathit{SO}(3)$. 
Using B-splines as a model of continuous-time trajectory has advantages:
(i) they reduce the number of pose states (a few control poses vs.~a different pose per event/timestamp);
(ii) they have a local support (each control pose has a limited influence on the overall trajectory);
(iii) they unify, allowing us to choose the degree of smoothness of the trajectories 
by varying one parameter (the order of the spline); 
(iv) they enable the fusion of data from different sensors (synchronous and asynchronous), in a principled way, 
querying each sensor at the precise timestamp of their measurements, 
and (v) they also facilitate the estimation of the time offset between sensors.

The pose at any time of interest $\Rot(t)$ is obtained by interpolation of a certain number of neighboring control poses $\Rot(t_i)$. 
For \emph{linear} splines, the pose at $t\in\left[t_i,t_{i+1}\right)$ is simply the linear interpolation in the Lie group sense of the two adjacent control poses $\Rot\left(t_i\right)$ and $\Rot\left( t_{i+1} \right)$, given by \cite{Barfoot15book}:
\begin{equation}
    \Rot(t) = \Rot(t_i) \exp \left( \frac{t-t_i}{t_{i+1}-t_i} \log \left( \Rot^\top(t_i) \,\Rot (t_{i+1}) \right) \right).
    \label{eq:linear_interp}
\end{equation}

Although linear splines are continuous, they have limited smoothness, hence many control poses are required to achieve sufficient smoothness.
Therefore, we also employ higher-order splines: cubic splines are smoother than linear ones. 

In the case of \emph{cubic} B-splines, the interpolation of the pose at $t\in[t_i,t_{i+1})$
requires four control poses, which occur at $\{t_{i-1}, t_{i}, t_{i+1}, t_{i+2}\}$. 
Following the cumulative cubic B-spline formulation \cite{PatronPerez15ijcv,Mueggler18tro}, 
we write the camera trajectory as: 
\begin{equation}
    \Rot(u(t)) = \Rot_{i-1} \prod_{j=1}^{3} \exp \left( \tilde{\mathbf{B}}_j(u(t)) \mathtt{\Omega}_{i+j-1} \right),
    \label{eq:cubic_interp}
\end{equation}
where $u(t) \doteq (t-t_i) / \Delta t \in [0,1)$ is the normalized time representation, 
$\Delta t \doteq t_{i+1} - t_{i}$,
\begin{equation}
    \mathtt{\Omega}_i \doteq \log \left( \Rot_{i-1}^{\top} \Rot_i \right)
\label{eq:incre_pose}
\end{equation}
is the incremental pose between two consecutive control poses,
and $\tilde{\mathbf{B}}_j$ indicates the $j$-th entry (0-based) of $\tilde{\mathbf{B}}$,
which is the matrix representation of the cumulative basis functions for the cubic B-splines:
\begin{equation}
    \tilde{\mathbf{B}}(u) \doteq \frac{1}{6}
    \begin{pmatrix}
        6 & 0 & 0 & 0 \\
        5 & 3 & -3 & 1 \\
        1 & 3 & 3 & -2 \\
        0 & 0 & 0 & 1
    \end{pmatrix}\!
    \begin{pmatrix}
        1 \\
        u \\
        u^2 \\
        u^3
    \end{pmatrix}.
\label{eq:basis_func}
\end{equation}

We denote the control poses of the linear and cubic splines slightly differently ($\Rot(t_i)$ and $\Rot_{i}$ respectively) because linear splines pass through the control points whereas cubic splines do not, in general.

Despite the elegance of the continuous-time trajectory representation, 
its high computational complexity (especially the derivative computation for trajectory optimization) presents a challenge for the application of cubic B-splines in trajectory estimation. 
We tackle this by adopting an efficient method recently presented in \cite{Sommer20cvpr} to compute the analytical derivatives of an $\mathit{SO}(3)$ spline trajectory with respect to its control poses. 
In this way, the complexity of the derivative computation becomes linear in the order of the spline, 
which promotes the efficiency of the proposed BA approach.

\subsubsection{Panoramic IWE}
\label{sec:panoramicIWE}
The global map $I$ in \eqref{eq:ideal_model} is generated by warping and counting aligned events on a panoramic image.
Every incoming event $e_k$ is rotated according to its corresponding pose $\Rot(t_k)$ on the spline trajectory:
\begin{equation}
    \bX_k' = \Rot\left(t_k\right) \bX_k,
\label{eq:event_warping}
\end{equation}
where $\bX_k = \mK^{-1}\,(\imgpoint_k^\top,1)^\top$ is the bearing direction (3D point) of $e_k$ in calibrated camera coordinates, and $\mK$ is the intrinsic parameter matrix of the camera. 
The rotated point $\bX_k' = (X'_k,Y'_k,Z'_k)^\top$ is then projected onto the panorama using the equirectangular projection:
\begin{equation}
    \bp_k \doteq
    \begin{pmatrix}
        \frac{w}{2} + \frac{w}{2\pi} \arctan \left( \frac{X_k'}{Z_k'}\right) \\
        \frac{h}{2} + \frac{h}{\pi} \arcsin \left( \frac{Y_k'}{\sqrt{X_k'^2 + Y_k'^2 + Z_k'^2}} \right)
    \end{pmatrix},
\label{eq:equirect_projection}
\end{equation}
where $w$ and $h$ are the width and height of the panoramic~map. 
The mapping $\imgpoint_k \stackrel{\Warp}{\mapsto} \bp_k$ defines the event warping from the sensor image plane to the panoramic map. 

Finally, the warped event is accumulated on the panoramic map $I$ using bilinear voting since it may not have integer coordinates \cite{Gallego17ral,Rebecq18ijcv}.
Note that event polarities are not used.

\subsection{\cmaxslam{} System}
\label{sec:method:pipeline}

The proposed rotational SLAM system consists of two parts (see \cref{fig:pipeline}). 
The \emph{front-end} takes raw events as input and estimates the camera's angular velocity $\angvel$ using CMax \cite{Gallego17ral}.
The \emph{back-end} integrates $\angvel$ to obtain a set of (absolute) rotations that is used to initialize the camera trajectory, which is refined via the above BA method.

\vspace{1ex}
\subsubsection{Front-end}
\label{sec:method:pipeline:frontend}

\begin{figure}[t]
     \centering
     \begin{subfigure}[c]{\linewidth}
         \centering
         \includegraphics[width=\linewidth]{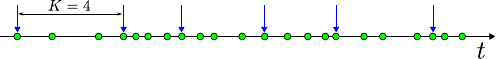}
         \caption{Constant event number $(K=4)$.}
         \label{fig:event_slicing:const_ev_num}
         \vspace{0.25cm}
     \end{subfigure}
     \begin{subfigure}[c]{\linewidth}
         \centering
         \includegraphics[width=\linewidth]{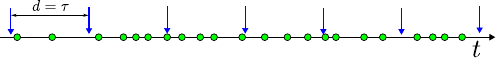}
         \caption{Constant duration $(d=\tau)$.}
         \vspace{0.25cm}
         \label{fig:event_slicing:const_duration}
     \end{subfigure}
     \begin{subfigure}[c]{\linewidth}
         \centering
         \includegraphics[width=\linewidth]{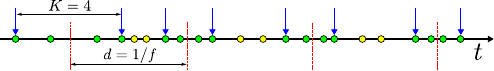}
         \caption{Ours $(K=4,\, d=1/f)$.}
         \label{fig:event_slicing:ours}
     \end{subfigure}
        \caption{Three event slicing strategies: (a) constant event count, (b) constant duration, 
        and (c) proposed strategy that selects a fixed number of events around equispaced timestamps (with output frequency $f$). 
        Green and yellow dots represent processed and skipped events, respectively. 
        Blue arrows are the boundaries of the event slices. 
        Red dashed lines indicate the selected timestamps for angular velocity estimation.
        \label{fig:event_slicing}}
        \vspace{-2ex}
\end{figure}

We adopt the CMax-based angular velocity estimator (\cref{sec:compared_methods:ral17}) as front-end of the proposed SLAM system and modify it to work with the back-end.
This method assumes constant angular velocity within each packet (or ``slice'') of events, so a proper slicing strategy of the event stream is critical. 
Two commonly-used possibilities are constant event count and constant time span \cite{Liu18bmvc}. 
The former creates packets of constant number of events $K$ (see \cref{fig:event_slicing:const_ev_num}),
which would guarantee sufficient data for every angular velocity estimate; 
however, the output rate would vary wildly depending on the event rate. 
The latter strategy, with packets of fixed time duration $d$ (see \cref{fig:event_slicing:const_duration}), 
would guarantee a fixed rate of angular velocity estimates, but accuracy would be unstable 
(a slow-moving camera would produce few events per packet, which would not be enough to reliably estimate motion). 

In our system, the back-end requires the front-end to provide pose estimates with constant rate and stable accuracy to achieve reliable trajectory initialization.
Hence, we combine the above two strategies. 
As shown in \cref{fig:event_slicing:ours}, we select a series of equispaced timestamps (i.e., constant frequency) at which the angular velocity will be estimated, and then slice a fixed number of events around each timestamp.
This event slicing strategy is similar to that in \cite{Rosinol18ral}; 
the difference lies in the fact that our event packets are centered at the selected timestamps, while the packets in \cite{Rosinol18ral} consist of the events before the selected timestamps.
If the time span of a single slice is too large (in our implementation, $10/f$, where $f$ is the angular velocity frequency), we assume the camera is not moving and therefore set the angular velocity to zero.

As depicted in \cref{fig:pipeline}, before event slicing we may perform an optional downsampling of the events,
(e.g., keeping one out of $Q$ events, where $Q$ is the event sampling rate), 
to allow for flexibly trading off accuracy and speed.

\vspace{1ex}
\subsubsection{Back-end}
\label{sec:method:pipeline:backend}

The BA introduced in \eqref{eq:ideal_model} is described as trajectory refinement, in which we already have all event data and an initial trajectory to start the optimization search. 
Next, let us explain how to adapt \eqref{eq:ideal_model} to an online SLAM system.

\paragraph{Trajectory Refinement via a Sliding Window}
There are two issues with the BA defined in \eqref{eq:ideal_model} that prevent its direct application to the back-end of a real SLAM system.
First, if the trajectory is temporally long, the amount of events and the number of control poses that need to be optimized will be very large, becoming intractable.
Second, the initial trajectory is required to be close enough to the real one, otherwise the initial panoramic map will be blurred, which may cause BA optimization to get stuck in a local optimum.
To this end, the BA in our back-end follows the structure of mature SLAM frameworks \cite{Cadena16tro}, working in a \emph{sliding-window} fashion (fixed time window).
Past control poses that fall outside the current time window are fixed,
while new poses in the current time window are initialized and refined as follows.

\emph{Initialization}. Once the angular velocities estimated by the front-end are available for the current time window, 
the back-end integrates them into a set of poses (starting from the last optimized pose in the previous window)
and initializes a new trajectory segment by fitting a spline curve to such front-end poses.
Given the $\mathit{SO}(3)$ nature of the problem, 
we follow a ``lift-solve-retract'' approach \cite{Forster17troOnmanifold,Barfoot15book} to solve for the control poses of the new trajectory segment:
\begin{enumerate}
    \item \emph{Lift}: use the first new front-end pose as offset, and compute the rotation increments with respect to it, so that the problem is lifted to the tangent vector space.
    \item \emph{Solve}: form the linear system \cite{Shene14spline} in the tangent space, and solve for the control rotation increments.
    \item \emph{Retract}: apply the offset to the control rotation increments, so that control poses in $\mathit{SO}(3)$ are obtained.
\end{enumerate}

\emph{Refinement}. After initialization, all events in the current time window are projected onto the global, panoramic IWE.
Following the optimization goal of the proposed BA (i.e., maximizing the contrast of the IWE), leads to an objective that only depends on the camera trajectory:
\begin{equation}
     \mathop{\arg \max}_{\left\{{\Rot_i}\right\}_{i \in \cR}}\; {\rm Var}\bigl(I_L\left(\left\{{\Rot_i}\right\}_{i \in \cR}, \cE_\text{curr} \right) + I_G \left( \cE_\text{prev} \right) \bigr),
\label{eq:init_obj_func}
\end{equation}
where $\cR$ is the set of indices of the varying control poses and $\cE_\text{curr}$ are the current events, which are projected to generate the local map $I_L$. 
The global map $I_G$ is built from all past events $\cE_\text{prev}$ and settled poses. 

\paragraph{Balancing Current and Past Events}
The current segment of the camera trajectory is refined by solving the BA problem \eqref{eq:init_obj_func} using CG-FR.
In practice, the number of events accumulated into the panoramic map increases rapidly as the camera rotates, which could lead to two issues.
First, the influence of the local map $I_L$ decreases with respect to the growing values of $I_G$, which may encourage local events to align to wrong edges of the global map. 
Second, if the camera kept moving around some regions while rarely observing others,
the majority of events would accumulate in the over-observed regions.
This could lead to a highly unbalanced edge map, especially for long-term tracking.
The following strategies mitigate the above two issues.

Regarding the first issue, we replace $I_L + I_G$ in \eqref{eq:init_obj_func} 
by a weighted sum, $I_L + \alpha I_G$, with $\alpha$ adaptively calculated from the event densities of $I_L$ and $I_G$. 
The event density $\rho(H)$ of an IWE $H$ is defined as the ratio of the number of events $N_e(H)$ to the image area that they occupy (event area, EA)~\cite{Gallego19cvpr}:
\begin{equation}
    \rho(H) \doteq N_e(H) \,/\operatorname{area}{}(H).
    \label{eq:event_density}
\end{equation}
We adopt the formula \cite{Gallego19cvpr}: 
$\operatorname{area}{}(H) = 1 - \exp\left(-(H(\bx)/\lambda_0\right)$, 
that is, pixels with $H(\bx) \gtrsim \lambda_0$ warped events contribute more to the area functional than those with $H(\bx) \lesssim \lambda_0$. 
Empirically, we use $\lambda_0 = 1$.
From the event densities of $I_L$ and $I_G$, the weight $\alpha$ is computed as
\begin{equation}
\label{eq:alphadensityratio}
    \alpha \doteq \rho(I_L) \,/\, \rho(I_G).
\end{equation}
Using \eqref{eq:event_density}--\eqref{eq:alphadensityratio}, 
we may rewrite the objective function for the \emph{sliding-window CMax-based BA} \eqref{eq:init_obj_func} as:
\begin{equation}
    \mathop{\arg\max}_{\{\Rot_i\}_{i \in \cR}}\; {\rm Var}\bigl(I_L(\{\Rot_i\}_{i \in \cR}, \cE_\text{curr}) + \alpha \, I_G( \cE_\text{prev}) \bigr).
\label{eq:new_obj_func}
\end{equation}
Note that $I_L$ would change during the optimization (and so would $\alpha$), 
however, the event density variations with respect to the initial $I_L$ (aligned by the \cmaxw{} front-end) are small, 
hence we use the latter to compute $\alpha$ and reutilize it during the optimization for each time window in \eqref{eq:new_obj_func}.

Regarding the second issue, we may stop updating the pixels of $I_G$ once we received sufficient evidence for them.
In the implementation, we monitor how long each map pixel is observed and stop accumulating events onto the pixel once the duration reaches a threshold (e.g., 10 s in our experiments).
A very low event rate in the current time window serves as a proxy to determine whether the camera is stationary, so that the BA is not performed, and consequently disregard such observation duration.
Note that this mitigation strategy is optional.
After the optimization, the image $I_G$ contains the sharp map $M$ produced by this approach.

For efficiency in \eqref{eq:new_obj_func}, we interpolate poses at the midpoint of each small batch of events (e.g., 100 events) and use it for all events in the batch, which reduces a lot the number of times for querying the camera pose and its derivatives in the spline trajectory.
This does not spoil the nature of continuous-time trajectory because the event rate is usually much higher (the average time span of a batch of 100 event is usually less than 0.5~ms).
As shown in \cref{fig:pipeline}, we can also perform (optional) event downsampling for the back-end, to trade off accuracy and speed depending on the application.

\paragraph{Analytical Derivatives and Locality Exploitation}
Derivatives are analytically computed during optimization:
\begin{equation}
    \frac{\partial}{\partial \bPhi} {\rm Var} \left( I\left(\bp; \bPhi \right) \right)
    = \frac{1}{|\Omega|} \int_{\Omega} 2\rho(\bp; \bPhi) \frac{\partial \rho(\bp; \bPhi)}{\partial \bPhi} d \bp
    \label{eq:contrast_deriv}
\end{equation}
where $\rho(\bp; \bPhi) \doteq I(\bp; \bPhi) - \mu\left( I(\bp; \bPhi) \right)$, 
and $\bPhi \doteq \{\Rot_i\}_{i \in \cR}$ is the set of all involved control poses. 
By the chain rule, 
\begin{equation}
    \frac{\partial I (\bp; \bPhi)}{\partial \bPhi}
    = -\sum_k \frac{\partial I}{\partial \bp_k}
    \; \frac{\partial \bp_k}{\partial \bX_k'}
    \; \frac{\partial \bX_k'}{\partial \bPhi}.
\label{eq:iwe_deriv}
\end{equation}

Due to the locality of B-splines, the contribution of each event to the panoramic IWE only depends on a few neighboring control poses (e.g., two for linear splines and four for cubic ones \cite{Mueggler18tro}).
Therefore, for each event, the Jacobian $\partial \bX_k'/\partial \bPhi$ in \eqref{eq:iwe_deriv} is sparse, with only few non-zero elements, which is beneficial for speeding up the computation of the derivatives. 

While $I = I_L + \alpha I_G$ in \eqref{eq:new_obj_func}, note that the derivative \eqref{eq:iwe_deriv}  only needs to be computed for $I_L$, since $I_G$ does not depend on the varying poses $\bPhi$ (i.e., its derivative vanishes).

\subsection{Discussion}
\label{sec:method:discussion}
The CMax-based BA defined in \eqref{eq:ideal_model} is an ideal model. 
It suggests the ``best'' theoretical trajectory under assumptions of purely rotational motion, perfect representation of the continuous camera motion, perfect projection model (including calibration), and lack of noise, which are unachievable conditions in the real world. 
We have proposed several techniques (i.e., approximations) to turn \eqref{eq:ideal_model} into a tractable model, \eqref{eq:new_obj_func}.

The \cmaxslam{} system does not have a separate mapping part, and the maintained global panoramic map is just a by-product which is only used for trajectory refinement. 
The system also does not rely on an explicit bootstrapping step.

The consistency between front-end and back-end is also notable. 
Both have the same goal (CMax), albeit with different motion parametrizations: constant angular velocity for short intervals (front-end) vs.~arbitrary continuous-time rotation (spline-based) for longer time intervals (back-end).
In both cases the IWE acts as an edge-like brightness map that is used for motion estimation.

While \cite{Kim21ral} also uses CMax, 
it operates on each incoming time interval of events (of 25 ms), 
estimating the angular velocity and absolute rotation for that interval alone. 
It works on a discrete set of poses that are updated one at a time \eqref{eq:ral21_update}.
Hence it has no continuous trajectory model and no capability to refine past poses (i.e., a back-end). 

\section{Experiments}
\label{sec:experiments}

This section presents a comprehensive comparative evaluation of the rotation-only motion estimators reviewed in \cref{sec:compared_methods}, as well as our proposed BA and \cmaxslam{} system.
First, we introduce the experimental setup, including datasets (\cref{sec:experiments:setup}) and metrics (\cref{sec:experiments:metrics}).
Then, we present the results on synthetic (\cref{sec:experiments:synth})
and real-world data (\cref{sec:experiments:real}), discussing the issues of the latter.
We also assess the runtime of the methods (\cref{sec:experiments:runtime}),
demonstrate \cmaxslam{} in complex scenes (\cref{sec:experiments:wild_experiments}) and show its super-resolution capabilities (\cref{sec:experiments:super_resolution}).
Finally, we present a sensitivity analysis (\cref{sec:experiments:sensitive_analysis}).

\subsection{Experimental Setup}
\label{sec:experiments:setup}

\subsubsection{Datasets}
\label{sec:experiments:datasets}
To evaluate the accuracy and robustness of rotation estimation, we conduct experiments on six synthetic sequences as well as six real-world sequences from standard datasets \cite{Mueggler17ijrr,Kim21ral}. 
All of them contain events, frames (not used), IMU data, and groundtruth (GT) poses.

The synthetic sequences are generated using the panoramic renderer of the ESIM simulator \cite{Rebecq18corl}, with input panoramas downloaded from the Internet. 
The generated sequences cover indoor, outdoor, daylight, night, human-made and natural scenarios.
The sizes of the input panoramas vary from 2K (\playroom{}), 4K (\bicycle{}), 6K (\city{} and \street{}), to 7K (\town{} and \bay{}) resolution.
In the simulation, \playroom{} is a classical sequence, generated with a DVS128 camera model ($128 \times 128$ px) and a duration of 2.5s,
while the other five sequences are generated with a DAVIS240C camera model (240 $\times$ 180 px) and a duration of 5~s.

All six real-world datasets are recorded by DAVIS240C event cameras, whose IMU operates at 1~kHz.
The Event Camera Dataset (ECD) \cite{Mueggler17ijrr} provides four \mbox{60 s} long rotational motion sequences: \shapes{}, \poster{}, \boxes{} and \dynamic{}. 
The first three sequences feature static indoor scenes with increasing texture complexity 
(yielding about 20 -- 200 million events), while the last one presents a dynamic scene ($\approx$ 70 million events). 
The camera moves with increasing speed, first rotating around each axis and then rotating freely in \mbox{3 DOFs}.
A motion capture (mocap) system outputs groundtruth poses with millimeter precision at \mbox{200 Hz}.
For these four ECD sequences, we use the first 30~s of data for accuracy evaluation, where the magnitude of translational motion is relatively small.
Two sequences from \cite{Kim21ral} are also considered: \indoor{} and \fastmotion{}. 
Both of them consist of $\approx$ 4.5 million events, with mocap GT poses given at \mbox{100 Hz}. 
In sequence \indoor{}, the camera rotates $360^\circ$ around the vertical axis, 
while in \fastmotion{}, the camera performs random fast rotations throughout the whole sequence.
Since the GT from the mocap(s) is noisy, it is filtered with a Gaussian low-pass filter in the Lie group sense using a time span of 50~ms.

\subsubsection{Hardware Configuration}
All experiments are conducted on a standard laptop (Intel Core i7-1165G7 CPU @ 2.80GHz), with the exception of those involving RTPT \cite{Reinbacher17iccp}, which runs on another laptop with an NVIDIA GeForce GTX 970M GPU due to software incompatibilities (running 2017 CUDA code).

\subsubsection{Downsampling}
Event downsampling is only enabled in the sensitivity study on the event sampling rate (\cref{sec:experiments:sensitive_analysis:event_sample_rate}),
and the nearly real-time demonstration shown in the supplementary video (\cref{sec:experiments:wild_experiments}).
In all other experiments of our methods, all input events are processed.

\subsection{Evaluation Metrics}
\label{sec:experiments:metrics}
To comprehensively characterize the performance of the methods, 
we evaluate both the output trajectories and obtained maps.
For the former, we calculate the absolute and relative errors of the estimated rotations with respect to the GT.
For the latter, we compute panoramic maps (i.e., IWEs) using the trajectories output by each approach,
and extend the reprojection error from feature-based SLAM, to assess the quality of trajectories and maps.

\subsubsection{Absolute Error}
The absolute rotation error quantifies the global consistency of the estimated camera poses \cite{Sturm12iros}. 
At timestamp $t_k$, the absolute rotation error $\Delta \Rot_k$ is given by
\begin{equation}
    \Delta \Rot_k = \Rot_k'^\top \Rot_k,
    \label{eq:abs_err}
\end{equation}
where $\Rot_k$ is the estimated rotation at $t_k$ and $\Rot_k'$ is the corresponding GT rotation (obtained by linear interpolation). 
In the benchmark, each method outputs rotations at a different rate, hence we calculate a series of absolute errors at these timestamps when the rotation is estimated, and further compute the Root Mean Square (RMS) to represent the accuracy. 
To evaluate the continuous trajectories refined by the proposed BA, we query poses and compute errors at \mbox{50 Hz}.

\subsubsection{Relative Error}
The relative rotation error measures the local accuracy of the pose estimates \cite{Sturm12iros}. 
At timestamp $t_k$, the relative rotation error $\delta \Rot_k$ is:
\begin{equation}
    \delta \Rot_k = \left( \Rot_k'^\top \Rot'_{k+\Delta} \right)^{-1} \left( \Rot_k^\top \Rot_{k+\Delta} \right),
    \label{eq:rel_err}
\end{equation}
where $\{ \Rot_k, \Rot_{k+\Delta} \}$ and $\{ \Rot'_k, \Rot'_{k+\Delta} \}$ are pairs of estimated and GT poses, respectively, which are selected based on some criterion $\Delta$. 
We set the pose pairs to have a time span of 1~s, 
measuring the error produced per second, and sample the trajectory to get these pose pairs every 0.1~s.

In the experiments, we use the angle of the difference between two rotations \cite{Barfoot15book} to measure estimation errors:
\begin{equation}
\label{eq:rot2angle}
    \angle \mB = \arccos\left( \frac{\operatorname{trace}(\mB) - 1}{2} \right),
\end{equation}
where $\mB$ is $\Delta \Rot_k$ or $\delta \Rot_k$, as adopted in \cite{Zhang18iros}.

In all experiments, we align the output trajectories to the GT at $t_0$ before evaluation:
$\hat{\Rot}_k = \Rot'(t_0) \, \Rot^\top(t_0) \, \Rot_k,$
where $\hat{\Rot}_k$ is the aligned pose, $\Rot'(t_0)$ is the GT pose at $t = t_0$.
For synthetic and real-world sequences, $t_0$ is set to 0.1~s and 1.0~s respectively.

\subsubsection{Reprojection Error}
For real-world data, where the camera motion is not purely rotational, 
problems arise to characterize rotational errors. 
As will be explained in \cref{sec:experiments:real:problem}, 
we propose assessing the performance of the methods by means of the reprojection error.
Without groundtruth data association, we will argue in favor of using the Event Area (EA) \cite{Gallego19cvpr} of the panoramic map as a proxy for the reprojection error.
We use the percentage of event area with respect to the total image area, as it is more intuitive to understand. 

We also report the Gradient Magnitude \cite{Gallego19cvpr} of the IWE:
\begin{equation}
\label{eq:gradmagIWE}
    \operatorname{GM}(I) = \Bigl(\frac{1}{N}\int_{\Omega} \Vert \nabla I(\bx) \Vert^2 d\bx\Bigr)^{\frac{1}{2}},
\end{equation}
where $N$ is the number of pixels 
and $\nabla I = (I_x,I_y)^\top$ is the spatial gradient of $I$ 
(e.g., calculated using Sobel's operator).

\subsection{Experiments on Synthetic Data}
\label{sec:experiments:synth}
\begin{table*}
\caption{\label{tab:synth_data}
\emph{Absolute} [$^\circ$] and \emph{Relative} [$^\circ/s$] RMSE on synthetic datasets.  
} 
\centering
\adjustbox{max width=\linewidth}{
\setlength{\tabcolsep}{4pt}
\begin{tabular}{llrrrrrrrrrrrr}
\toprule
& Sequence & \multicolumn{2}{c}{\text{playroom}}
         & \multicolumn{2}{c}{\text{bicycle}}
         & \multicolumn{2}{c}{\text{city}}
         & \multicolumn{2}{c}{\text{street}}
         & \multicolumn{2}{c}{\text{town}}
         & \multicolumn{2}{c}{\text{bay}} \\
\cmidrule(l{1mm}r{1mm}){3-4}
\cmidrule(l{1mm}r{1mm}){5-6}
\cmidrule(l{1mm}r{1mm}){7-8}
\cmidrule(l{1mm}r{1mm}){9-10}
\cmidrule(l{1mm}r{1mm}){11-12}
\cmidrule(l{1mm}r{1mm}){13-14}
& & Abs & Rel
& Abs & Rel
& Abs & Rel
& Abs & Rel
& Abs & Rel 
& Abs & Rel \\

\midrule
\multirow{5}{*}{\shortstack{Front-end \\ \cref{sec:experiments:synth:frontend}}}

& PF-\smt{} \cite{Kim14bmvc} & 9.038 & 6.457 & - & - & - & - & - & - & - & - & - & - \\

& EKF-\smt{} \cite{Kim18phd} & 6.059 & 4.270 & {1.382} & {0.935} & {1.715} & {1.376} & 3.481 & {1.706} & 4.331 & {1.599} & {2.649} & {1.908} \\
 
& \rtpt{} \cite{Reinbacher17iccp} & - & - & - & - & - & - & - & - & - & - & - & - \\
 
& \cmaxgae{} \cite{Gallego17ral} & 4.794 & 4.178 & 1.666 & 1.397 & - & - & - & - & 6.970 & 4.568 & - & - \\

& \cmaxw{} front-end \cite{Gallego17ral} & {3.228} & {2.641} & 1.731 & 1.576 & 5.322 & 3.988 & {1.917} & 1.712 & {1.910} & 1.887 & 3.696 & 3.078 \\

\midrule
\multirow{3}{*}{\shortstack{BA: linear \\ \cref{sec:experiments:synth:backend}}}

& PF-\smt{} \cite{Kim14bmvc} & 0.631 & 1.140 & N/A & N/A & N/A & N/A & N/A & N/A & N/A & N/A & N/A & N/A \\

& EKF-\smt{} \cite{Kim18phd} & 0.250 & 0.257 & 0.434 & 0.289 & 0.299 & 0.321 & 0.331 & 0.334 & 0.954 & 0.507 & 0.476 & 0.353 \\

& \cmaxgae{} \cite{Kim21ral} & 1.588 & 0.869 & 0.637 & 0.538 & N/A & N/A & N/A & N/A & 1.640 & 0.855 & N/A & N/A \\

\midrule
\multirow{3}{*}{\shortstack{BA: cubic \\ \cref{sec:experiments:synth:backend}}}

& PF-\smt{} \cite{Kim14bmvc} & 10.242 & 8.518 & N/A & N/A & N/A & N/A & N/A & N/A & N/A & N/A & N/A & N/A \\

& EKF-\smt{} \cite{Kim18phd} & 1.265 & 0.774 & 0.627 & 0.510 & 0.312 & 0.359 & 0.508 & 0.538 & 2.031 & 1.135 & 2.912 & 0.659 \\
 
& \cmaxgae{} \cite{Kim21ral} & 1.274 & 1.789 & 1.004 & 0.737 & N/A & N/A & N/A & N/A & 1.337 & 0.880 & N/A & N/A \\

\midrule
\multirow{2}{*}{\shortstack{System \\ \cref{sec:experiments:synth:system}}}

& \cmaxslam{} (linear) & 0.763 & 0.593 & 0.327 & 0.414 & 0.509 & 0.721 & 0.470 & 0.584 & 0.553 & 0.554 & 0.617 & 0.475 \\

& \cmaxslam{} (cubic) & 0.796 & 0.538 & 0.368 & 0.472 & 0.499 & 0.621 & 0.412 & 0.524 & 0.541 & 0.527 & 0.674 & 0.424 \\

\bottomrule
\end{tabular}
}
\begin{tablenotes}
\item 
``-'' means the method fails on that sequence, 
and ``N/A'' indicates that BA refinement is not applicable because the corresponding front-end failed on this sequence.
\end{tablenotes}

\end{table*}
\def\figWidth{0.158\linewidth}
\begin{figure*}[t]
	\centering
    {\small
    \setlength{\tabcolsep}{1pt}
	\begin{tabular}{
	>{\centering\arraybackslash}m{0.4cm} 
	>{\centering\arraybackslash}m{\figWidth} 
	>{\centering\arraybackslash}m{\figWidth}
	>{\centering\arraybackslash}m{\figWidth}
	>{\centering\arraybackslash}m{\figWidth}
        >{\centering\arraybackslash}m{\figWidth}
        >{\centering\arraybackslash}m{\figWidth}}

        \rotatebox{90}{\makecell{\bicycle{}}}
		&\gframe{\includegraphics[width=\linewidth]{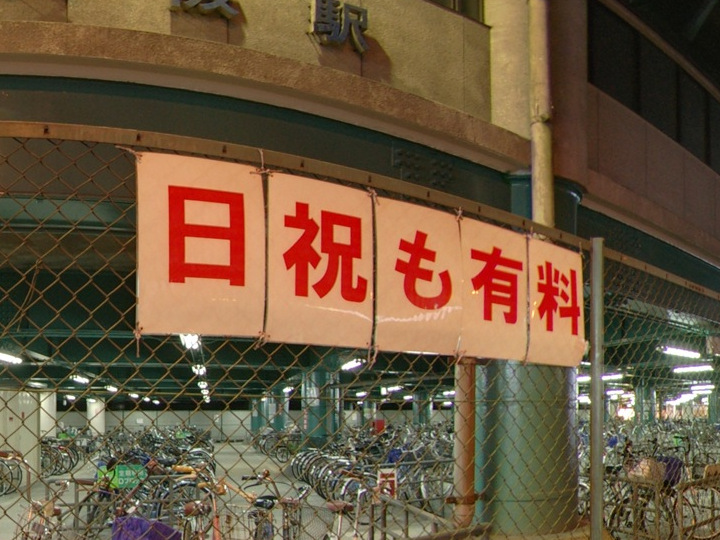}}
		&\gframe{\includegraphics[width=\linewidth]{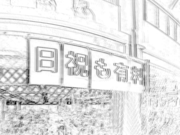}}
		&\gframe{\includegraphics[width=\linewidth]{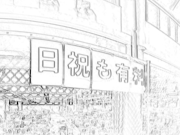}}
            &\gframe{\includegraphics[width=\linewidth]{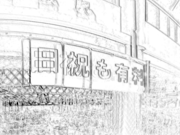}}
            &\gframe{\includegraphics[width=\linewidth]{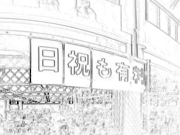}}
            &\gframe{\includegraphics[width=\linewidth]{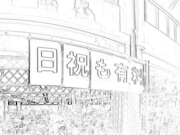}}
		\\
        
        \rotatebox{90}{\makecell{\town{}}}
		&\gframe{\includegraphics[width=\linewidth]{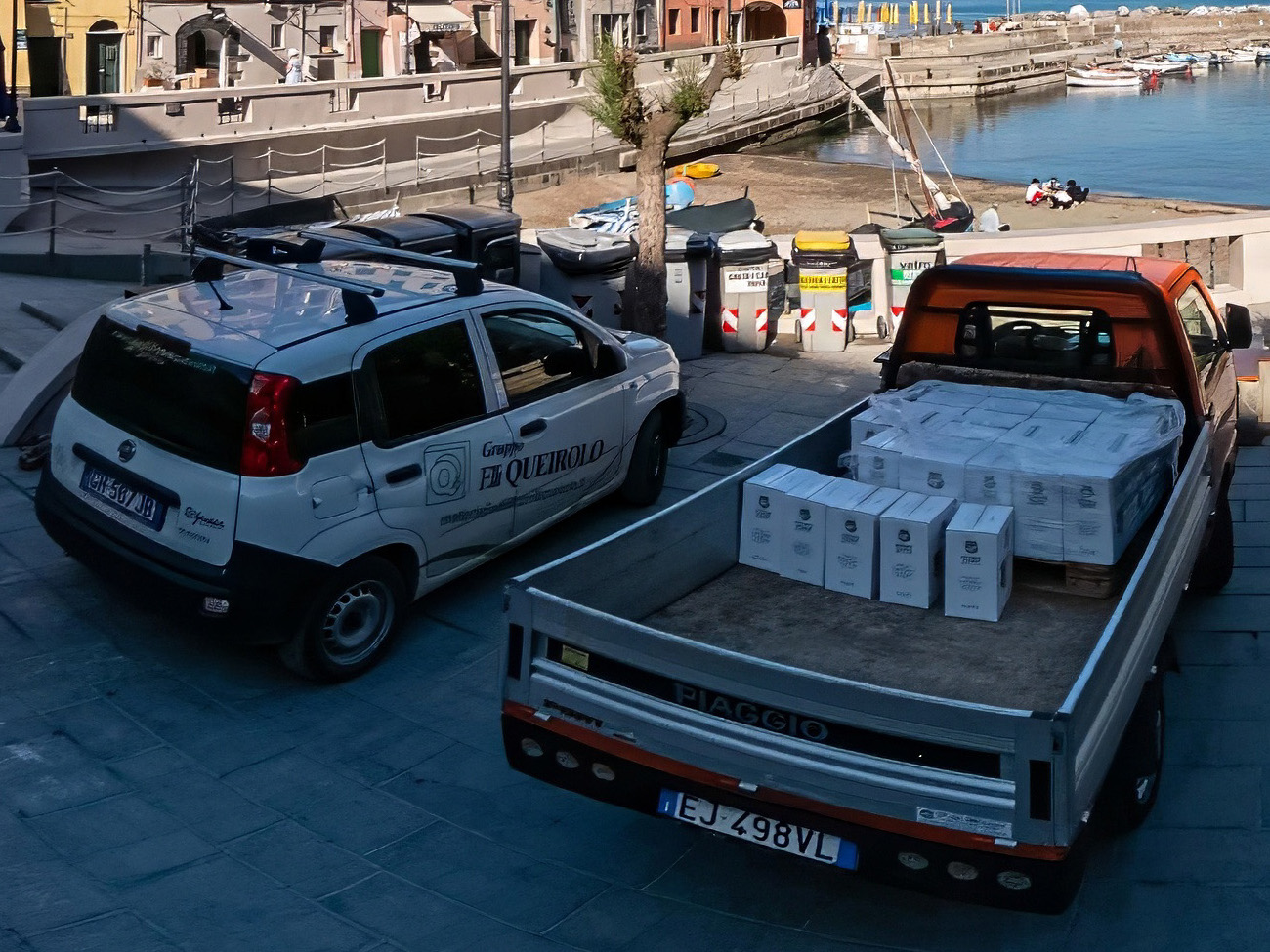}}
		&\gframe{\includegraphics[width=\linewidth]{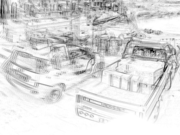}}
		&\gframe{\includegraphics[width=\linewidth]{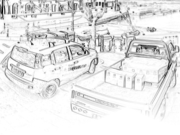}}
            &\gframe{\includegraphics[width=\linewidth]{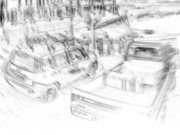}}
            &\gframe{\includegraphics[width=\linewidth]{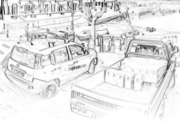}}
            &\gframe{\includegraphics[width=\linewidth]{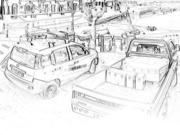}}
		\\

        & (a) Scenes
        & (b) EKF-\smt{}
        & (c) Refined EKF-\smt{} (linear)
        & (d) \cmaxgae{} trajectories
        & (e) Refined \cmaxgae{} (linear)
        & (f) Groundtruth\\
	\end{tabular}
	}
    \caption{\emph{Effect of Bundle Adjustment (Offline smoothing)}.
    Parts of the panoramic IWEs obtained with the estimated trajectories (before / after BA refinement) and GT.
    Synthetic data, as in \cref{tab:synth_data}.
    Gamma correction $\gamma = 0.75$ applied for better visualization. 
    \label{fig:synth_iwe}
    }
\end{figure*}

In this section, we compare the front-end methods, the proposed BA and \cmaxslam{} on the six synthetic sequences.
More details of the front-ends are given in Appendix~\ref{appendix:baselines}. 

\subsubsection{Comparison of Front-ends}
\label{sec:experiments:synth:frontend}

First, we benchmark all front-ends with synthetic data.
The top part of \cref{tab:synth_data} reports the accuracy of the corresponding camera trajectories.

\emph{\smt{}}:
Overall, EKF-\smt{} achieves the best performance among all front-end methods, 
while PF-\smt{} only completes tracking for \playroom{}.
Experimentally, EKF-\smt{} is more stable than PF-\smt{} and also less sensitive to the map quality.

\emph{\rtpt{}}:
Due to the probabilistic operation of \rtpt{}, it may output different results in different trials.
Hence, we report the best results we obtained. 
\rtpt{} fails on all synthetic sequences because of its limitation on the range of camera motions that can be tracked.
It is due to the fact that \rtpt{} monitors the tracking quality during operation and stops the map update when the quality decreases below a threshold, which often happens if the camera's FOV gets close to the left or right boundary of the panoramic map.

\emph{CMax}:
\cmaxw{} achieves the best absolute errors in three out of six sequences, which demonstrates the effectiveness of the event slicing strategy (\cref{sec:method:pipeline:frontend}).
\cmaxgae{} fails \city{}, \street{} and \bay{}, which have highly textured areas to trigger a massive amount of events.
On these sequences, events are densely clustered on the map of \cmaxgae{} (\cref{fig:map_compare:CMAX-GAE}), which presents an obstacle for its optimizer to align events.

\subsubsection{Back-end (BA)}
\label{sec:experiments:synth:backend}

Next, we use the proposed BA method to smooth the trajectories estimated by the front-end methods that do not lose track.
The refinement is conducted offline, in a sliding window manner (as described in \cref{sec:method:pipeline:backend}).

As reported in the middle part of \cref{tab:synth_data}, the proposed BA is able to decrease both the absolute and relative RMSE of all front-ends that do not lose track.
For example, for \bicycle{}, the absolute RMSE of EKF-\smt{} is reduced from 1.382$^\circ$ (front-end) to 0.434$^\circ$ (BA: linear) or 0.627$^\circ$ (BA: cubic). 
The trajectories estimated by \cmaxgae{} are also considerably refined by our BA method.
The improvement in the refinement of the trajectories is also noticeable in the quality of the map, as shown in \cref{fig:synth_iwe}; the IWEs look sharper after refinement.

\subsubsection{CMax-SLAM}
\label{sec:experiments:synth:system}

\begin{table*}
\caption{\label{tab:real_data}
\emph{Absolute} [$^\circ$] and \emph{Relative} [$^\circ/s$] RMSE on real-world datasets.  
} 
\centering
\adjustbox{max width=\linewidth}{
\setlength{\tabcolsep}{4pt}
\begin{tabular}{llrrrrrrrrrrrr} 
\toprule
& Sequence & \multicolumn{2}{c}{\text{shapes} \cite{Mueggler17ijrr}}
         & \multicolumn{2}{c}{\text{poster} \cite{Mueggler17ijrr}}
         & \multicolumn{2}{c}{\text{boxes} \cite{Mueggler17ijrr}}
         & \multicolumn{2}{c}{\text{dynamic} \cite{Mueggler17ijrr}}
         & \multicolumn{2}{c}{\text{360 indoor} \cite{Kim21ral}}
         & \multicolumn{2}{c}{\text{fast motion} \cite{Kim21ral}}\\
\cmidrule(l{1mm}r{1mm}){3-4}
\cmidrule(l{1mm}r{1mm}){5-6}
\cmidrule(l{1mm}r{1mm}){7-8}
\cmidrule(l{1mm}r{1mm}){9-10}
\cmidrule(l{1mm}r{1mm}){11-12}
\cmidrule(l{1mm}r{1mm}){13-14}
& & Abs & Rel
& Abs & Rel
& Abs & Rel
& Abs & Rel
& Abs & Rel
& Abs & Rel \\

\midrule
\multirow{6}{*}{\shortstack{Front-end \\ \cref{sec:experiments:real:frontend}}}

& PF-\smt{} \cite{Kim14bmvc} & - & - & - & - & - & - & - & - & 10.562 & 12.041 & - & - \\

& EKF-\smt{} \cite{Kim18phd} & - & - & - & - & - & - & - & - & 11.317 & 7.596 & {2.358} & {1.656} \\

& \rtpt{} \cite{Reinbacher17iccp} & {3.224} & 3.167 & 8.135 & 9.423 & {2.973} & {2.375} & {1.873} & 1.991 & - & - & 10.566 & 14.111 \\
 
& \cmaxgae{} \cite{Kim21ral} & 3.961 & {2.777} & {4.010} & {3.612} & - & - & 4.840 & {1.755} & {5.145} & {1.574} & 4.010 & 3.850 \\

& \cmaxw{} front-end \cite{Gallego17ral} & 7.164 & 6.536 & 7.263 & 5.985 & 9.241 & 6.846 & 5.535 & 3.696 & 7.889 & 1.609 & 3.519 & 2.485 \\

& IMU dead reckoning & 46.803 & 6.470 & 45.743 & 5.746 & 46.979 & 6.911 & 44.531 & 4.183 & 63.396 & 9.156 & 15.236 & 10.035 \\

\midrule
\multirow{2}{*}{\shortstack{BA \\ \cref{sec:experiments:real:backend}}}

& IMU (linear) & 4.939 & 7.159 & 5.624 & 6.658 & 5.423 & 7.063 & 3.289 & 3.643 & 4.579 & 1.370 & 1.695 & 2.844 \\

& IMU (cubic) & 4.934 & 6.791 & 5.716 & 6.803 & 5.393 & 6.963 & 3.337 & 3.645 & 4.990 & 1.380 & 2.053 & 3.053 \\

\midrule
\multirow{2}{*}{\shortstack{System \\ \cref{sec:experiments:real:system}}}

& \cmaxslam{} (linear) & 4.953 & 6.712 & 5.653 & 6.357 & 5.418 & 6.751 & 3.380 & 3.586 & 5.046 & 1.564 & 1.322 & 2.182 \\

& \cmaxslam{} (cubic) & 4.974 & 6.821 & 5.674 & 6.497 & 5.391 & 6.761 & 3.374 & 3.618 & 5.002 & 1.578 & 1.320 & 2.038 \\

\bottomrule
\end{tabular}
}
\begin{tablenotes}
\item \qquad 
``-'' means the method fails on that sequence.
\end{tablenotes}
\end{table*}
\begin{figure*}[t]
     \centering
     \begin{subfigure}[b]{0.75\linewidth}
         \centering
         \includegraphics[width=\linewidth]{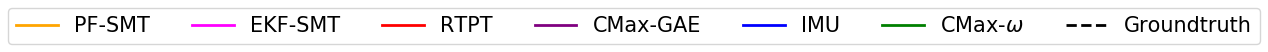}
     \end{subfigure}
     
    \begin{subfigure}[b]{0.245\linewidth}
         \centering
         \includegraphics[width=\linewidth]{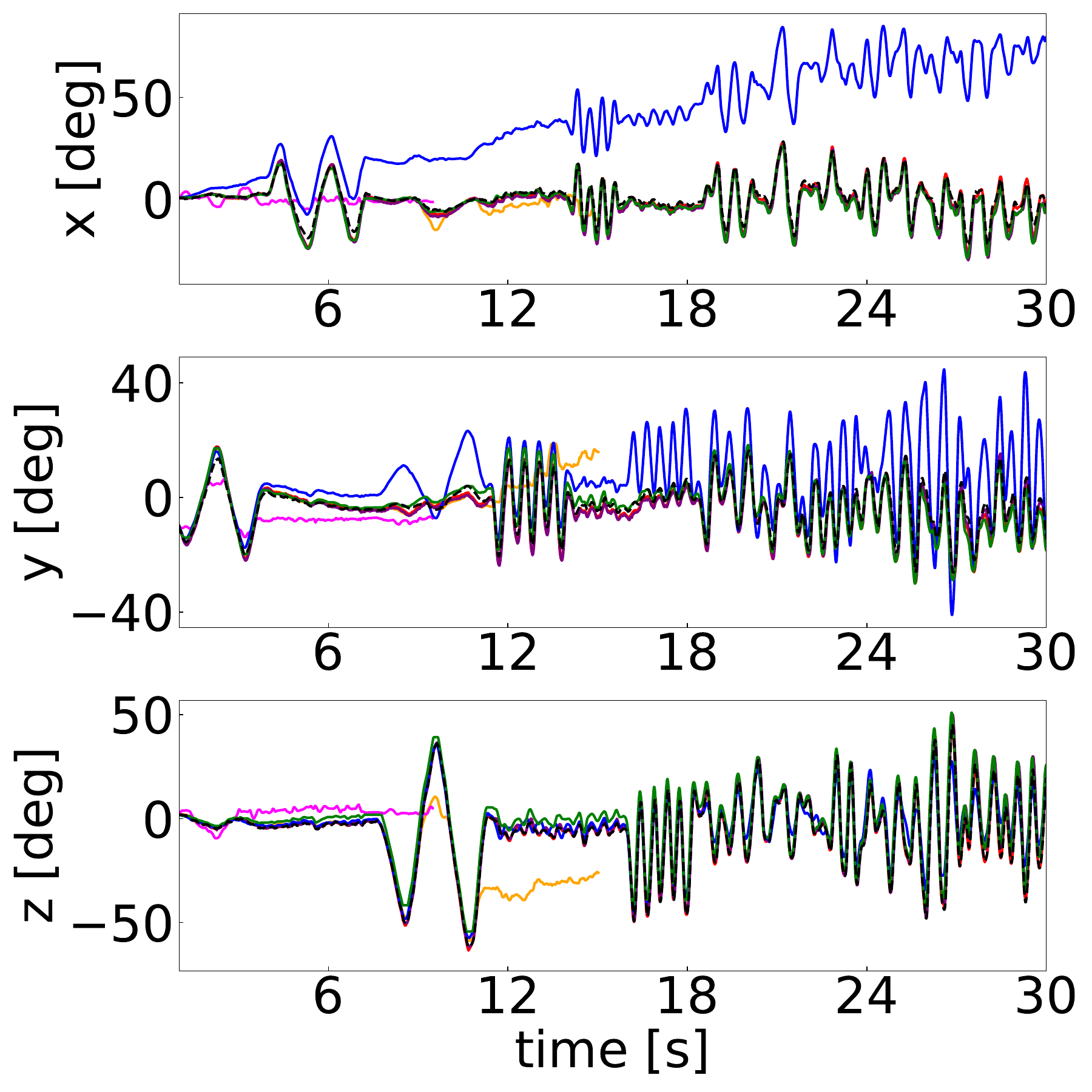}
         \caption{shapes}
         \label{fig:front-end_traj:fast_motion}
     \end{subfigure}
     \hspace{-1ex}
     \begin{subfigure}[b]{0.245\linewidth}
         \centering
         \includegraphics[width=\linewidth]{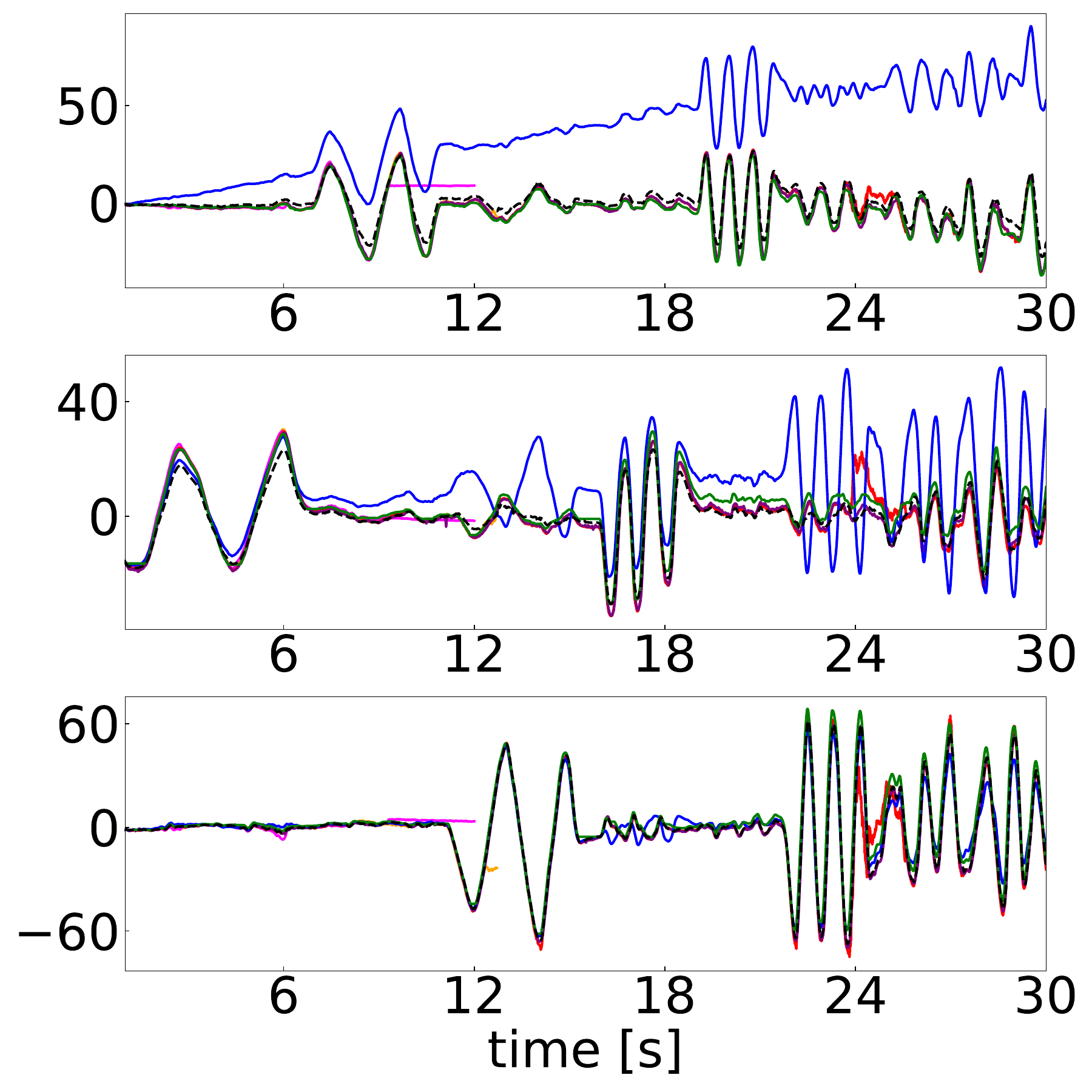}
         \caption{poster}
         \label{fig:front-end_traj:poster}
     \end{subfigure}
     \hspace{-2ex}
     \begin{subfigure}[b]{0.245\linewidth}
         \centering
         \includegraphics[width=\linewidth]{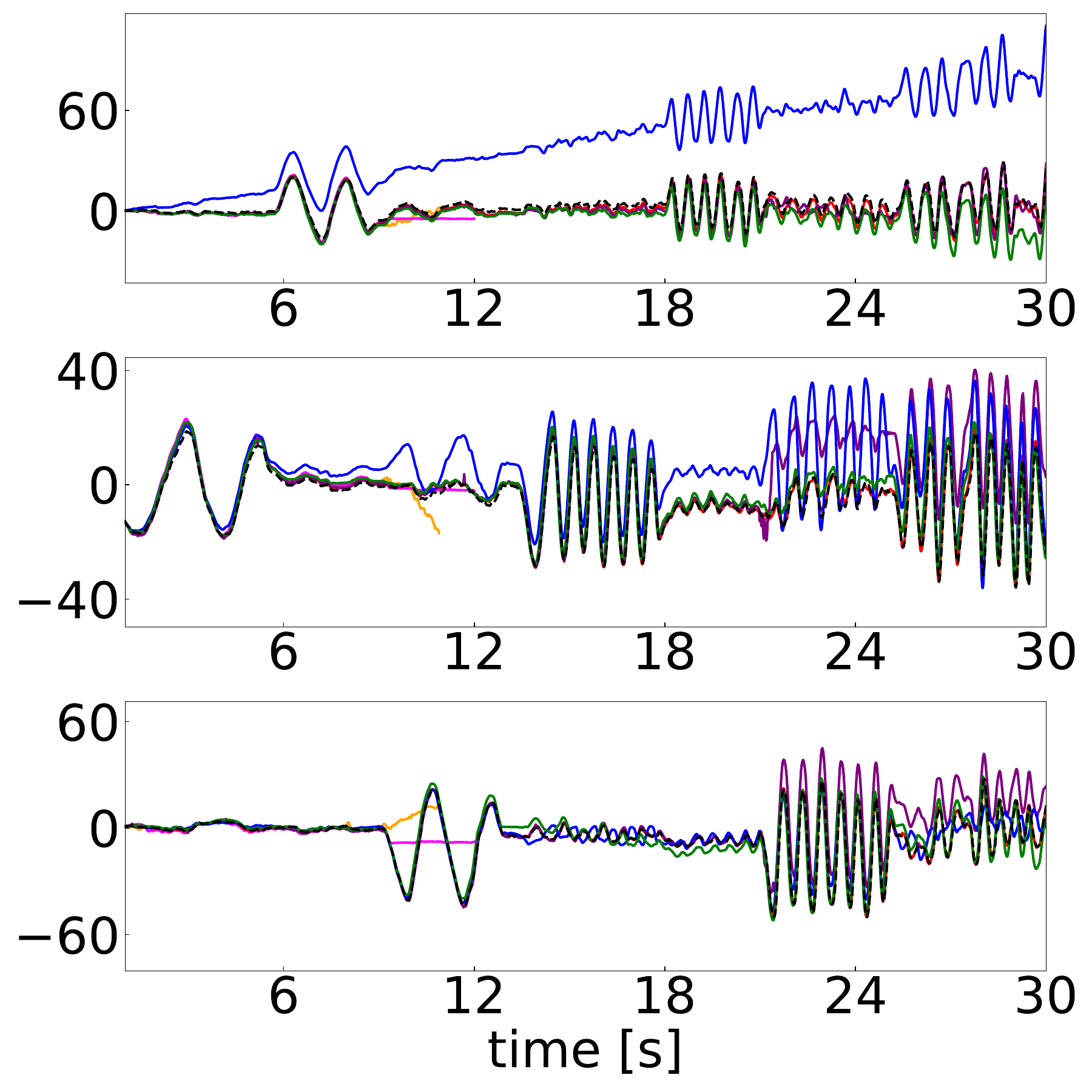}
         \caption{boxes}
         \label{fig:front-end_traj:boxes}
     \end{subfigure}
     \hspace{-2ex}
     \begin{subfigure}[b]{0.245\linewidth}
         \centering
         \includegraphics[width=\linewidth]{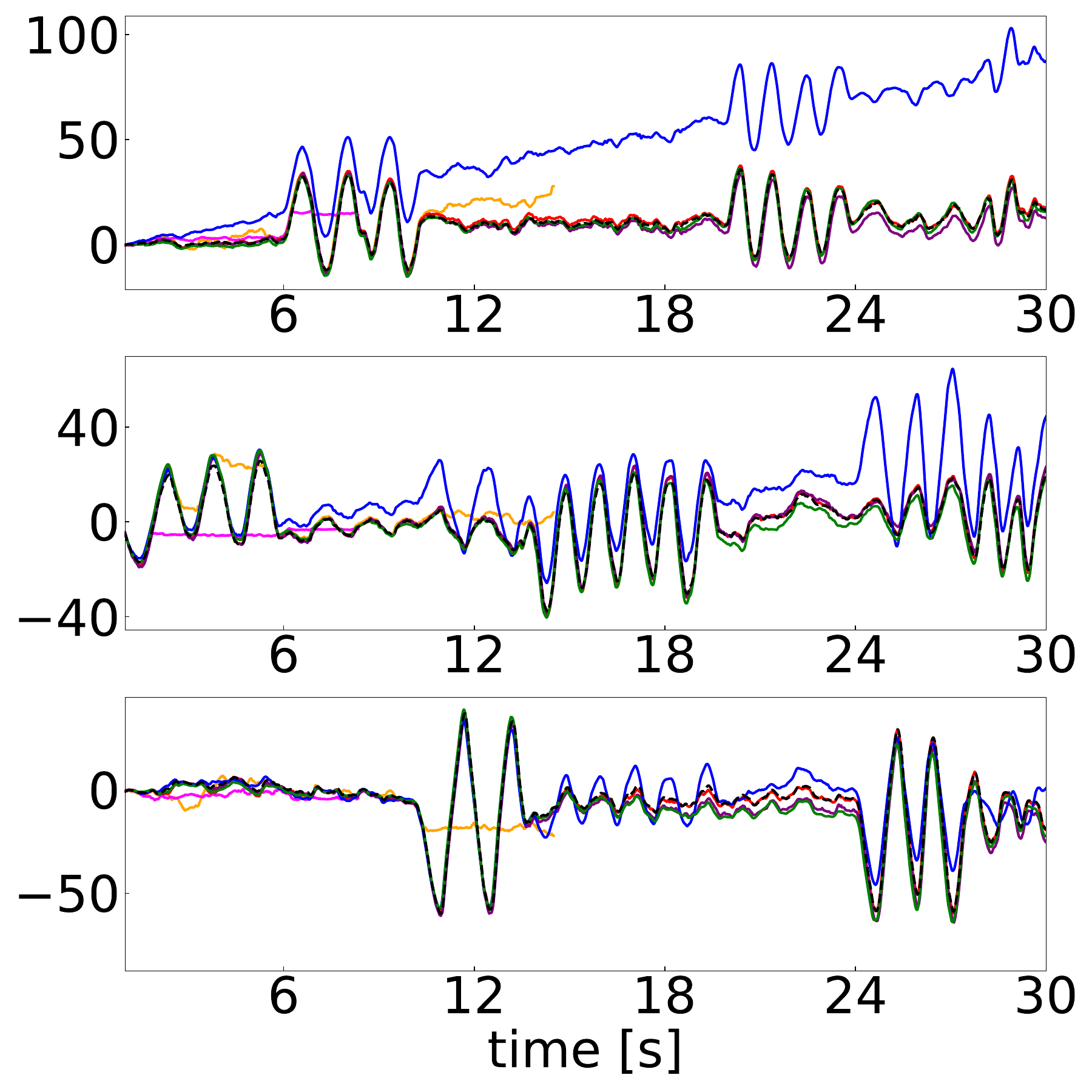}
         \caption{dynamic}
         \label{fig:front-end_traj:dynamic}
    \end{subfigure}
        \caption{\emph{Front-ends (before BA)}. 
        Comparison of camera trajectories of all front-end methods involved in the benchmark.
        EKF-\smt{} and PF-\smt{} do not show up because they fail all sequences from ECD \cite{Mueggler17ijrr} dataset.
        \label{fig:front-end_traj}}
\vspace{1ex}
    {
     \begin{subfigure}[b]{0.50\linewidth}
         \centering
         \includegraphics[width=\linewidth]{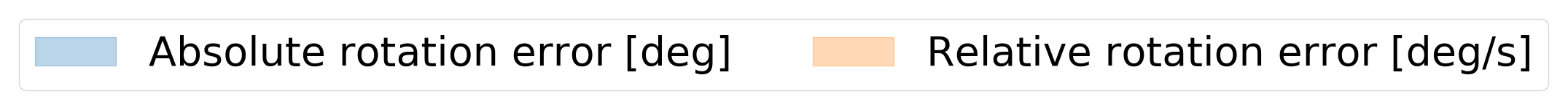}
     \end{subfigure}\\
     \begin{subfigure}[b]{0.245\linewidth}
         \centering
         \includegraphics[width=\linewidth, trim={1.0cm 1.0cm 1.0cm 0.5cm}]{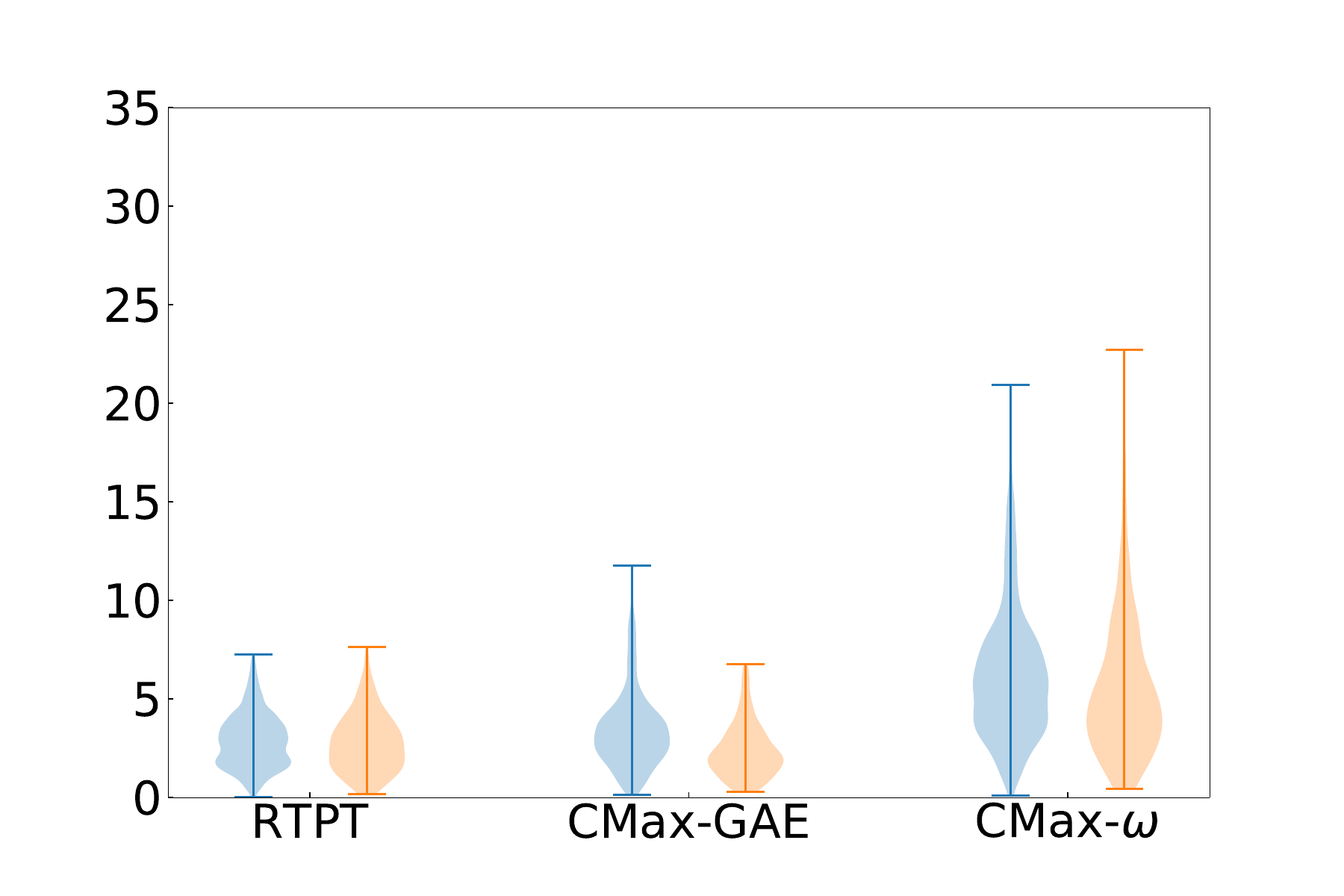}
         \caption{shapes}
         \label{fig:front-end_err:shapes}
     \end{subfigure}
    \hspace{-2ex}
     \begin{subfigure}[b]{0.245\linewidth}
         \centering
         \includegraphics[width=\linewidth, trim={1.0cm 1.0cm 1.0cm 0.5cm}]{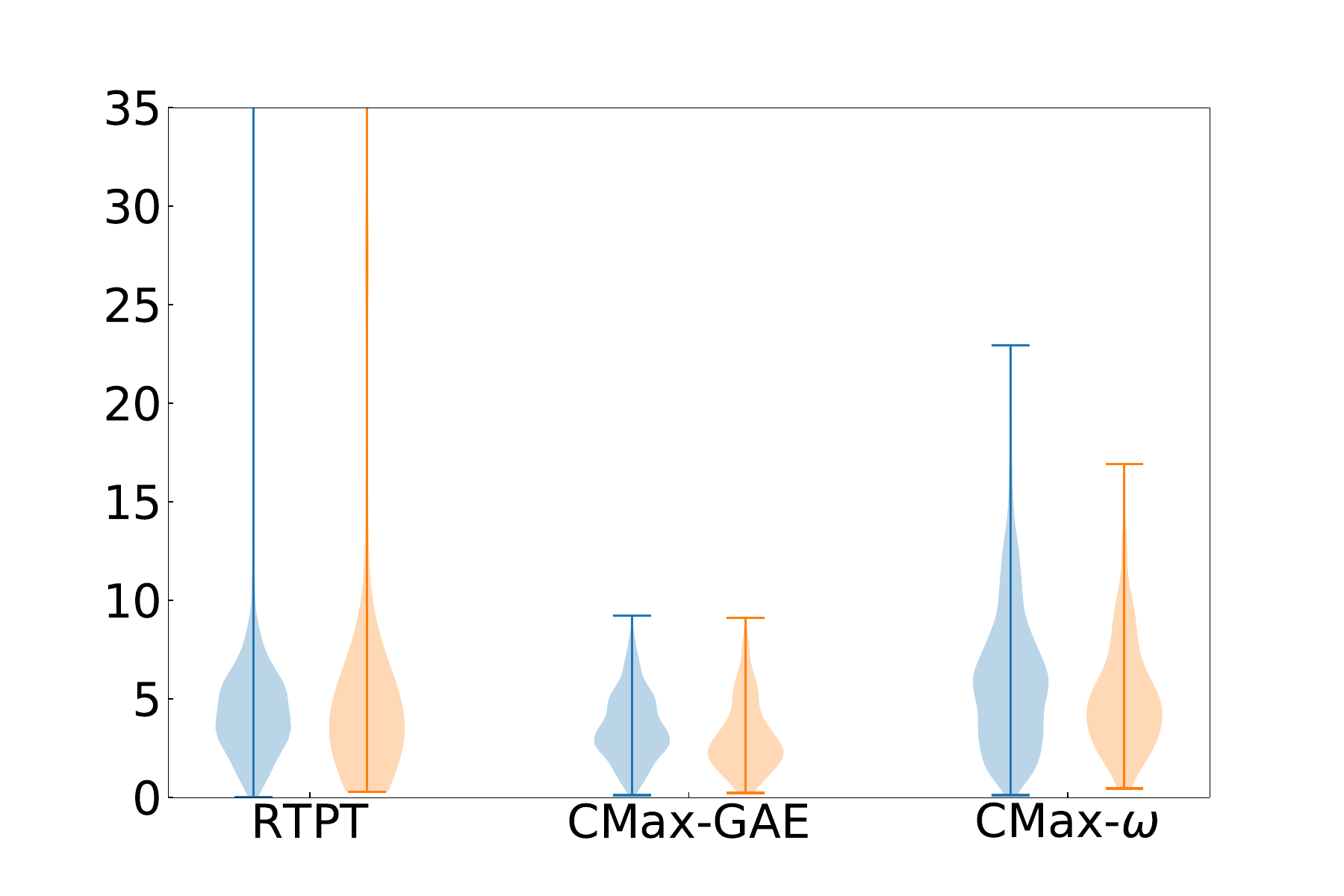}
         \caption{poster}
         \label{fig:front-end_err:poster}
     \end{subfigure}
    \hspace{-2ex}
     \begin{subfigure}[b]{0.245\linewidth}
         \centering
         \includegraphics[width=\linewidth, trim={1.0cm 1.0cm 1.0cm 0.5cm}]{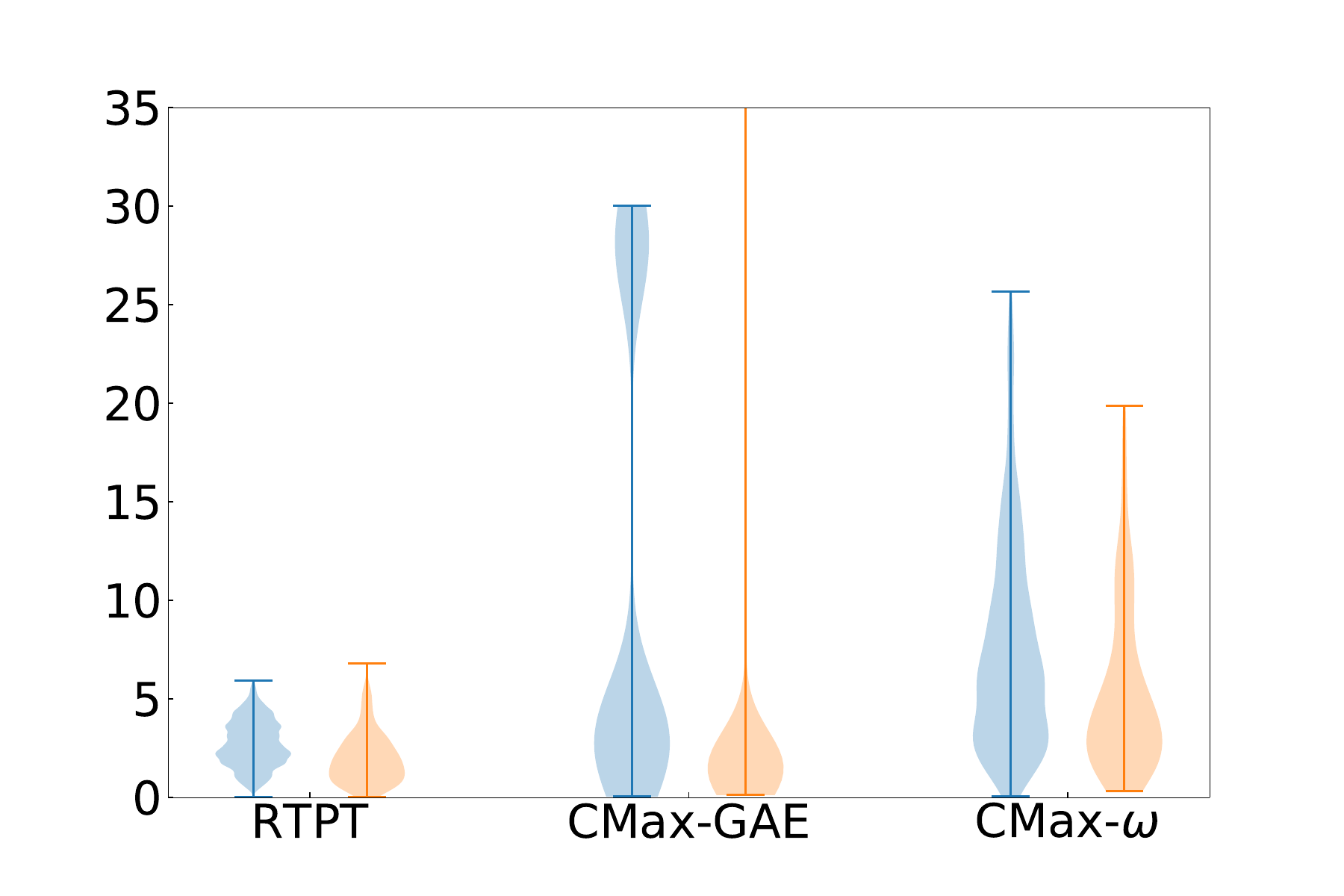}
         \caption{boxes}
         \label{fig:front-end_err:boxes}
     \end{subfigure}
    \hspace{-2ex}
     \begin{subfigure}[b]{0.245\linewidth}
         \centering
         \includegraphics[width=\linewidth, trim={1.0cm 1.0cm 1.0cm 0.5cm}]{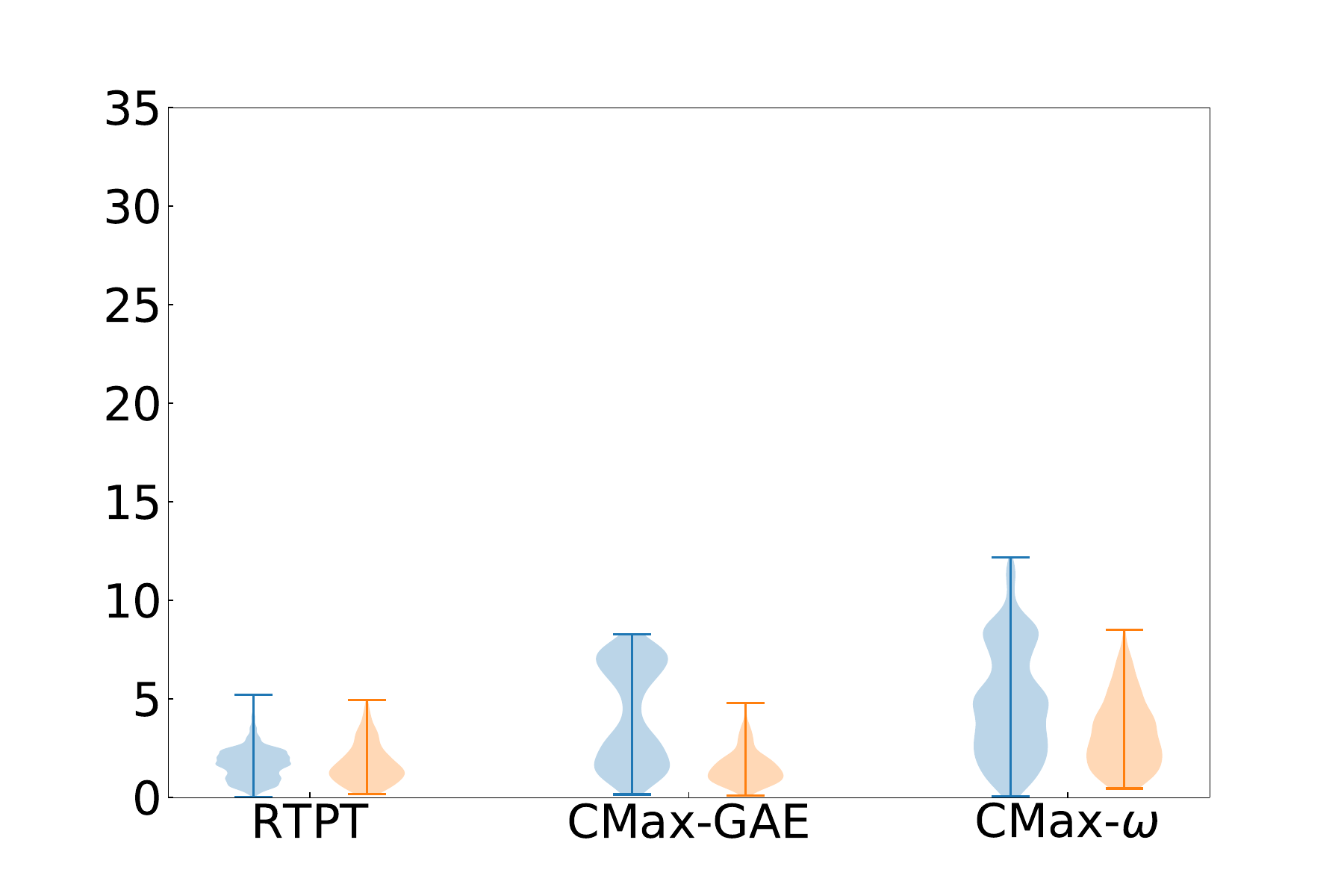}
         \caption{dynamic}
         \label{fig:front-end_err:dynamic}
     \end{subfigure}
     }
    \addtocounter{figure}{-1}
    \captionof{figure}{\emph{Front-ends (before BA)}.
    Absolute and relative errors of the visual front-end methods. 
    \smt{} is not included since it does not work on most sequences.
    \label{fig:front-end_err}}
\end{figure*}

Finally, the bottom part of \cref{tab:synth_data} reports the accuracy numbers of the proposed \cmaxslam{} system on the same sequences.
\cmaxslam{} outperforms all baseline methods on all synthetic sequences. 
Additionally, there is an obvious improvement from \cmaxw{} to \cmaxslam{} in all trials.
This further supports the effectiveness of the proposed BA back-end, which works in an online manner here.
The linear and cubic spline trajectories report similar accuracy.
In short, \cmaxslam{} is more accurate than all previously existing rotation estimation methods.

\subsection{Experiments on Real-world Data}
\label{sec:experiments:real}
This section compares the front-end methods, the proposed BA and \cmaxslam{} on real-world data.

\begin{figure*}
     \centering
     \begin{subfigure}[b]{0.7\linewidth}
         \centering
         \includegraphics[width=\linewidth]{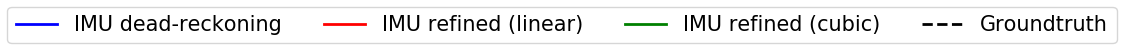}
     \end{subfigure}
     
    \begin{subfigure}[b]{0.245\linewidth}
         \centering
         \includegraphics[width=\linewidth]{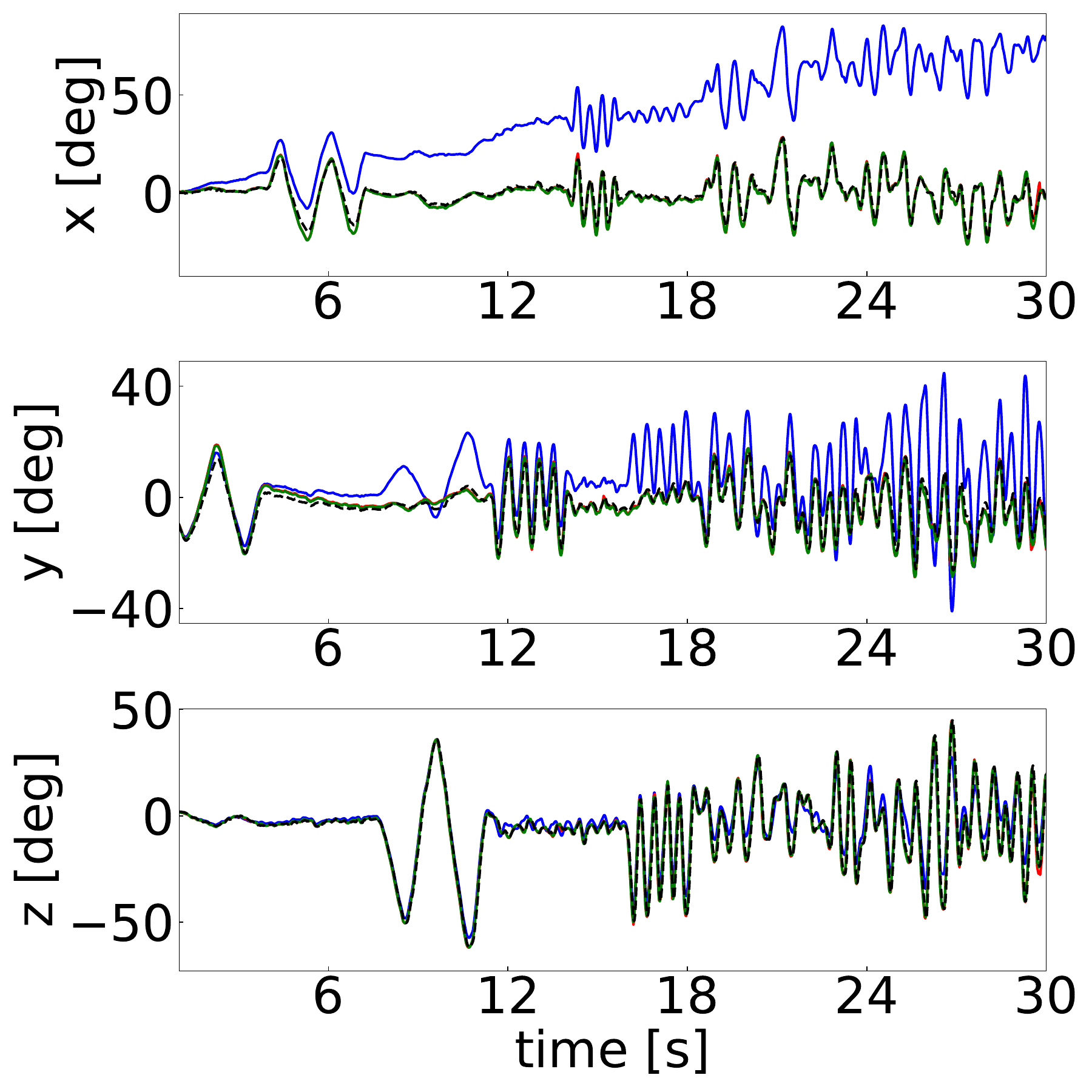}
         \caption{shapes}
         \label{fig:refined_traj:shapes}
     \end{subfigure}
     \hspace{-1ex}
     \begin{subfigure}[b]{0.245\linewidth}
         \centering
         \includegraphics[width=\linewidth]{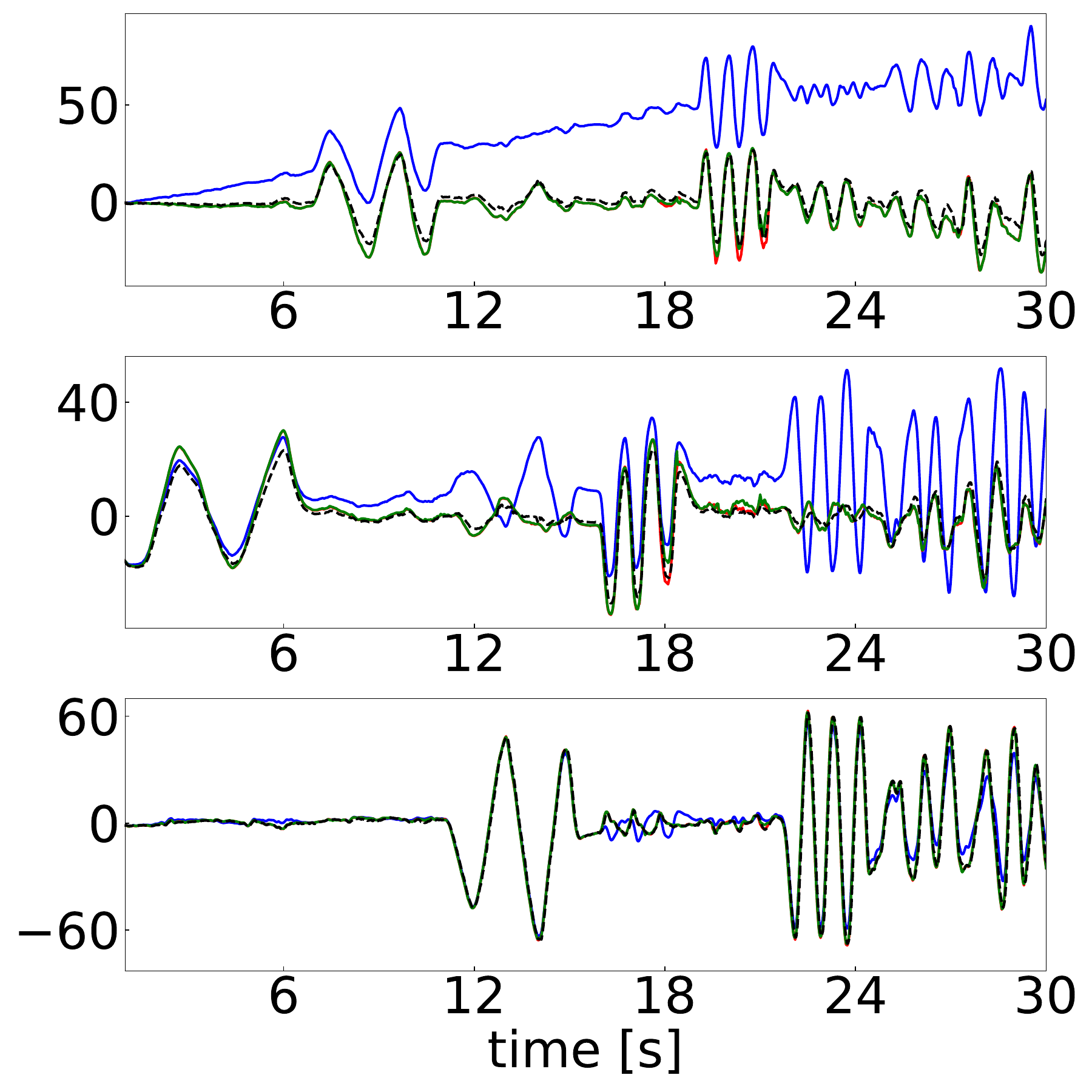}
         \caption{poster}
         \label{fig:refined_traj:poster}
     \end{subfigure}
     \hspace{-2ex}
     \begin{subfigure}[b]{0.245\linewidth}
         \centering
         \includegraphics[width=\linewidth]{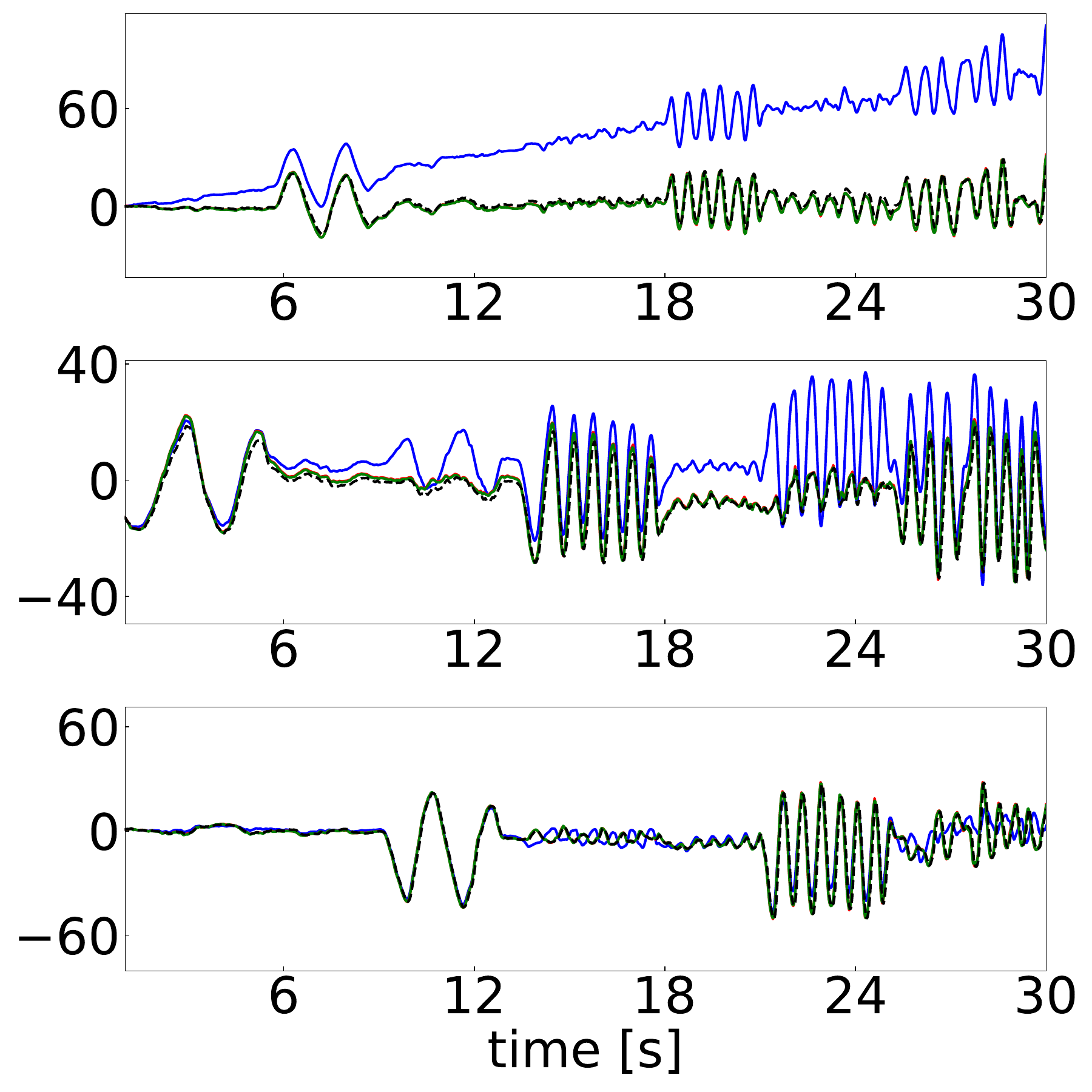}
         \caption{boxes}
         \label{fig:refined_traj:boxes}
     \end{subfigure}
     \hspace{-2ex}
     \begin{subfigure}[b]{0.245\linewidth}
         \centering
         \includegraphics[width=\linewidth]{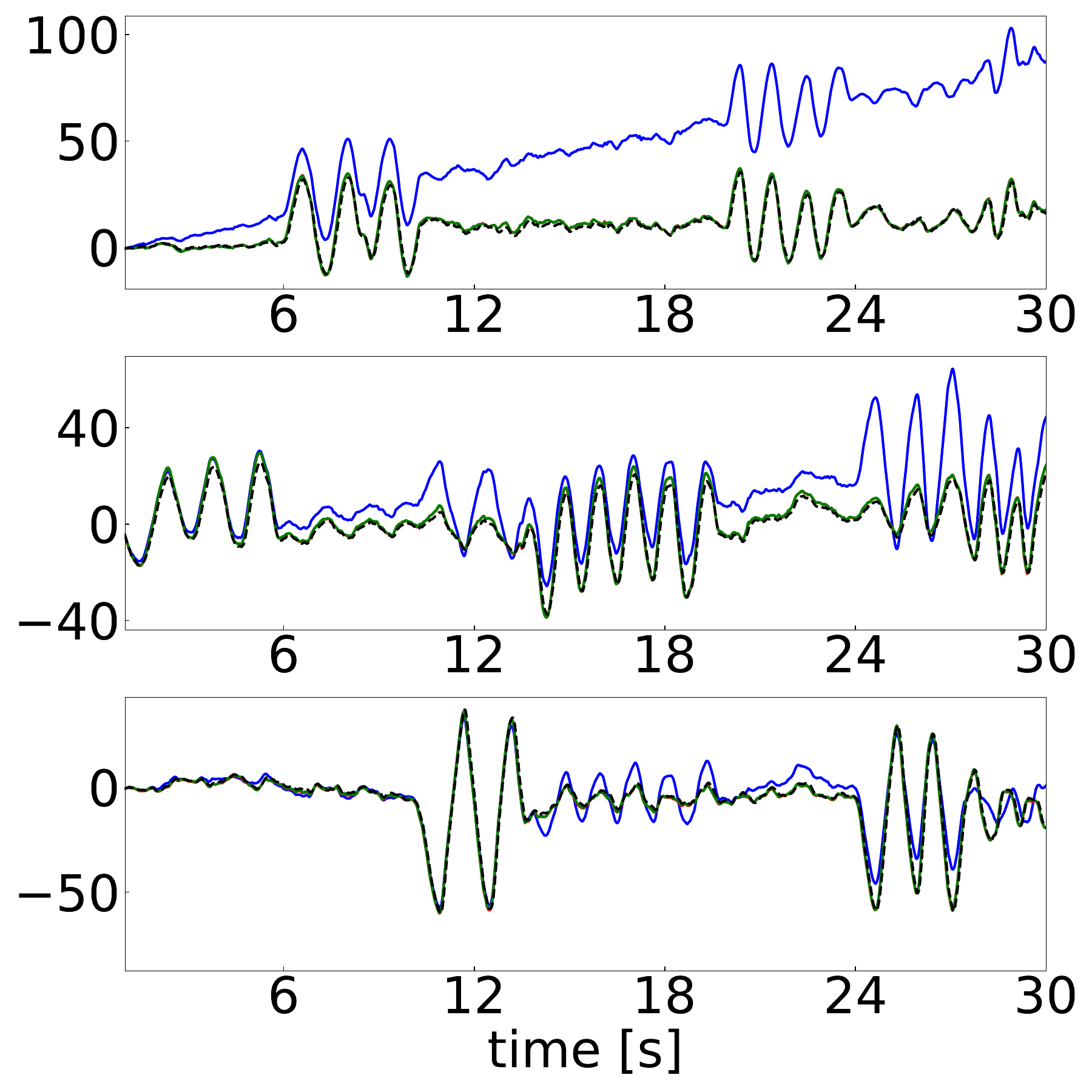}
         \caption{dynamic}
         \label{fig:refined_traj:dynamic}
    \end{subfigure}
        \caption{\emph{Camera trajectories}: 
        IMU dead reckoning and its refined trajectories (linear and cubic) using Bundle Adjustment.
        \label{fig:refined_traj}}
\vspace{1ex}
     \begin{subfigure}[b]{0.50\linewidth}
         \centering
         \includegraphics[width=\linewidth]{images/error/legend/err_violinplot_legend.png}
     \end{subfigure}
     
     \begin{subfigure}[b]{0.245\linewidth}
         \centering
         \includegraphics[width=\linewidth, trim={1.0cm 1.0cm 1.0cm 0.5cm}]{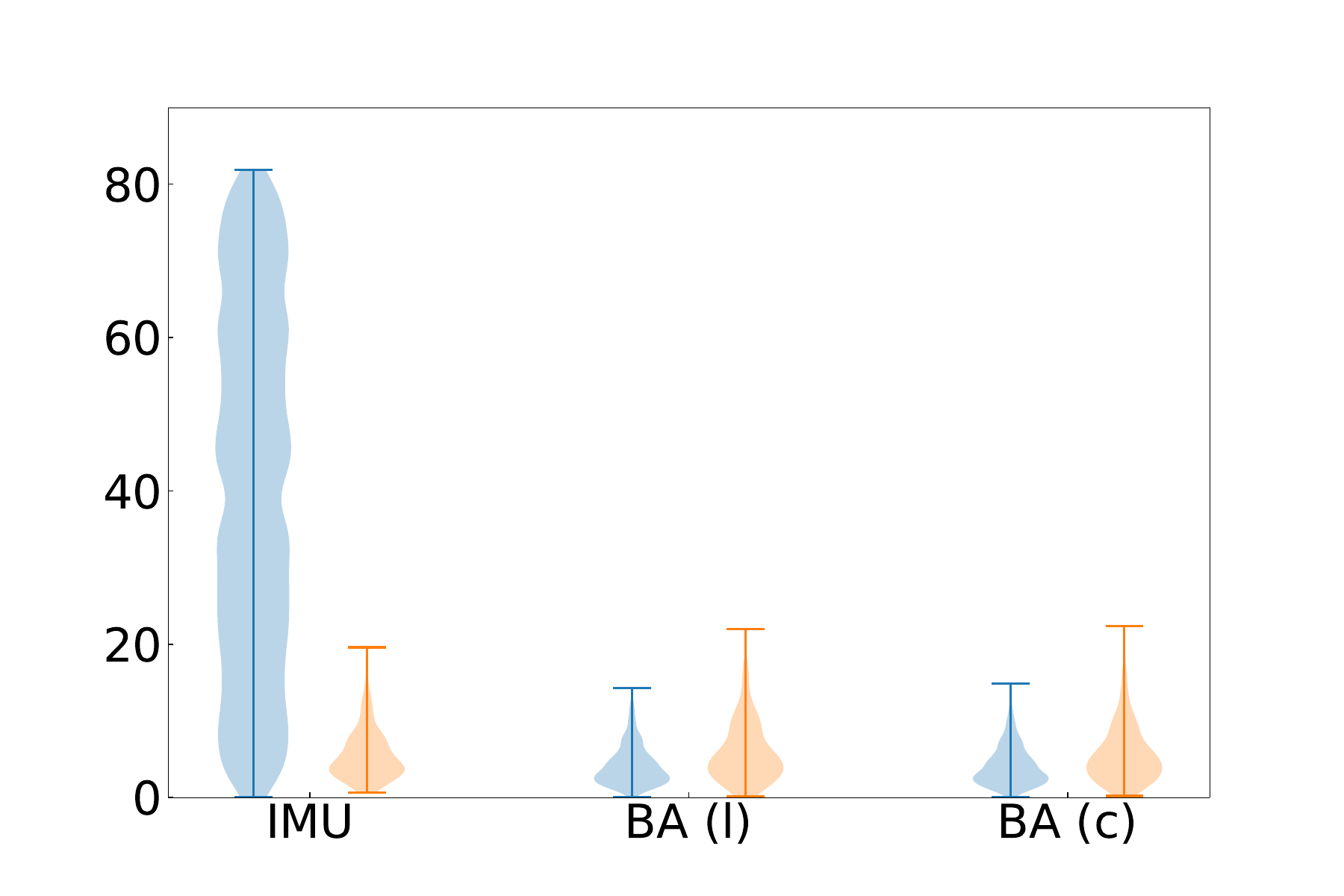}
         \caption{shapes}
         \label{fig:refined_traj_err:fast_motion}
     \end{subfigure}
    \hspace{-2ex}
     \begin{subfigure}[b]{0.245\linewidth}
         \centering
         \includegraphics[width=\linewidth, trim={1.0cm 1.0cm 1.0cm 0.5cm}]{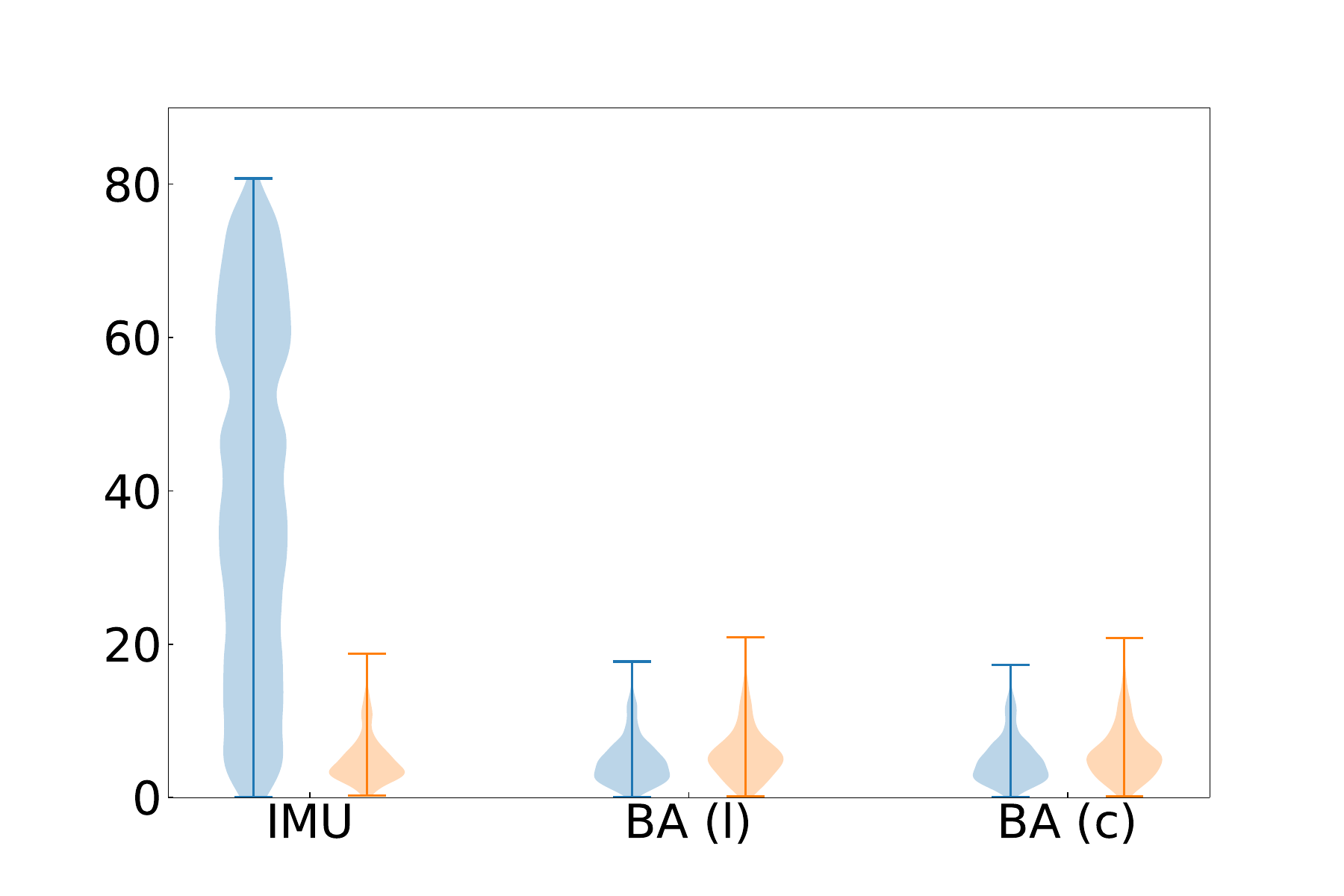}
         \caption{poster}
         \label{fig:refined_traj_err:poster}
     \end{subfigure}
    \hspace{-2ex}
     \begin{subfigure}[b]{0.245\linewidth}
         \centering
         \includegraphics[width=\linewidth, trim={1.0cm 1.0cm 1.0cm 0.5cm}]{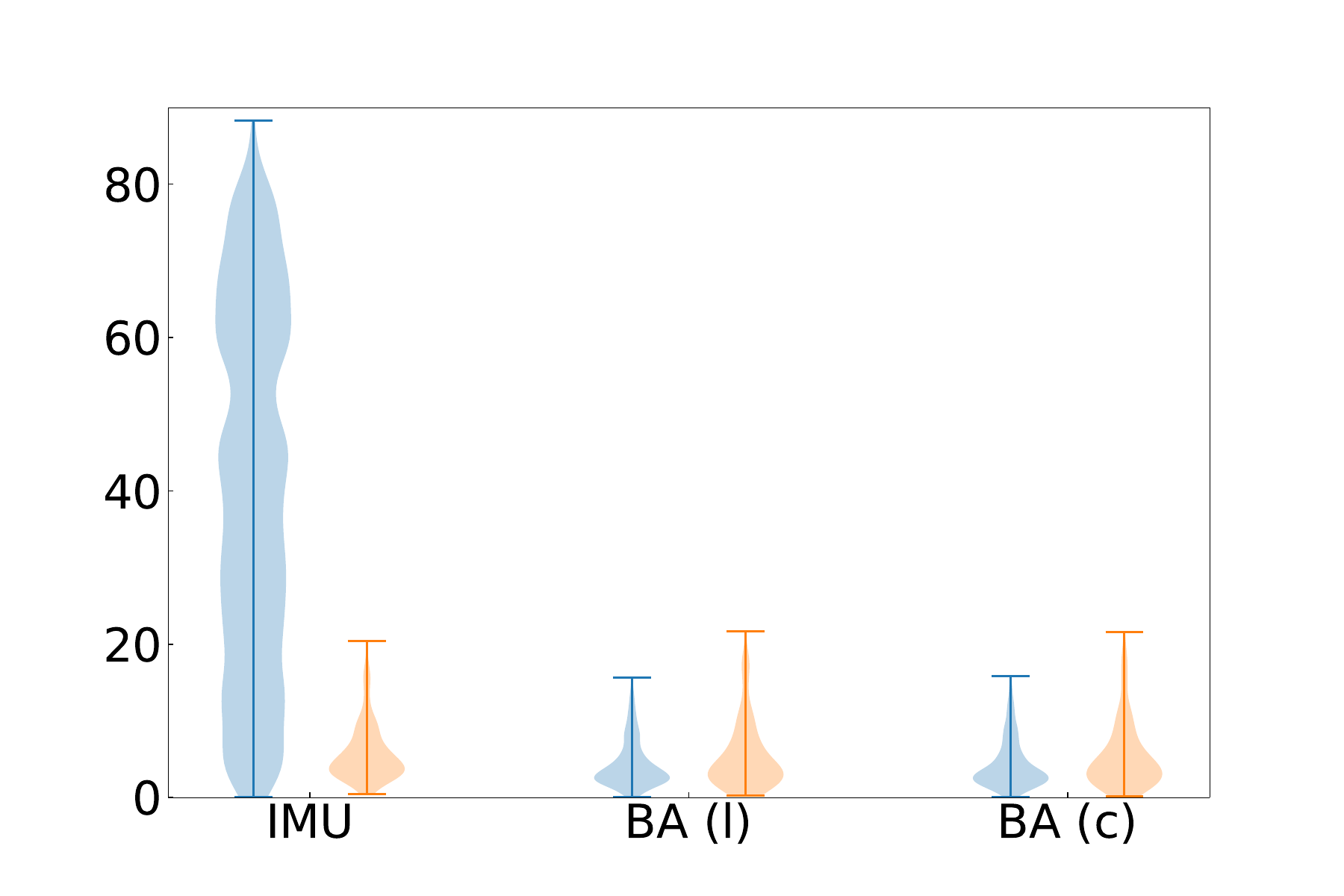}
         \caption{boxes}
         \label{fig:refined_traj_err:boxes}
     \end{subfigure}
    \hspace{-2ex}
     \begin{subfigure}[b]{0.245\linewidth}
         \centering
         \includegraphics[width=\linewidth, trim={1.0cm 1.0cm 1.0cm 0.5cm}]{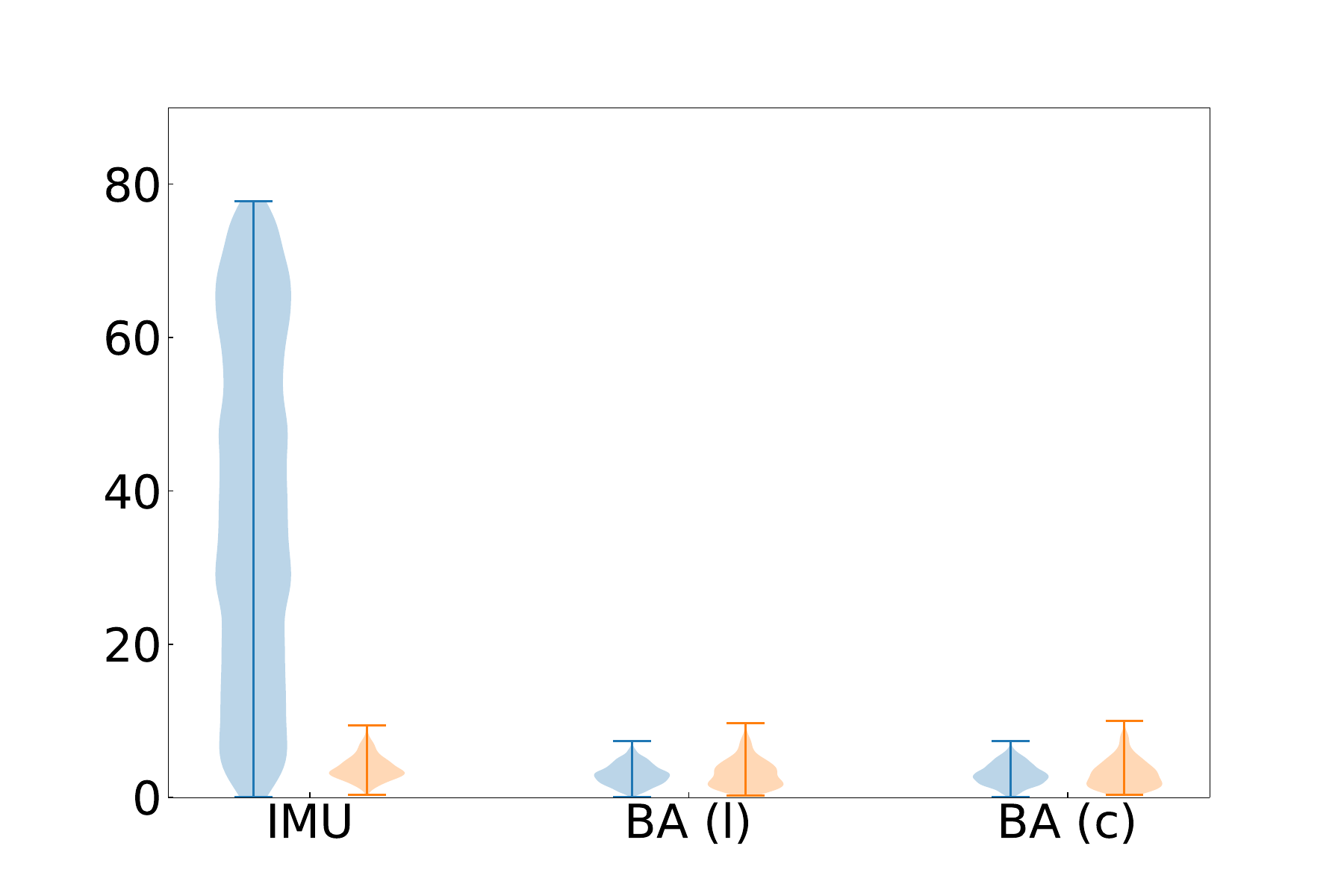}
         \caption{dynamic}
         \label{fig:refined_traj_err:dynamic}
     \end{subfigure}
    \addtocounter{figure}{-1}
    \captionof{figure}{
    \emph{Refinement of IMU dead reckoning using Bundle Adjustment (Offline smoothing)}.
    Absolute and relative errors of the input and refined trajectories.
    ``l'' and ``c'' indicate the BA algorithm with linear and cubic splines, respectively.
    \label{fig:refined_traj_err}}
\end{figure*}

\subsubsection{Evaluation Issues on Real-world Data}
\label{sec:experiments:real:problem}
The main difficulty of real-world evaluation lies in finding real data that is compatible with the purely rotational motion assumption of the problem. 
Real-world sequences from established datasets \cite{Mueggler17ijrr,Kim21ral} were recorded hand-held, hence they contain some translational motion.
Meanwhile, all methods impose the rotational motion constraint.
Hence, problems arise when testing 3-DOF methods on non-strictly 3-DOF data.

In the input events, one cannot disentangle the rotational part from the translational part. 
Hence, by design, all methods explain the additional DOFs in the events using only rotational DOFs.
If the translational motion is non-negligible, comparing the rotations that explain additional DOFs to the rotational component of the GT provided by a 6-DOF motion-capture system \cite{Mueggler17ijrr,Kim21ral} can be misleading (increased RMSE) 
and inconsistent with the visual results (e.g., maps with double and blurred edges).
Therefore, a more sensible figure of merit (FOM) is needed to characterize rotation estimation methods.

In classical BA, the reprojection error (in the image domain) is the FOM that measures the goodness of fit between the unknowns (scene structure/map and camera motion) and the visual data.
Lifting the visual data onto 3D through noisy poses leads to multiple misaligned copies of the same scene part, which when projected gives rise to reprojection errors. 
In our event-based rotational motion scenario, such a misalignment is noticeable on the panoramic map (IWE) in the form of blurred edges or multiple copies of the same scene edge (as seen on synthetic data in \cref{fig:synth_iwe}).
Hence, the thickness of the edges or area occupied by the warped events on the IWE (i.e., the Event Area (EA) \cite{Gallego19cvpr}) is a proxy for the reprojection error. 
The approximation character is due to the fact that reprojection errors require explicit data association (indicating which events correspond to the same edge), but since there is no groundtruth data association we adopt the implicit and soft data association of the IWE: the closer the warped events, the higher their correspondence \cite{Gallego18cvpr}. 
The proxy reprojection error highlights better the issues of the BA problem considered, 
hence we use it to guide our discussion in the upcoming experiments.

\begin{table*}
\caption{\label{tab:reprojection_error}
\emph{Proxy reprojection error} given by the Event Area (EA) of the panoramic IWE. 
The Gradient Magnitude (GM) \eqref{eq:gradmagIWE} of the IWE is also reported.
Data from \bicycle{}, \town{}, and the events in [1,11] s of the ECD \cite{Mueggler17ijrr} sequences (\cref{fig:iwe_sharpness}).
}
\centering
\adjustbox{max width=\linewidth}{
\setlength{\tabcolsep}{3pt}
\begin{tabular}{llrrrrrrrrrrrr} 
\toprule
& Sequence & \multicolumn{2}{c}{\text{bicycle} (synth)}
         & \multicolumn{2}{c}{\text{town} (synth)}
         & \multicolumn{2}{c}{\text{shapes} (real)}
         & \multicolumn{2}{c}{\text{poster} (real)}
         & \multicolumn{2}{c}{\text{boxes} (real)}
         & \multicolumn{2}{c}{\text{dynamic} (real)} \\
\cmidrule(l{1mm}r{1mm}){3-4}
\cmidrule(l{1mm}r{1mm}){5-6}
\cmidrule(l{1mm}r{1mm}){7-8}
\cmidrule(l{1mm}r{1mm}){9-10}
\cmidrule(l{1mm}r{1mm}){11-12}
\cmidrule(l{1mm}r{1mm}){13-14}
& & EA [\%] $\downarrow$ & GM $\uparrow$
& EA [\%] $\downarrow$ & GM $\uparrow$
& EA [\%] $\downarrow$ & GM $\uparrow$
& EA [\%] $\downarrow$ & GM $\uparrow$
& EA [\%] $\downarrow$ & GM $\uparrow$
& EA [\%] $\downarrow$ & GM $\uparrow$\\

\midrule
\multirow{5}{*}{\shortstack{Front-end \\ \cref{sec:experiments:synth:frontend} \\ \cref{sec:experiments:real:frontend}}}

& EKF-\smt{} \cite{Kim18phd} & 4.390 & 47.693 & 8.452 & 108.893 & - & - & - & - & - & - & - & - \\
 
& \rtpt{} \cite{Reinbacher17iccp} & - & - & - & - & 0.459 & 14.448 & 3.309 & 22.231 & 2.942 & 15.744 & 2.167 & 17.143 \\
 
& \cmaxgae{} \cite{Kim21ral} & 4.424 & 48.474 & 8.082 & 115.517 & 0.514 & 10.186 & 3.412 & 16.718 & 2.980 & 12.432 & 2.273 & 12.611 \\

& IMU dead reckoning & 4.152 & 65.198 & 7.393 & 172.964 & 0.604 & 5.467 & 3.601 & 7.830 & 3.146 & 9.371 & 2.473 & 10.507 \\

& Groundtruth & 4.152 & 65.216 & 7.391 & 173.055 & 0.532 & 6.410 & 3.396 & 9.648 & 2.957 & 11.812 & 2.222 & 12.602 \\

\midrule
\multirow{4}{*}{\shortstack{BA: linear \\ \cref{sec:experiments:synth:backend} \\ \cref{sec:experiments:real:backend}}}

& EKF-\smt{} \cite{Kim18phd} & 4.163 & 64.634 & 7.453 & 169.618 & N/A & N/A & N/A & N/A & N/A & N/A & N/A & N/A \\

& \rtpt{} \cite{Reinbacher17iccp} & N/A & N/A & N/A & N/A & 0.332 & 35.187 & 3.038 & 46.776 & 2.724 & 38.690 & 1.856 & 46.882 \\

& \cmaxgae{} \cite{Kim21ral} & 4.185 & 63.472 & 7.506 & 165.736 & 0.334 & 34.818 & 3.039 & 46.679 & 2.742 & 37.055 & 1.859 & 46.373 \\

& IMU dead reckoning & N/A & N/A & N/A & N/A & 0.331 & 35.389 & 3.041 & 46.557 & 2.724 & 38.695 & 1.848 & 47.813 \\

\midrule
\multirow{4}{*}{\shortstack{BA: cubic \\ \cref{sec:experiments:synth:backend} \\ \cref{sec:experiments:real:backend}}}

& EKF-\smt{} \cite{Kim18phd} & 4.210 & 61.765 & 7.623 & 159.119 & N/A & N/A & N/A & N/A & N/A & N/A & N/A & N/A \\
 
& \rtpt{} \cite{Reinbacher17iccp} & N/A & N/A & N/A & N/A & 0.335 & 34.587 & 3.041 & 46.420 & 2.733 & 37.787 & 1.878 & 43.811 \\

& \cmaxgae{} \cite{Kim21ral} & 4.205 & 62.256 & 7.561 & 161.898 & 0.344 & 32.861 & 3.054 & 45.299 & 2.878 & 26.509 & 1.906 & 40.789 \\

& IMU dead reckoning & N/A & N/A & N/A & N/A & 0.333 & 34.965 & 3.064 & 44.208 & 2.725 & 38.623 & 1.855 & 46.932 \\

\midrule
\multirow{2}{*}{\shortstack{\cref{sec:experiments:synth:system} \\ \cref{sec:experiments:real:system}}}

& \cmaxslam{} (linear) & 4.166 & 64.305 & 7.494 & 166.333 & 0.332 & 35.243 & 3.037 & 46.961 & 2.725 & 38.602 & 1.850 & 47.443 \\

& \cmaxslam{} (cubic) & 4.166 & 64.244 & 7.471 & 168.254 & 0.340 & 33.977 & 3.040 & 46.708 & 2.727 & 38.378 & 1.862 & 45.882 \\

\bottomrule
\end{tabular}
}
\begin{tablenotes}
\item 
``-'' means the method fails on a sequence (same as in \cref{tab:real_data}), and ``N/A'' indicates that the BA refinement is not applicable because the corresponding front-end failed on this sequence. 
BA is marked as ``N/A'' on trajectories from IMU dead reckoning on synthetic data because here the IMU is noise-free.
\end{tablenotes}
\end{table*}

\subsubsection{Comparison of Front-ends}
\label{sec:experiments:real:frontend}
The top part of \cref{tab:real_data} reports the RMSE of the trajectories estimated by all front-ends on real-world sequences.
Additionally, \cref{fig:front-end_traj} depicts the trajectories produced by all front-ends on the ECD sequences \cite{Mueggler17ijrr}, compared against the GT,
while the corresponding error statistics are displayed in \cref{fig:front-end_err}.

\emph{\smt{}}:
\indoor{} is the only sequence that PF-\smt{} is able to track completely. 
EKF-\smt{} reports the highest accuracy on \fastmotion{}, while shows similar performance as PF-\smt{} on \indoor{}. 
However, both \smt{} variants fail on all ECD sequences. 
As shown in \cref{fig:front-end_traj}, \smt{} tracks accurately at the beginning (thanks to the stable IMU initialization), but then it loses track suddenly at some point, instead of accumulating drift (like the IMU dead reckoning does).
Tracking failure happens mostly when the camera changes the rotation direction abruptly (e.g., turning back).
We suspect it is due to the error propagation between the tracking and mapping threads. 
Small errors in the poses or the map are amplified, corrupting the states and their uncertainty in the Bayesian filters.

\emph{\rtpt{}}
manages to track well the sequences where the camera's FOV moves around the center of the panoramic map \cite{Mueggler17ijrr}.
However, once the camera explores a larger region, \rtpt{} reports dramatically increased errors (e.g., \fastmotion{}), or even loses tracking (e.g., \indoor{}).
Again, this evidences a limitation on the range of trackable camera motions.

\emph{CMax}:
Our \cmaxw{} front-end computes accurate angular velocities, producing good results on all sequences.
As shown in \cref{fig:front-end_traj}, it is clearly better than IMU integration.
On real-world data, it shows robustness and does not suffer from dramatic increase in errors.
\cmaxgae{} jointly estimates angular velocities and absolute rotations, 
which endows better consistency for long-term tracking, as shown by its competitive performance on most real-world sequences.
Due to reason described in \cref{sec:experiments:synth:frontend}, it fails on \boxes{}, where the texture of the scene is higher than in others.

\subsubsection{Back-end (BA)}
\label{sec:experiments:real:backend}

\def\figWidth{0.19\linewidth}
\begin{figure*}[t]
	\centering
    {\small
    \setlength{\tabcolsep}{1pt}
	\begin{tabular}{
	>{\centering\arraybackslash}m{0.4cm} 
	>{\centering\arraybackslash}m{\figWidth} 
    >{\centering\arraybackslash}m{\figWidth} 
	>{\centering\arraybackslash}m{\figWidth}
	>{\centering\arraybackslash}m{\figWidth}
	>{\centering\arraybackslash}m{\figWidth}}
 	 
		\rotatebox{90}{\makecell{shapes}}
        &\gframe{\includegraphics[width=\linewidth]{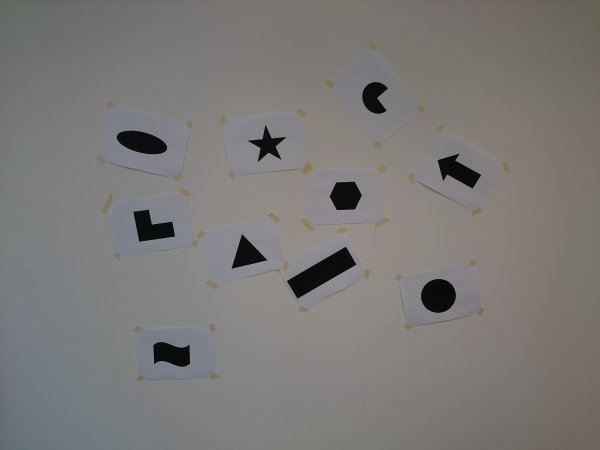}}
		&\gframe{\includegraphics[width=\linewidth]{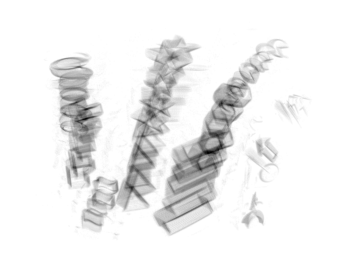}}
		&\gframe{\includegraphics[width=\linewidth]{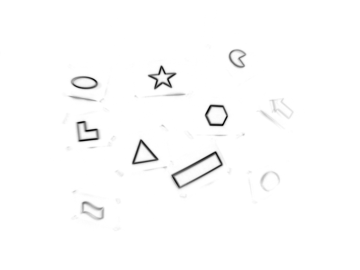}}
		&\gframe{\includegraphics[width=\linewidth]{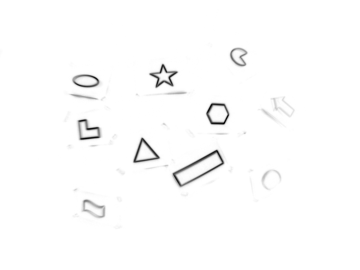}}
		&\gframe{\includegraphics[width=\linewidth]{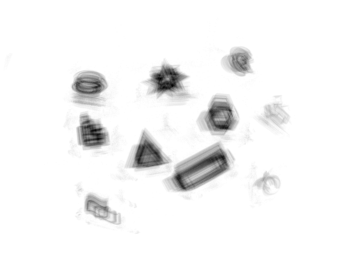}}
		\\
		
		\rotatebox{90}{\makecell{poster}}
        &\gframe{\includegraphics[width=\linewidth]{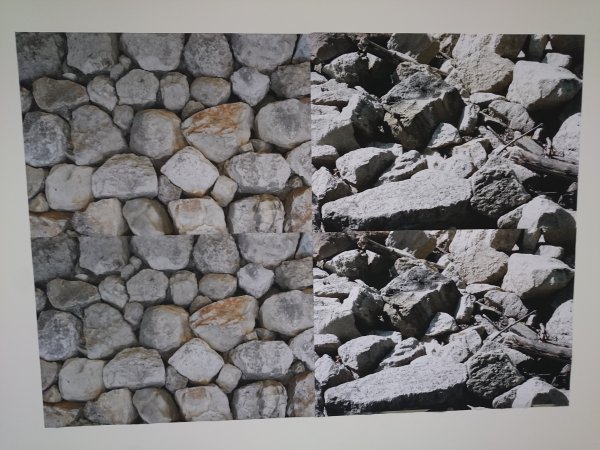}}
		&\gframe{\includegraphics[width=\linewidth]{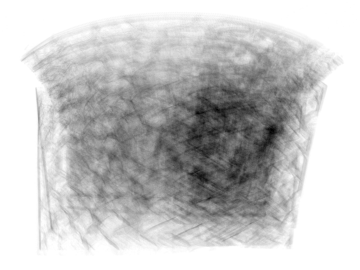}}
		&\gframe{\includegraphics[width=\linewidth]{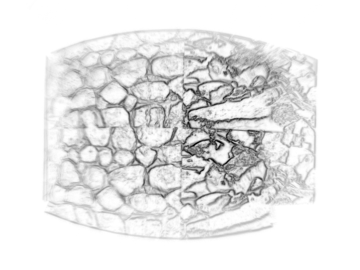}}
		&\gframe{\includegraphics[width=\linewidth]{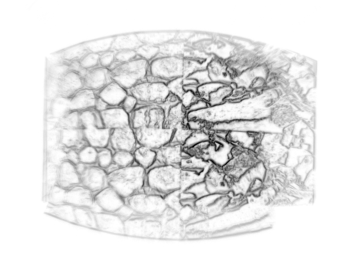}}
		&\gframe{\includegraphics[width=\linewidth]{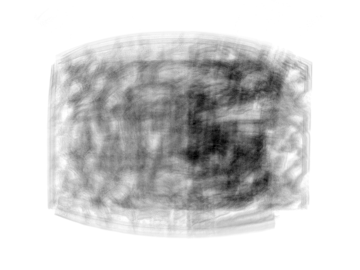}}
		\\

        \rotatebox{90}{\makecell{boxes}}
        &\gframe{\includegraphics[width=\linewidth]{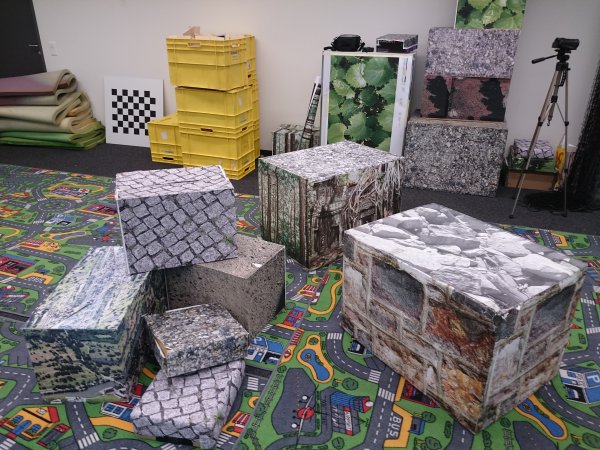}}
		&\gframe{\includegraphics[width=\linewidth]{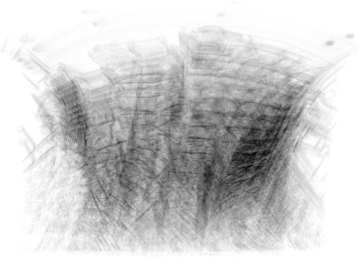}}
		&\gframe{\includegraphics[width=\linewidth]{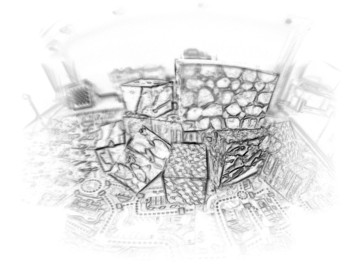}}
		&\gframe{\includegraphics[width=\linewidth]{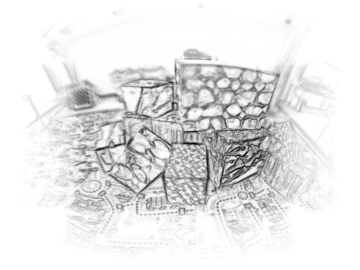}}
		&\gframe{\includegraphics[width=\linewidth]{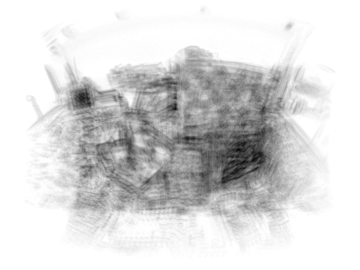}}
		\\
        
        \rotatebox{90}{\makecell{dynamic}}
        &\gframe{\includegraphics[width=\linewidth]{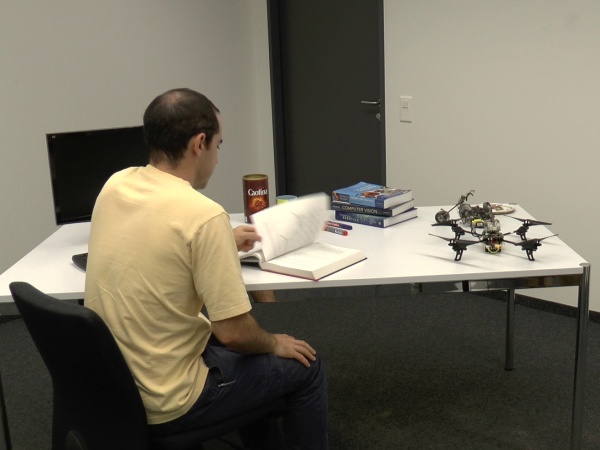}}
		&\gframe{\includegraphics[width=\linewidth]{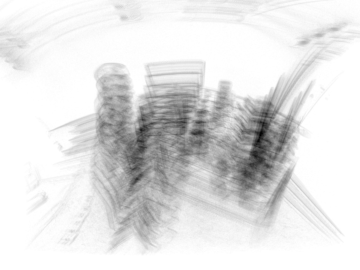}}
		&\gframe{\includegraphics[width=\linewidth]{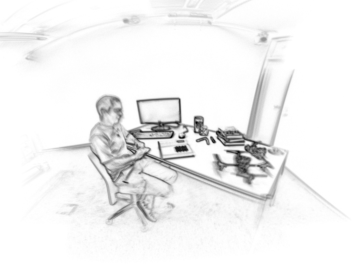}}
		&\gframe{\includegraphics[width=\linewidth]{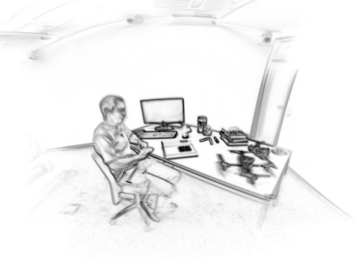}}
		&\gframe{\includegraphics[width=\linewidth]{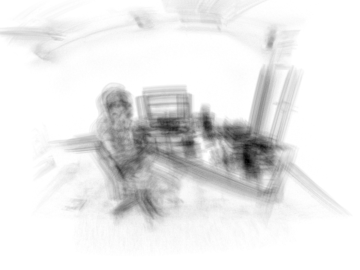}}
		\\

        & (a) Scene
        & (b) IMU dead reckoning
        & (c) Refined trajectories
        & (d) Refined trajectories
        & (e) Groundtruth\\
        &
        & 
        & (linear splines)
        & (cubic splines)
        & (only rotation)
	\end{tabular}
	}
    \caption{\emph{Effect of Bundle Adjustment (Offline smoothing)}.
    Central part of the panoramic IWEs generated using the estimated trajectories (before/after BA refinement) and GT.
    Gamma correction: $\gamma=0.75$. 
    Data from ECD \cite{Mueggler17ijrr}, using events in $[1,11]$s.
    \label{fig:iwe_sharpness}}
\end{figure*}
\def\figWidth{0.19\linewidth}
\begin{figure*}[t]
	\centering
    {\small
    \setlength{\tabcolsep}{1pt}
	\begin{tabular}{
	>{\centering\arraybackslash}m{0.4cm} 
	>{\centering\arraybackslash}m{\figWidth} 
	>{\centering\arraybackslash}m{\figWidth}
	>{\centering\arraybackslash}m{\figWidth}
	>{\centering\arraybackslash}m{\figWidth}
        >{\centering\arraybackslash}m{\figWidth}}
 	 
		\rotatebox{90}{\makecell{shapes}}
		&\includegraphics[width=\linewidth]{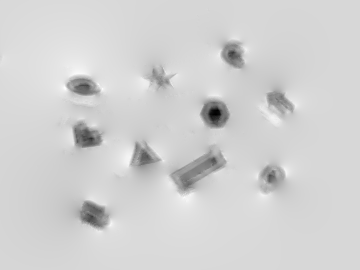}
		&\includegraphics[width=\linewidth]{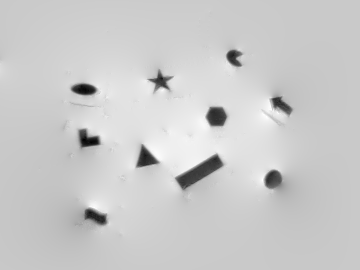}
        &\includegraphics[width=\linewidth]{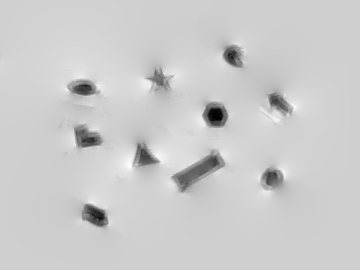}
		  &\includegraphics[width=\linewidth]{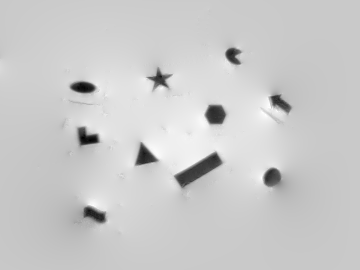}
        &\includegraphics[width=\linewidth]{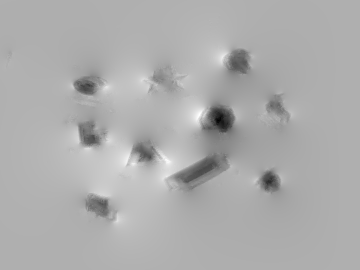}
		\\
		
		\rotatebox{90}{\makecell{poster}}
		&\includegraphics[width=\linewidth]{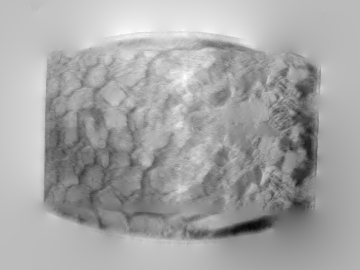}
		&\includegraphics[width=\linewidth]{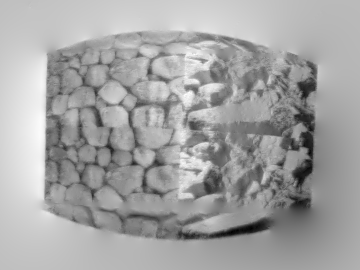}
        &\includegraphics[width=\linewidth]{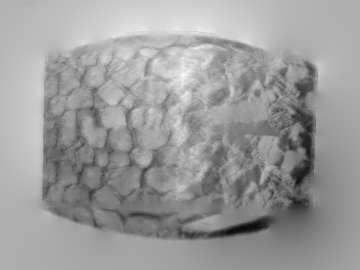}
		  &\includegraphics[width=\linewidth]{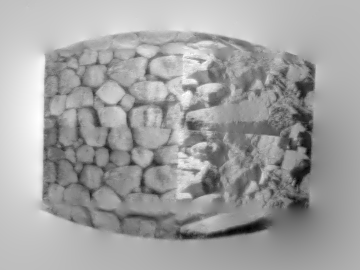}
        &\includegraphics[width=\linewidth]{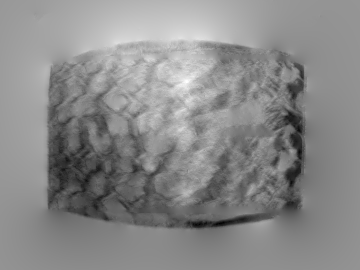}
		\\

        \rotatebox{90}{\makecell{boxes}}
		&\includegraphics[width=\linewidth]{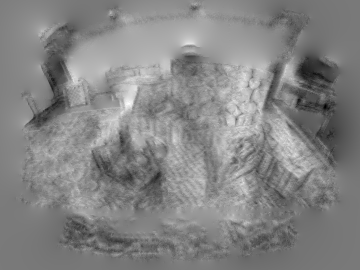}
		&\includegraphics[width=\linewidth]{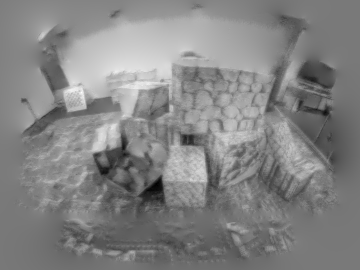}
        &\includegraphics[width=\linewidth]{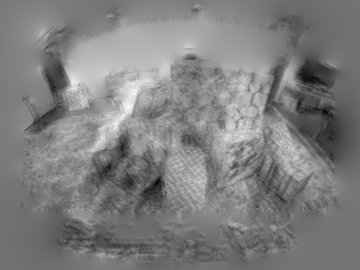}
        &\includegraphics[width=\linewidth]{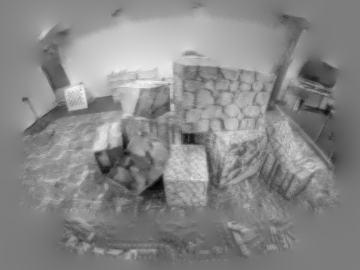}
        &\includegraphics[width=\linewidth]{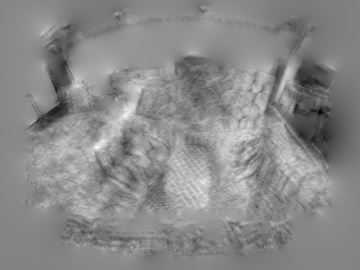}
		\\
        
        \rotatebox{90}{\makecell{dynamic}}
		&\includegraphics[width=\linewidth]{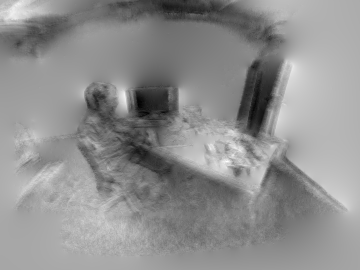}
		&\includegraphics[width=\linewidth]{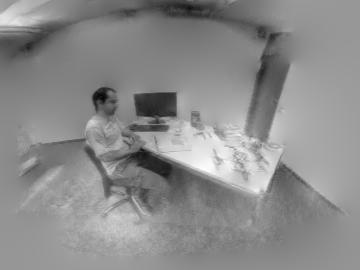}
        &\includegraphics[width=\linewidth]{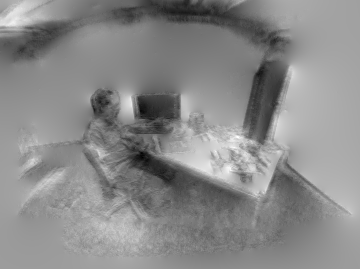}
        &\includegraphics[width=\linewidth]{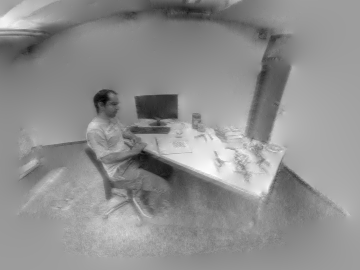}
        &\includegraphics[width=\linewidth]{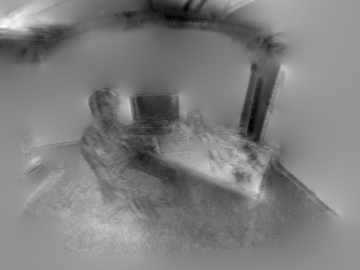}
		\\

        & (a) \cmaxgae{} trajectories
        & (b) Refined \cmaxgae{} trajectories, linear splines.
        & (c) \rtpt{} trajectories
        & (d) Refined \rtpt{} trajectories, linear splines.
        & (e) Groundtruth (only rotation) \\
	\end{tabular}
	}
    \caption{\emph{Effect of Bundle Adjustment (Offline smoothing)}.
    Central part of the panoramic grayscale maps generated using the estimated trajectories (before/after BA refinement) and GT.
    The grayscale map size is $1024 \times 512$ px.
    Same events as \cref{fig:iwe_sharpness}.\label{fig:graymap}}
\end{figure*}

Next, our offline BA approach is tested on real-world data.
Its improvement is most salient on camera trajectories that drift considerably from the GT, like IMU dead reckoning.
\Cref{fig:refined_traj} shows the estimated camera trajectories of the ECD sequences before/after refinement.
The refined trajectories are very close to the GT and have smaller and more concentrated absolute error distributions than the initial ones (\cref{fig:refined_traj_err} and the middle part of \cref{tab:real_data}).
The relative errors do not change considerably,
which is reasonable because they approximately measure the error in angular velocity, and the IMU provides accurate angular velocity data. 

A qualitative comparison of the maps associated to the rotations before/after BA refinement is given in \cref{fig:iwe_sharpness}.
The maps become considerably sharper with the refinement. 
\Cref{tab:reprojection_error} reports that the corresponding reprojection errors (EA) also decrease after BA refinement.
For comparison, the last column of \cref{fig:iwe_sharpness} shows the maps obtained with the GT rotations; 
they are not sharp (because the camera motion is not purely rotational), 
which is also noticeable in higher reprojection errors (EA) than for the refined rotations.

Similar observations apply to the refined trajectories from the visual front-ends.
\Cref{fig:graymap} compares the grayscale maps obtained using the trajectories estimated by \cmaxgae{} and \rtpt{} before/after linear BA refinement, as well as the GT rotations.
The grayscale maps are obtained by feeding the trajectory and events to the mapping module of \smt{}.
The \cmaxgae{}, \rtpt{} and GT rotations result in blurred grayscale panoramas (columns a, c and e).
In contrast, the proposed BA improves the alignment of the visual data (columns b and d).
Quantitatively, this is supported by the reduction of the proxy reprojection error (EA) in \cref{tab:reprojection_error}.

\subsubsection{CMax-SLAM}
\label{sec:experiments:real:system}
\begin{figure*}
     \centering
     \begin{subfigure}[b]{0.6\linewidth}
         \centering
         \includegraphics[width=\linewidth]{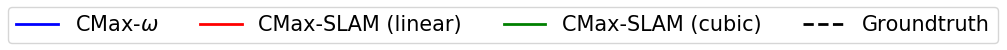}
     \end{subfigure}
     
    \begin{subfigure}[b]{0.245\linewidth}
         \centering
         \includegraphics[width=\linewidth]{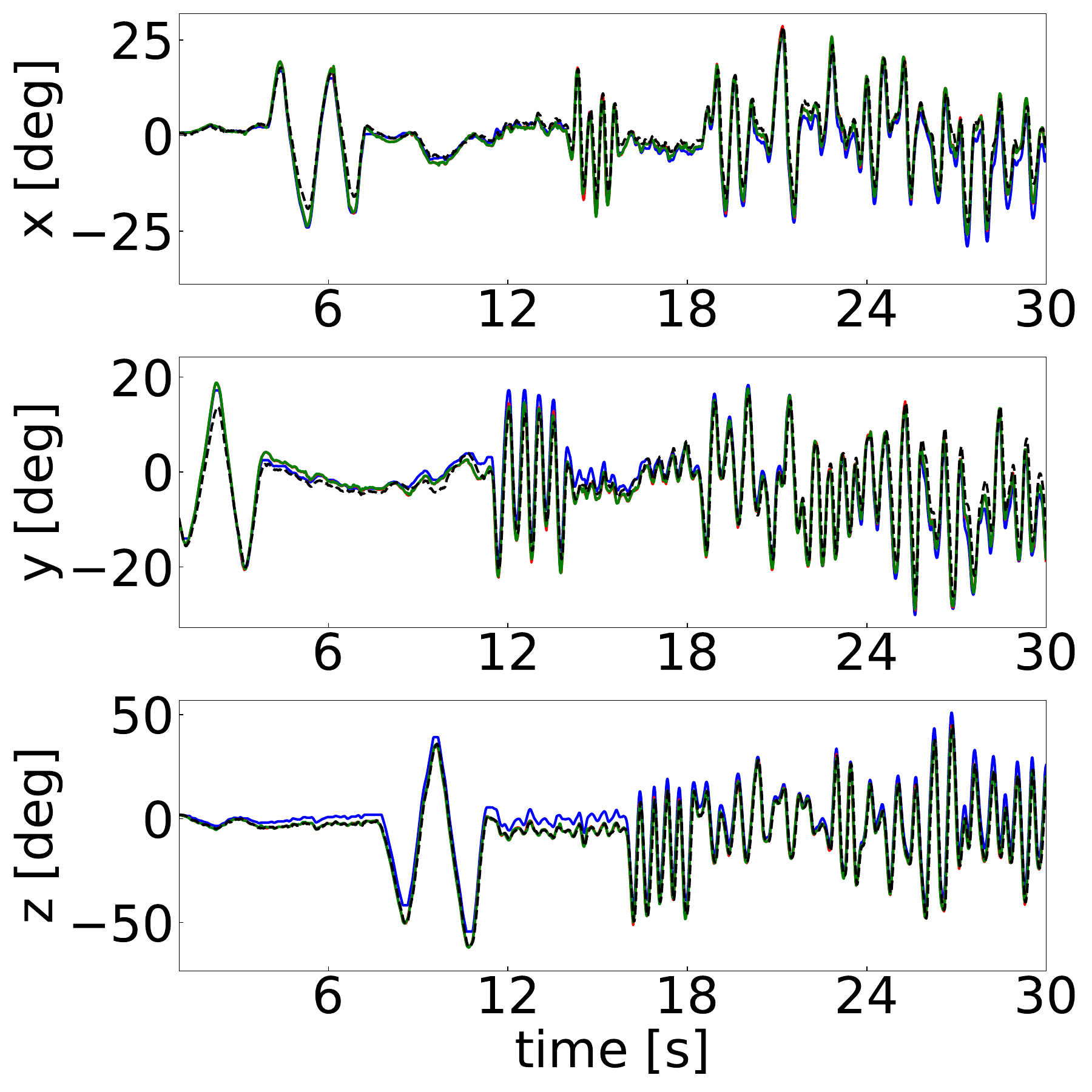}
         \caption{shapes}
         \label{fig:slam_traj:shapes}
     \end{subfigure}
     \hspace{-1ex}
     \begin{subfigure}[b]{0.245\linewidth}
         \centering
         \includegraphics[width=\linewidth]{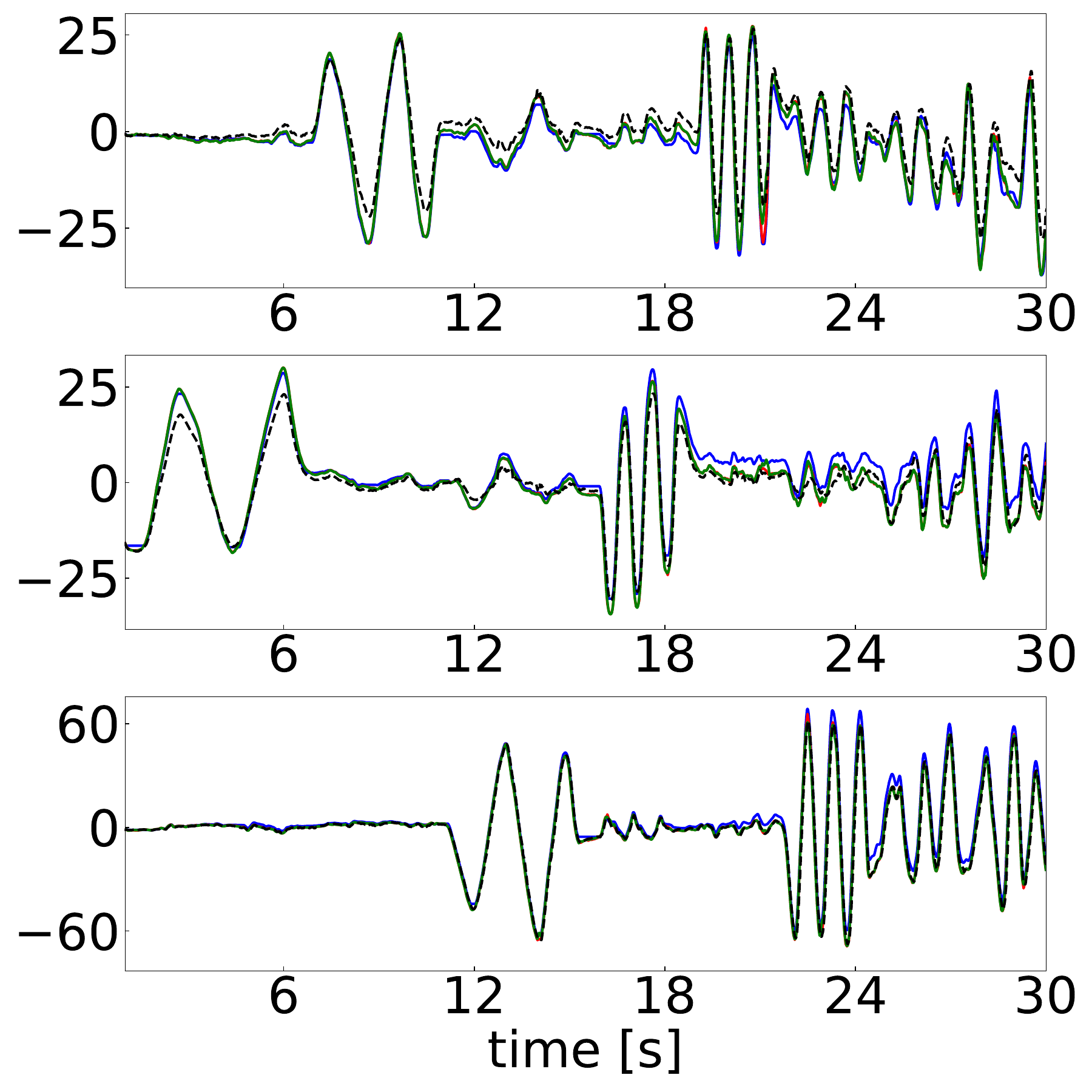}
         \caption{poster}
         \label{fig:slam_traj:poster}
     \end{subfigure}
     \hspace{-2ex}
     \begin{subfigure}[b]{0.245\linewidth}
         \centering
         \includegraphics[width=\linewidth]{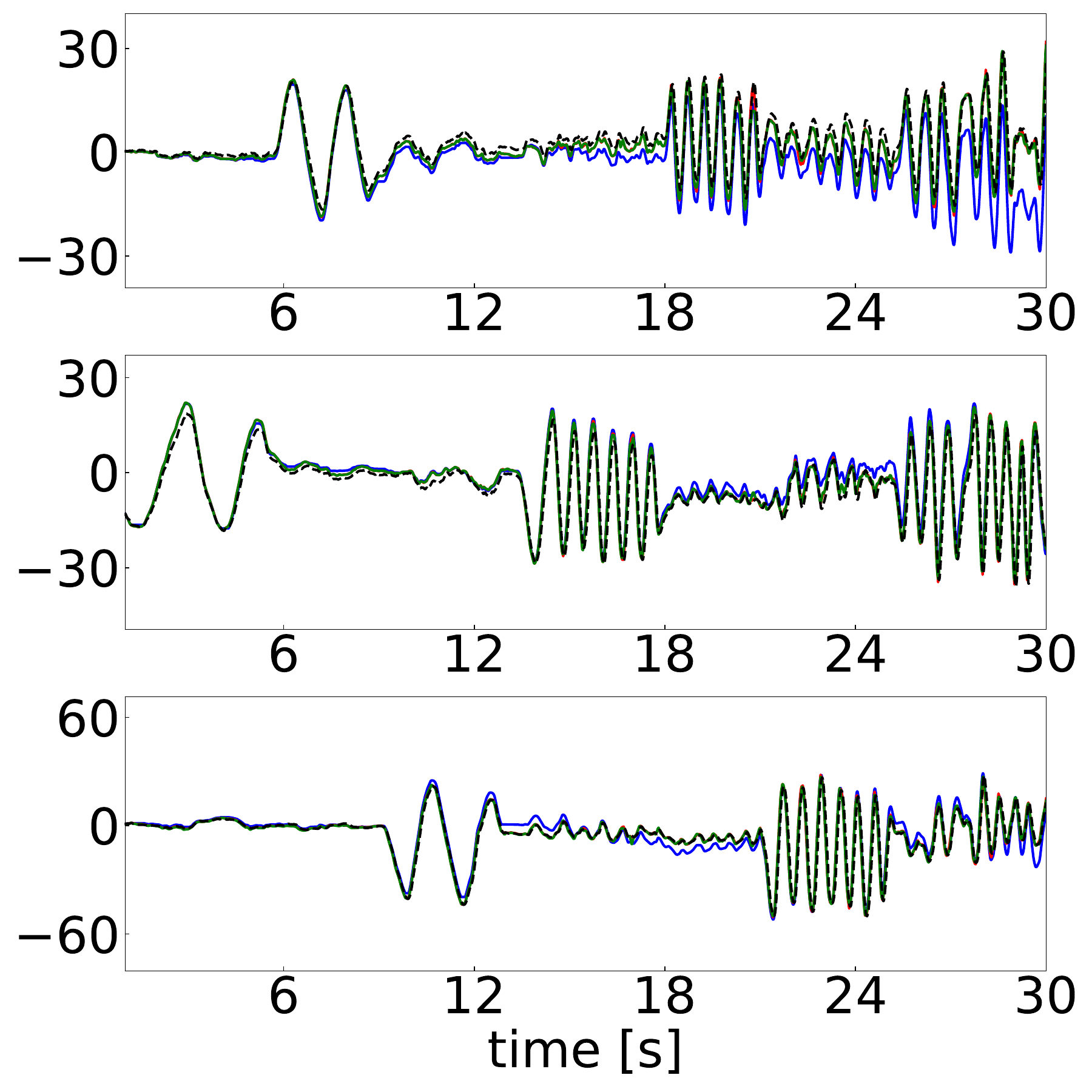}
         \caption{boxes}
         \label{fig:slam_traj:boxes}
     \end{subfigure}
     \hspace{-2ex}
     \begin{subfigure}[b]{0.245\linewidth}
         \centering
         \includegraphics[width=\linewidth]{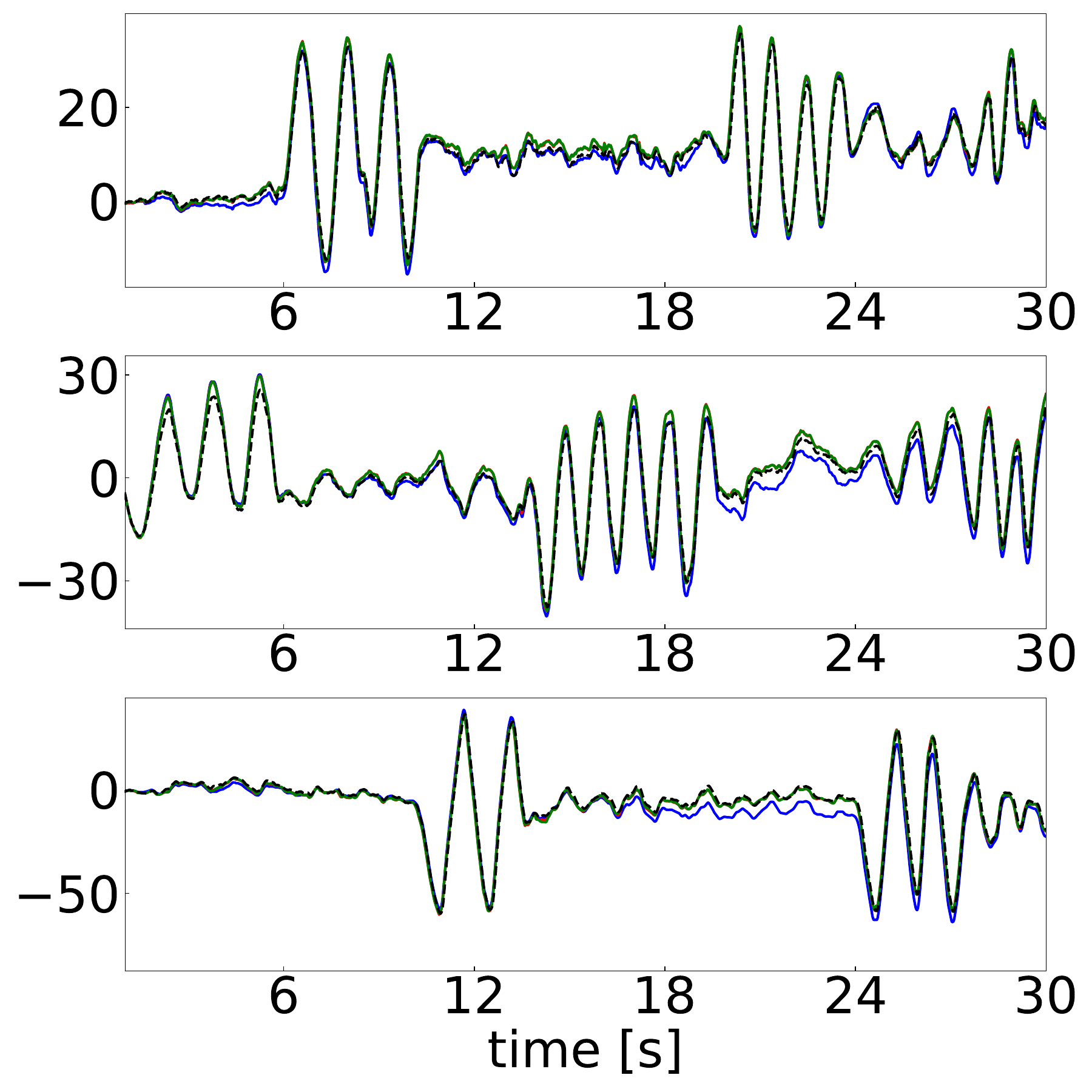}
         \caption{dynamic}
         \label{fig:slam_traj:dynamic}
    \end{subfigure}
        \caption{\emph{\cmaxslam{} (Online)}. 
        Trajectory comparison of CMax front-end and \cmaxslam{} (linear and cubic). 
        \label{fig:slam_traj}}
\vspace{1ex}
     \begin{subfigure}[b]{0.50\linewidth}
         \centering
         \includegraphics[width=\linewidth]{images/error/legend/err_violinplot_legend.png}
     \end{subfigure}
     
     \begin{subfigure}[b]{0.245\linewidth}
         \centering
         \includegraphics[width=\linewidth, trim={1.0cm 1.0cm 1.0cm 0.5cm}]{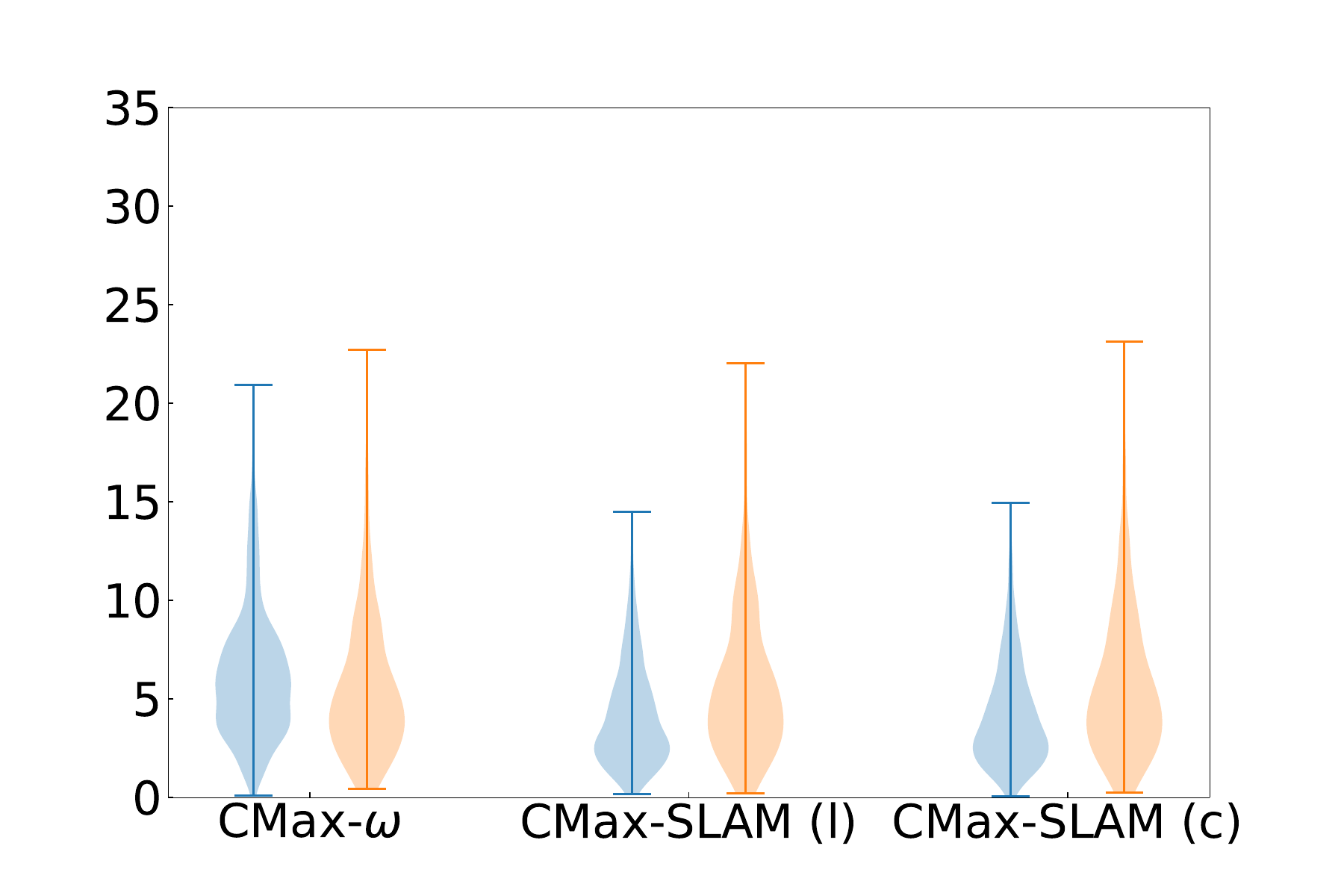}
         \caption{shapes}
         \label{fig:slam_err_violin:shapes}
     \end{subfigure}
    \hspace{-2ex}
     \begin{subfigure}[b]{0.245\linewidth}
         \centering
         \includegraphics[width=\linewidth, trim={1.0cm 1.0cm 1.0cm 0.5cm}]{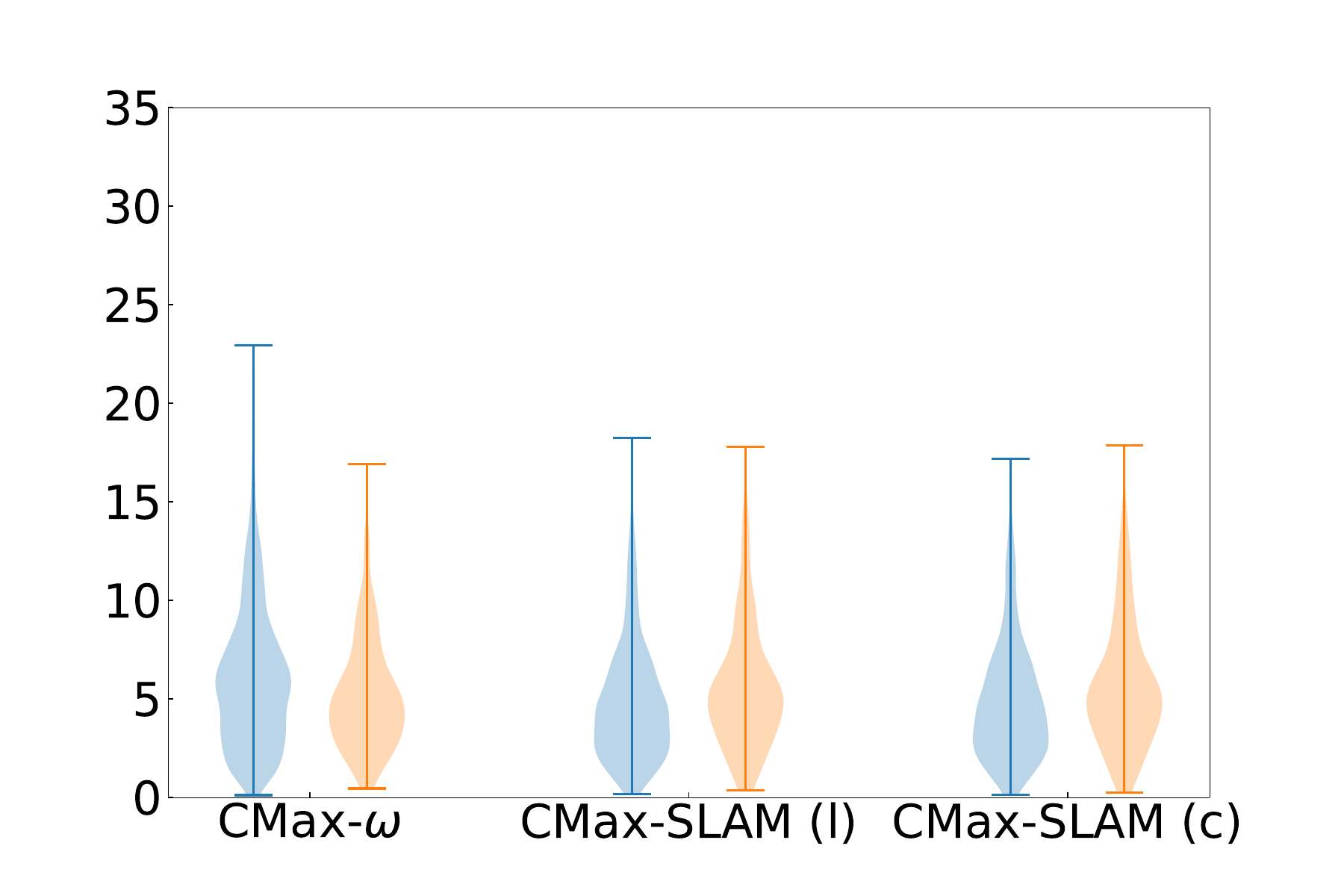}
         \caption{poster}
         \label{fig:slam_err_violin:poster}
     \end{subfigure}
    \hspace{-2ex}
     \begin{subfigure}[b]{0.245\linewidth}
         \centering
         \includegraphics[width=\linewidth, trim={1.0cm 1.0cm 1.0cm 0.5cm}]{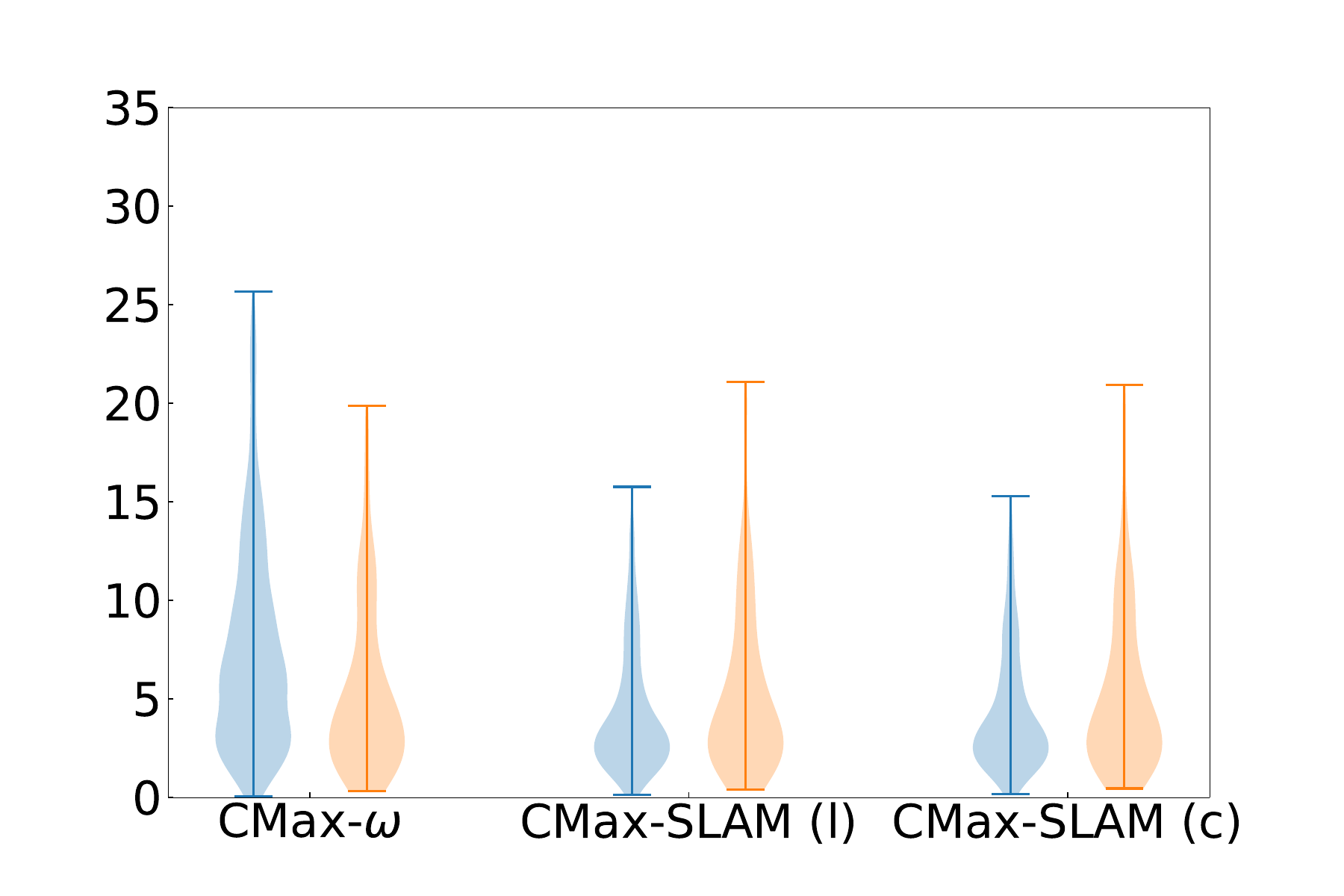}
         \caption{boxes}
         \label{fig:slam_err_violin:boxes}
     \end{subfigure}
    \hspace{-2ex}
     \begin{subfigure}[b]{0.245\linewidth}
         \centering
         \includegraphics[width=\linewidth, trim={1.0cm 1.0cm 1.0cm 0.5cm}]{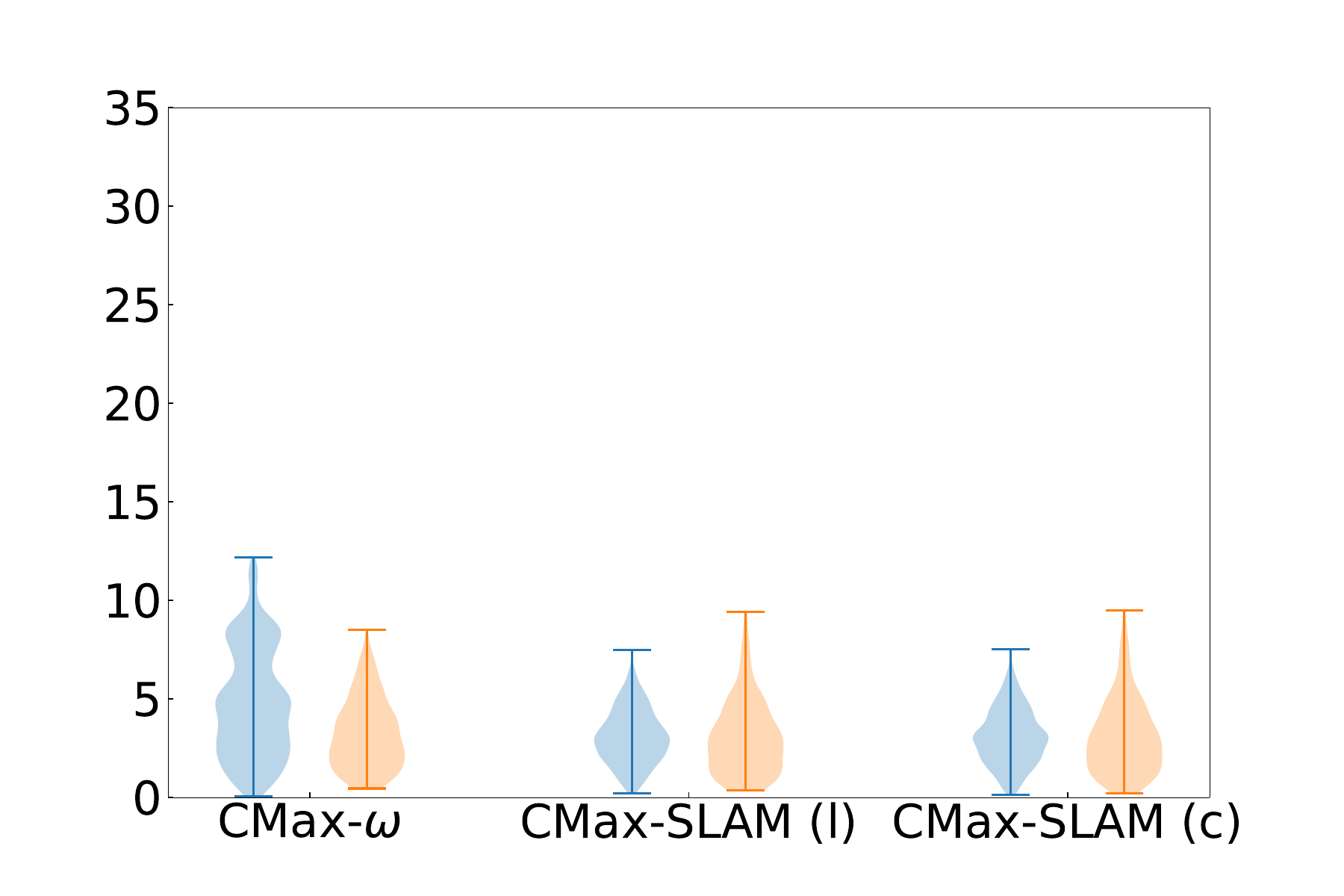}
         \caption{dynamic}
         \label{fig:slam_err_violin:dynamic}
     \end{subfigure}
     \addtocounter{figure}{-1}
    \captionof{figure}{\emph{\cmaxslam{} (Online)}. 
    Absolute and relative errors of the compared \cmaxslam{}. 
    ``l'' and ``c'' indicate the CMax-SLAM with linear and cubic spline trajectories, respectively.
    \label{fig:slam_err_violin}}
\end{figure*}

Finally, we test the proposed \cmaxslam{} system on the same sequences. 
The estimated trajectories and errors are plotted in \cref{fig:slam_traj,fig:slam_err_violin}, 
RMSE numbers are given in the bottom part of \cref{tab:real_data}, 
and proxy reprojection errors (EA) are given in \cref{tab:reprojection_error}.
The trajectories estimated by \cmaxslam{} are very close to the GT (\cref{fig:slam_traj}).
As the violin plots in \cref{fig:slam_err_violin} show, in most cases, the absolute errors decrease and their distributions become more concentrated than with respect to the \cmaxw{} front-end. 
The effect is most pronounced on \shapes{} and \dynamic{}, where the amount of translational motion is smaller than \poster{} and \boxes{}.
The relative errors remain almost unchanged, thus not spoiling the accurate angular velocities estimated by the front-end.
In terms of proxy reprojection errors, \cmaxslam{} achieves the smallest values on all ECD sequences (\cref{tab:reprojection_error}).

\emph{Remark}: \Cref{tab:reprojection_error} also reports the gradient magnitude (GM) \eqref{eq:gradmagIWE} of the panoramic IWE. 
It provides a measure of IWE sharpness different from the one used in the optimization (i.e., variance).
For completeness, \cref{tab:reprojection_error} also reports the values of synthetic sequences \bicycle{} and \town{}. 
The numbers show that, both in real-world and synthetic data, the criteria are aligned: 
the sharper the IWE (the larger the GM), the smaller the proxy reprojection errors (EA).

\subsection{Runtime Evaluation}
\label{sec:experiments:runtime}

\subsubsection{Comparison of Front-ends}
\label{sec:experiments:runtime:frontend}

Since the pipelines of the tested methods are completely different, to compare their efficiency we run them on selected segments from the ECD dataset and measure their processing time (\cref{tab:frontend_runtime_eval}).
All front-ends process all events except for \cmaxgae{}, which samples one out of four events (i.e., only processes 25\% of the events, as per its original implementation).
EKF-\smt{} is faster than the others in high-texture scenes (\poster{} and \boxes{}),
while \cmaxgae{} reports the shortest runtime in low-texture scenes (\shapes{} and \dynamic{}) by throwing away events.
\cmaxw{} is competitive while processing all events and producing poses at \mbox{100 Hz} (vs. \mbox{40 Hz} of \cmaxgae{}).
Overall, none of the front-end methods show real-time performance except on \shapes{}, whose texture is very simple.

\begin{table}
\caption{\label{tab:frontend_runtime_eval}Runtime evaluation of compared front-end methods [s]. 
Data from ECD \cite{Mueggler17ijrr}, using events from 0 to 8 s.
}
\centering
\adjustbox{max width=\columnwidth}{
\setlength{\tabcolsep}{3pt}
\begin{tabular}{lrrrrr}
\toprule 
Front-end  & PF-\smt{} & EKF-\smt{} & \cmaxw{} & \cmaxgae{} & \rtpt{} \\
methods    & (CPU) & (CPU) & (CPU) & (CPU) & (GPU) \\
\midrule
\shapes{} & 28.29 & 7.92 & 10.65 & 6.03 & 6.50\\
\poster{} & 174.78 & 26.90 & 52.34 & 36.25 & 38.43\\
\boxes{} & 170.66 & 27.08 & 47.65 & 42.12 & 38.24\\
\dynamic{} & 132.38 & 18.13 & 23.04 & 16.77 & 29.15\\
\bottomrule
\end{tabular}
}
\end{table}

\subsubsection{\cmaxslam{}}
\label{sec:experiments:runtime:system}

\Cref{tab:system_runtime_eval} reports the computational cost of \cmaxslam{} for several sequences at two different event camera resolutions: 
$240\times 180$ px (DAVIS240C in the ECD dataset \cite{Mueggler17ijrr}) and $346\times 260$ px (DAVIS346).
It shows that the higher the camera resolution, the longer the processing time.
In addition, the cubic spline trajectory representation is reasonably more expensive than the linear one. 
Currently, the \cmaxslam{} system has been implemented without optimizing for real-time performance.
It has the potential for nearly real-time performance by processing fewer events and reducing the size of the panoramic map (see \cref{sec:experiments:wild_experiments,sec:experiments:super_resolution}).

Additionally, in the accompanying video, 
we use a DAVIS346 to run \cmaxslam{} in nearly real-time, by processing only a small portion of events and using a small panoramic map.
This demonstrates the trade-off mentioned in \cref{sec:method:pipeline:frontend,sec:method:pipeline:backend}, which is further analyzed in the sensitivity studies (\cref{sec:experiments:sensitive_analysis:pano_map_size,sec:experiments:sensitive_analysis:event_sample_rate}) and can serve users to tune the system according to their needs.

\begin{table}
\caption{
\label{tab:system_runtime_eval}\cmaxslam{} runtime evaluation [\si{\mu\s}/event processed], 
for a map of size $1024 \times 512$ px.
Data from ECD \cite{Mueggler17ijrr}, using events from 0 to 10 s.
}
\centering
\adjustbox{max width=\columnwidth}{
\begin{tabular}{lrrr}
\toprule 
Sequence & Front-end & Back-end (linear) & Back-end (cubic) \\ 
\midrule
\shapes{} (DAVIS240C) & 1.725 & 11.204 & 20.406 \\
\poster{} (DAVIS240C) & 1.369 & 5.021 & 8.045 \\
\boxes{} (DAVIS240C) & 1.374 & 4.638 & 7.642 \\
\dynamic{} (DAVIS240C) & 1.344 & 5.888 & 10.740 \\
DAVIS346 & 1.710 & 7.872 & 13.565 \\ 
\bottomrule
\end{tabular}
}
\end{table}

\subsection{Experiments in the Wild for \cmaxslam{}}
\label{sec:experiments:wild_experiments}

\begin{table}
\centering
\caption{\emph{Self-recorded dataset} with VGA-resolution event camera; description of sequences. 
The last column indicates the angular range that the camera rotates around the $Y$~axis).
}
\label{tab:sequences}
\adjustbox{max width=\columnwidth}{
\setlength{\tabcolsep}{3pt}
\begin{tabular}{llrrr}
\toprule 
Camera setup & Sequence & $\sharp$ events [Mev] & Duration [s] & $Y$-angle [$^\circ$] \\
\midrule
Motorized & \gate{} & 97.4 & 8.0 & 360 \\
mount & \palace{} & 115.4 & 8.4 & 360 \\
 & \column{} & 4.9 & 10.0 & 90 \\
 & \mainbuilding{} & 116.2 & 8.5 & 360 \\
 & \marbuilding{} & 33.8 & 4.0 & $\approx$ 90 \\
 & \platzcenter{} & 112.9 & 8.8 & 360\\
 & \platzside{} & 120.1 & 8.0 & 360 \\
\midrule
Hand-held & \crossroad{} & 124.3 & 10.2 & 360 \\
 & \river{} & 69.3 & 5.5 & random \\
 & \bridge{} & 89.0 & 7.5 & random \\
\bottomrule
\end{tabular}
}
\end{table}

\def\figWidth{0.95\linewidth}
\begin{figure*}[t]
	\centering
    {\small
    \setlength{\tabcolsep}{1pt}
	\begin{tabular}{
	>{\centering\arraybackslash}m{0.4cm} 
	>{\centering\arraybackslash}m{\figWidth}}

        \rotatebox{90}{\makecell{\gate{}}}
		&\gframe{\includegraphics[width=\linewidth]{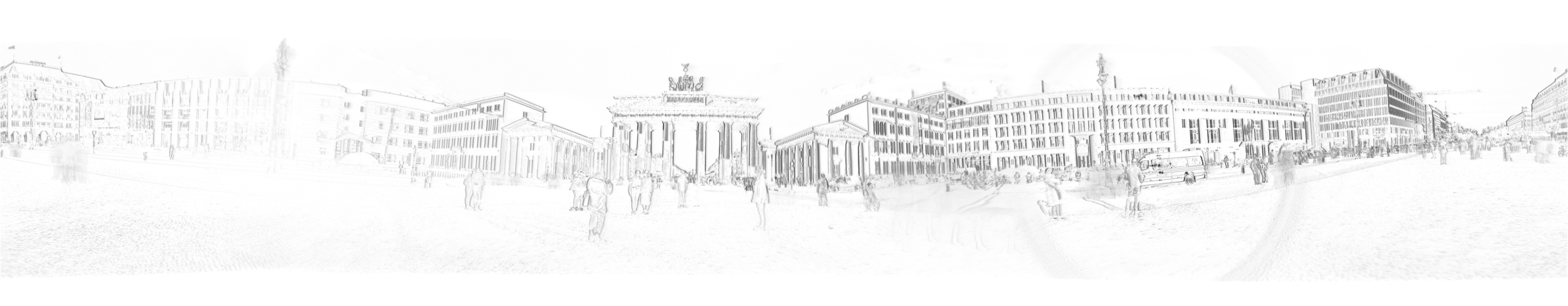}} \\

        \rotatebox{90}{\makecell{\palace{}}}
		&\gframe{\includegraphics[width=\linewidth]{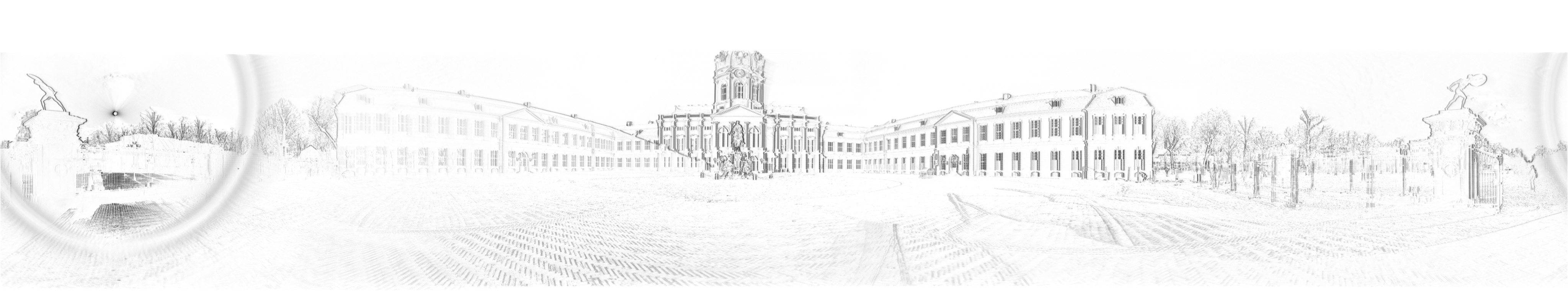}} \\

        \rotatebox{90}{\makecell{\crossroad{}}}
		&\gframe{\includegraphics[width=\linewidth]{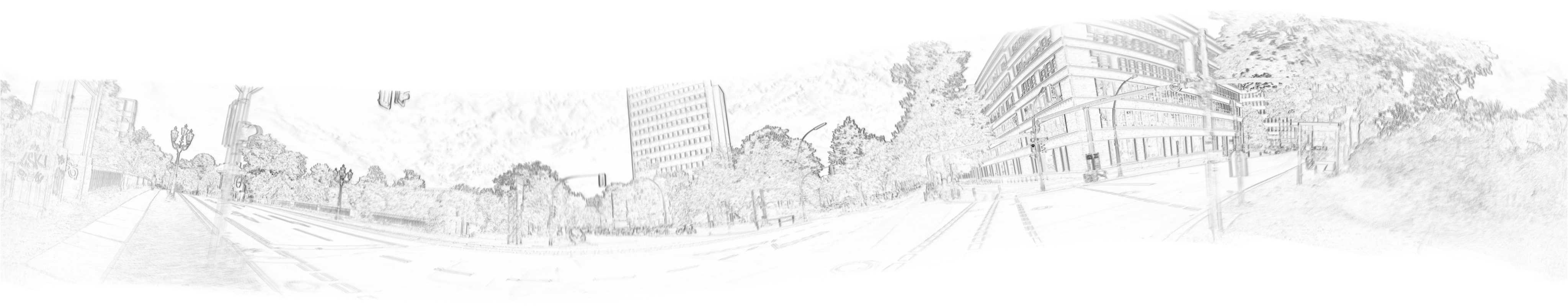}} \\
  
	\end{tabular}
	}
    \caption{\emph{Experiments in the wild}. 
    Panoramic IWEs produced by \cmaxslam{} at $4096 \times 2048$ px resolution, on the data from a $640 \times 480$ px camera. 
    The map sharpness is a proxy for the motion estimation quality. 
    Gamma correction: $\gamma = 0.75$. 
    \label{fig:wild_experiment}
    }
\end{figure*}

To assert that \cmaxslam{} can work reliably in complex natural scenes (i.e., with arbitrary photometric variations), we test it on sequences in the wild, where no GT is available.

We create a new real-world event camera dataset for rotational motion study (\cref{tab:sequences}), which contains ten sequences recorded with a DVXplorer from iniVation AG ($640 \times 480$ px). 
All sequences have events and IMU data (operating at around 800~Hz).
For some sequences we place the camera on a motorized mount (Suptig RSX-350) to produce approximate uniform rotational motion around the $Y$ axis. 
Despite the motorized mount, the camera cannot perform pure rotational motion because the center of rotation still deviates from the camera's optical center.
For the remaining sequences (\crossroad{}, \bridge{} and \river{}), the camera is hand-held, which may introduce more irregular residual translations.

Compared to the six indoor sequences in mocap rooms \cite{Mueggler17bmvc,Kim21ral}, these outdoor sequences contain more difficult brightness conditions (e.g., reflections in the river and windows), dynamic objects (e.g., moving pedestrians, bicycles, cars, leaves and water on the river) and direct sunlight observation causing glare in the lens, which make this dataset more challenging. 

We run \cmaxslam{} on the above sequences and produce panoramic IWEs.
The results in \cref{fig:wild_experiment} indicate that \cmaxslam{} recovers precise global, sharp IWEs for the above challenging scenes.
Results on additional scenes are presented in the accompanying video (suppl. mat.).
For the sequences recorded with the motorized mount, the camera dominantly pans, so the edges that are parallel to the ground trigger very few events.
Therefore, for \gate{} and \palace{} in \cref{fig:wild_experiment}, vertical edges are quite dark and sharp while horizontal edges are not as clear.
However, this situation is alleviated for \crossroad{} because the motion is hand-held.
Both types of motion demonstrate the capability of \cmaxslam{} to work robustly in outdoor scenes.

\subsection{Super-resolution}
\label{sec:experiments:super_resolution}

\subsubsection{\cmaxslam{} at super-resolution}
\label{sec:experiments:super_resolution:cmax_slam}

\def\figWidth{0.19\linewidth}
\begin{figure*}[t]
	\centering
    {\small
    \setlength{\tabcolsep}{1pt}
	\begin{tabular}{
	>{\centering\arraybackslash}m{0.4cm} 
	>{\centering\arraybackslash}m{\figWidth} 
	>{\centering\arraybackslash}m{\figWidth}
	>{\centering\arraybackslash}m{\figWidth}
	>{\centering\arraybackslash}m{\figWidth}
        >{\centering\arraybackslash}m{\figWidth}}

        \rotatebox{90}{\makecell{IWE}}
		&\gframe{\includegraphics[width=\linewidth]{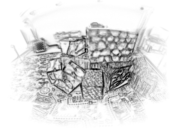}}
		&\gframe{\includegraphics[width=\linewidth]{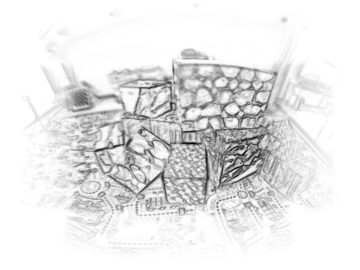}}
		&\gframe{\includegraphics[width=\linewidth]{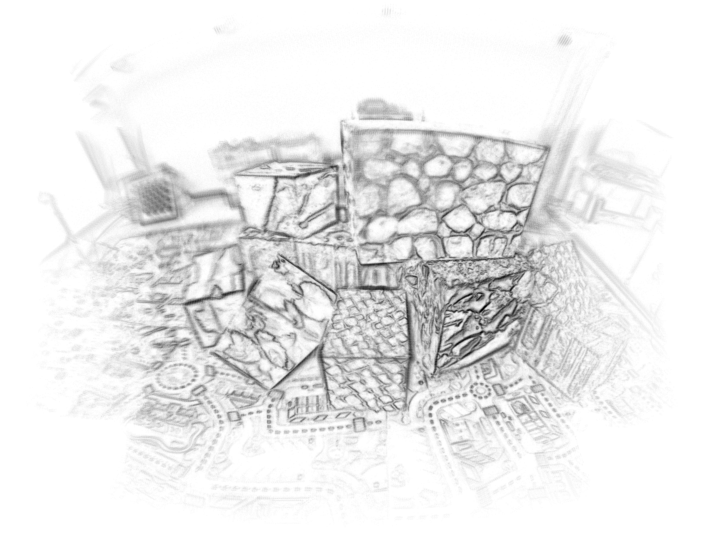}}
            &\gframe{\includegraphics[width=\linewidth]{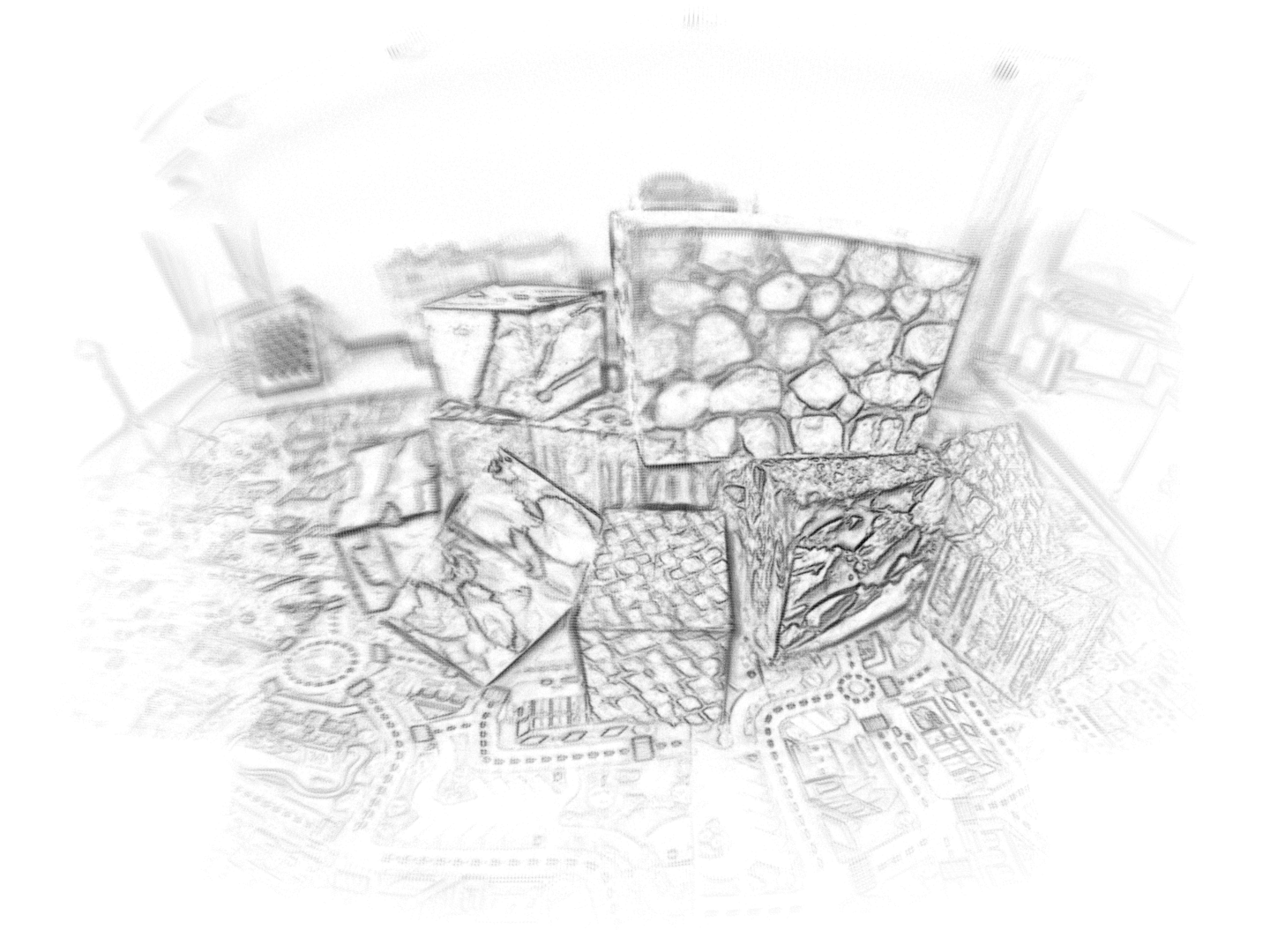}}
            &\gframe{\includegraphics[width=\linewidth]{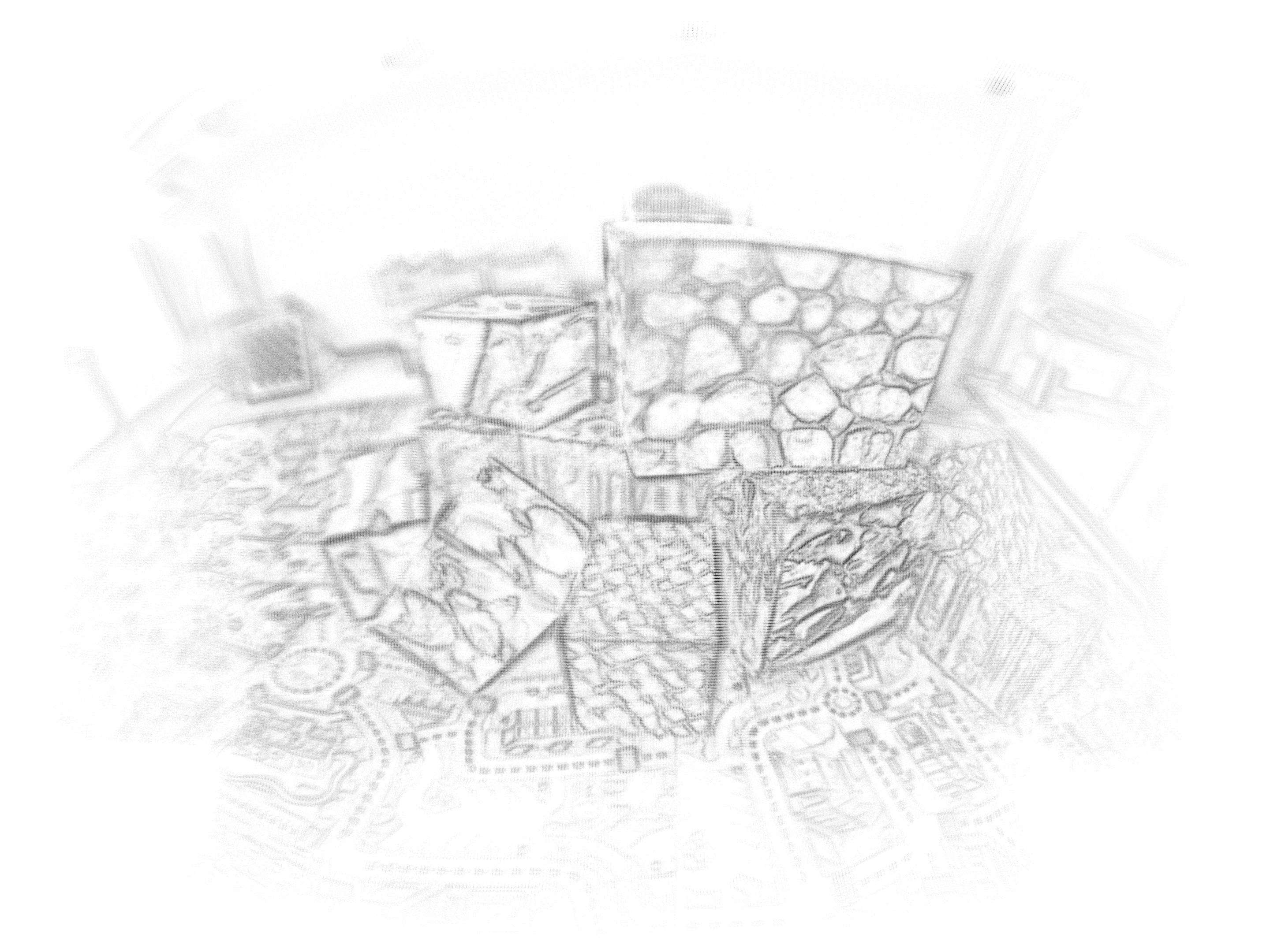}}
		\\
        
        \rotatebox{90}{\makecell{zoomed in}}
		&\gframe{\includegraphics[width=\linewidth]{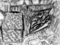}}
		&\gframe{\includegraphics[width=\linewidth]{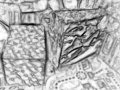}}
		&\gframe{\includegraphics[width=\linewidth]{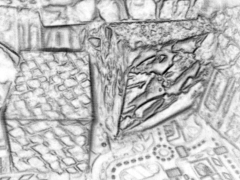}}
            &\gframe{\includegraphics[width=\linewidth]{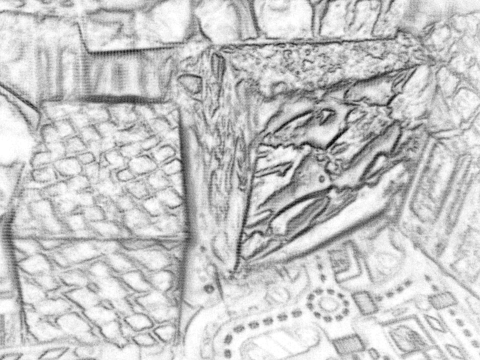}}
            &\gframe{\includegraphics[width=\linewidth]{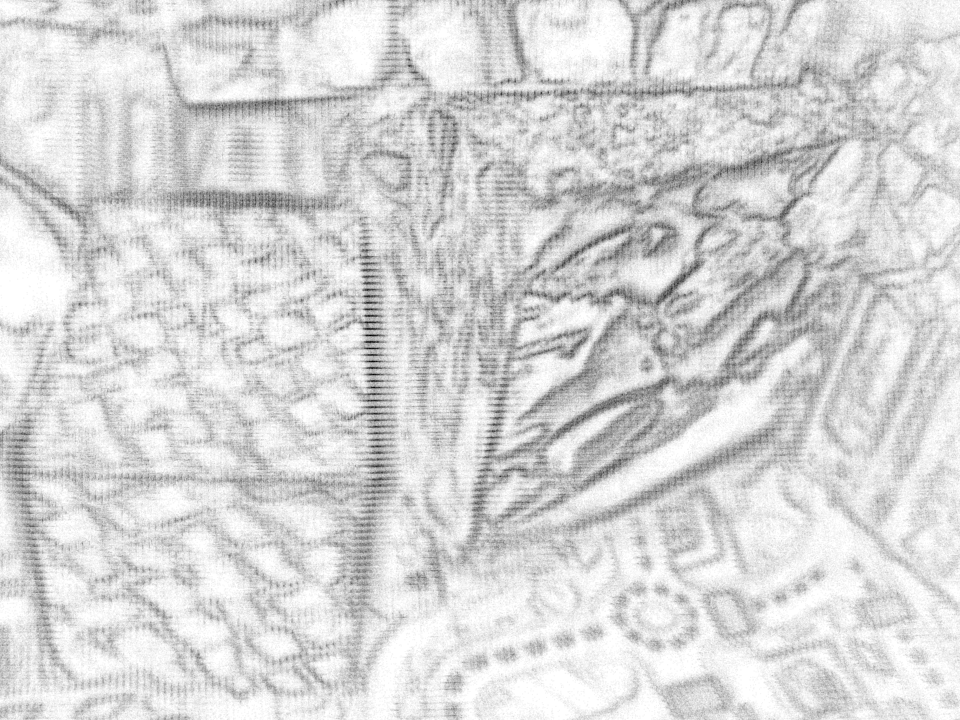}}
		\\

        & (a) $512 \times 256$ px
        & (b) $1024 \times 512$ px
        & (c) $2048 \times 1024$ px
        & (d) $4096 \times 2048$ px
        & (e) $8192 \times 4096$ px \\
	\end{tabular}
	}
    \caption{\emph{\cmaxslam{} at several resolutions}.
    Results of running \cmaxslam{} at different map resolutions (columns), 
    and using the estimated trajectories to generate panoramic IWEs 
    at the corresponding resolutions. 
    Gamma correction: $\gamma$ = 0.75. 
    Data from \boxes{} (same events as \cref{fig:iwe_sharpness}).\label{fig:super_resolution:cmaxslam}
    }
\end{figure*}
\begin{figure}
     \centering
     \begin{subfigure}[b]{0.48\linewidth}
         \centering
         \includegraphics[width=\linewidth]{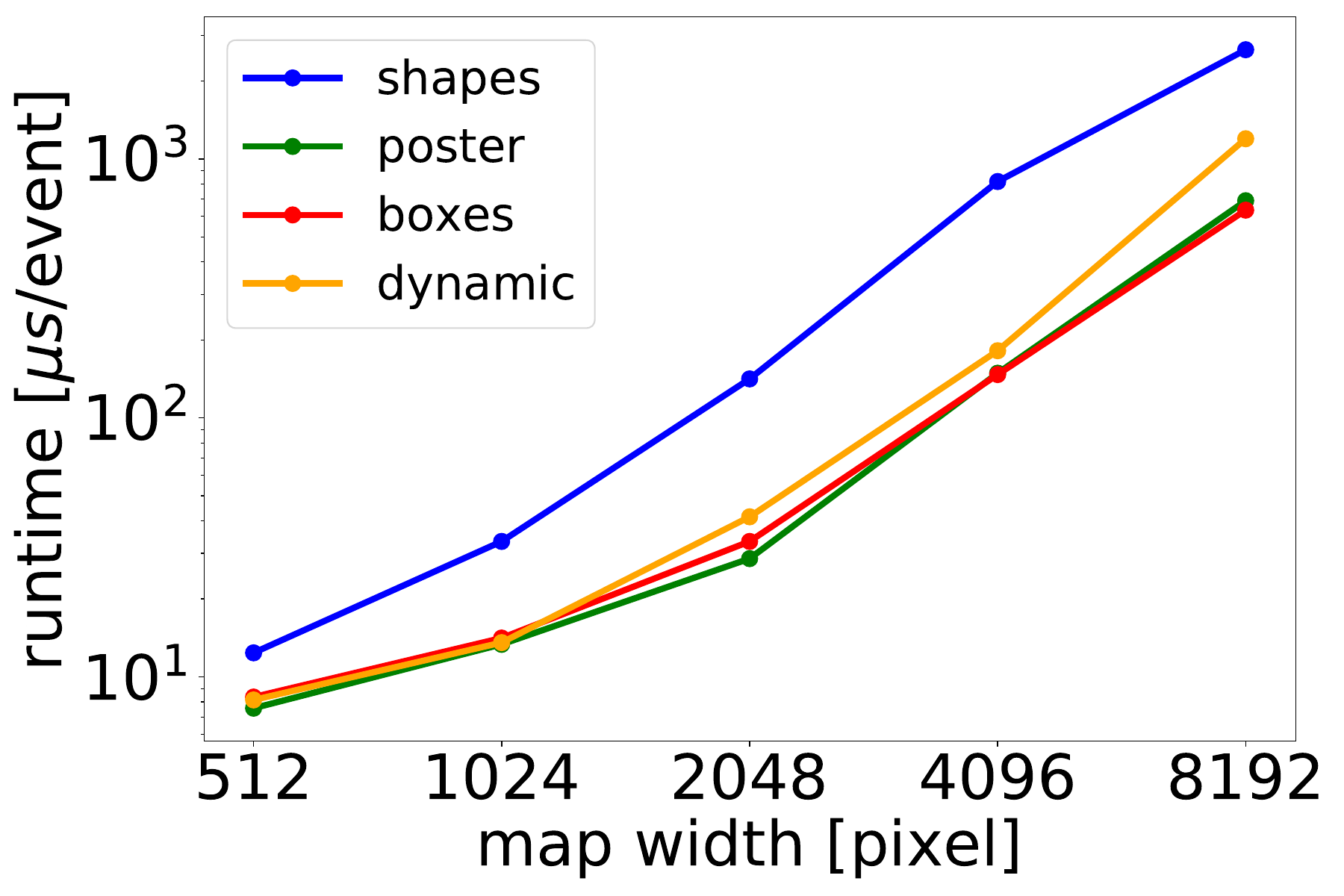}
         \caption{Runtime}
         \label{fig:super_resolution:higher_cmax_slam:runtime}
     \end{subfigure}
     \begin{subfigure}[b]{0.48\linewidth}
         \centering
         \includegraphics[width=\linewidth]{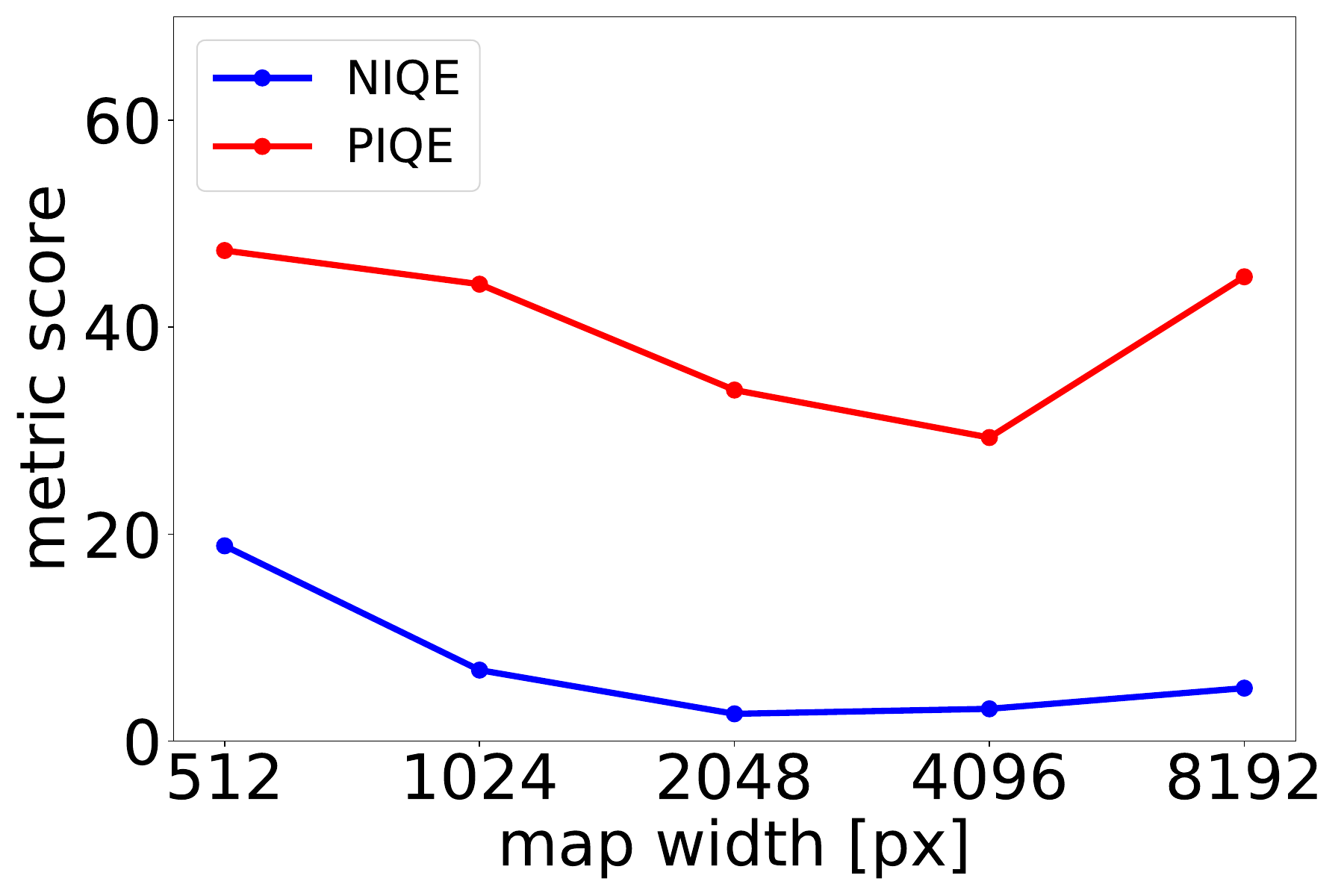}
         \caption{Map quality}
         \label{fig:super_resolution:higher_cmax_slam:map_quality}
     \end{subfigure}
        \caption{\emph{Results of running CMax-SLAM at super resolution} (\cref{fig:super_resolution:cmaxslam}).
        (a) Runtime evaluation of \cmaxslam{} back-end at different resolutions.
        (b) Map quality evaluation of running \cmaxslam{} at different resolutions.
        The lower the NIQE and PIQE scores, the better.
        }
        \label{fig:super_resolution:higher_cmax_slam}
\end{figure}

The resolution of the panoramic map in the back-end is, to some extent, independent of the resolution of the event camera. 
\Cref{fig:super_resolution:cmaxslam} shows the results of running \cmaxslam{} with several map resolutions, from $512 \times 256$ px to $8192 \times 4096$ px.
The larger the map, the larger the memory requirements and the slower the bilinear voting used in the IWE (due to memory access, despite voting complexity being linear with the number of events), hence the slower the optimization. 
As shown in \cref{fig:super_resolution:higher_cmax_slam:runtime}, every time the map size is doubled, the computational cost of the \cmaxslam{} back-end (BA) to process one event approximately increases exponentially.
A very low resolution map warps many events into few map pixels. 
A very large resolution map may contain empty pixels (without warped events), which could be filled in by smoothing the map. 
In both extreme cases, the ego-motion algorithm may fail.
In intermediate cases (\cref{fig:super_resolution:cmaxslam}, columns (b)--(d)), the \cmaxslam{} works well, producing sharp maps because the continuous-time camera trajectory and the high temporal resolution of the events allows us to warp the events in an almost continuous way, thus recovering fine details by converting high temporal resolution into high spatial resolution.
A sensible choice of the map size consists of making the individual map pixels cover approximately the same scene area as those of the event camera.

We evaluate the quality of the central parts of the panoramic IWEs (the first row in \cref{fig:super_resolution:cmaxslam}) by means of the Naturalness Image Quality Evaluator (NIQE) \cite{Mittal13spl} and the Perception‐based Image Quality Evaluator (PIQE) \cite{Venkatanath15ncc}: the lower the scores, the higher the image quality.
As depicted in \cref{fig:super_resolution:higher_cmax_slam:map_quality}, the panoramic IWE at the resolution of $2048 \times 1024$ px has the highest quality, 
closely followed by that of $4096 \times 2048$ px, which agrees with a visual inspection. 

\subsubsection{Grayscale map reconstruction at super-resolution}
\label{sec:experiments:super_resolution:mosaic}

\begin{figure*}[t]
	\centering
    {\small
    \setlength{\tabcolsep}{1pt}
	\begin{tabular}{
	>{\centering\arraybackslash}m{0.4cm} 
	>{\centering\arraybackslash}m{\figWidth} 
	>{\centering\arraybackslash}m{\figWidth}
	>{\centering\arraybackslash}m{\figWidth}
	>{\centering\arraybackslash}m{\figWidth}
        >{\centering\arraybackslash}m{\figWidth}}
 	 
		\rotatebox{90}{\makecell{grayscale map}}
		&\includegraphics[width=\linewidth]{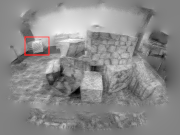}
		&\includegraphics[width=\linewidth]{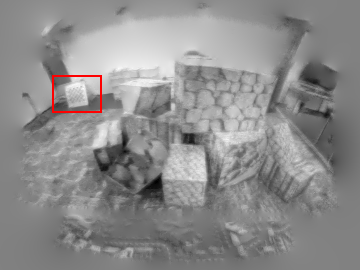}
		&\includegraphics[width=\linewidth]{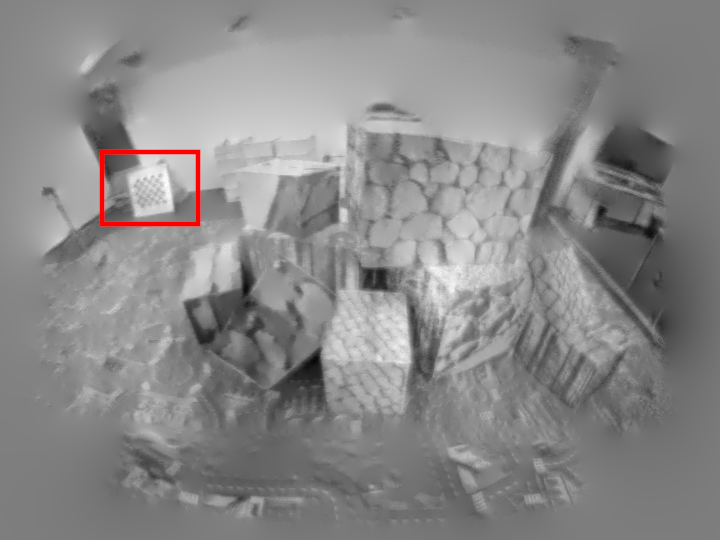}
        &\includegraphics[width=\linewidth]{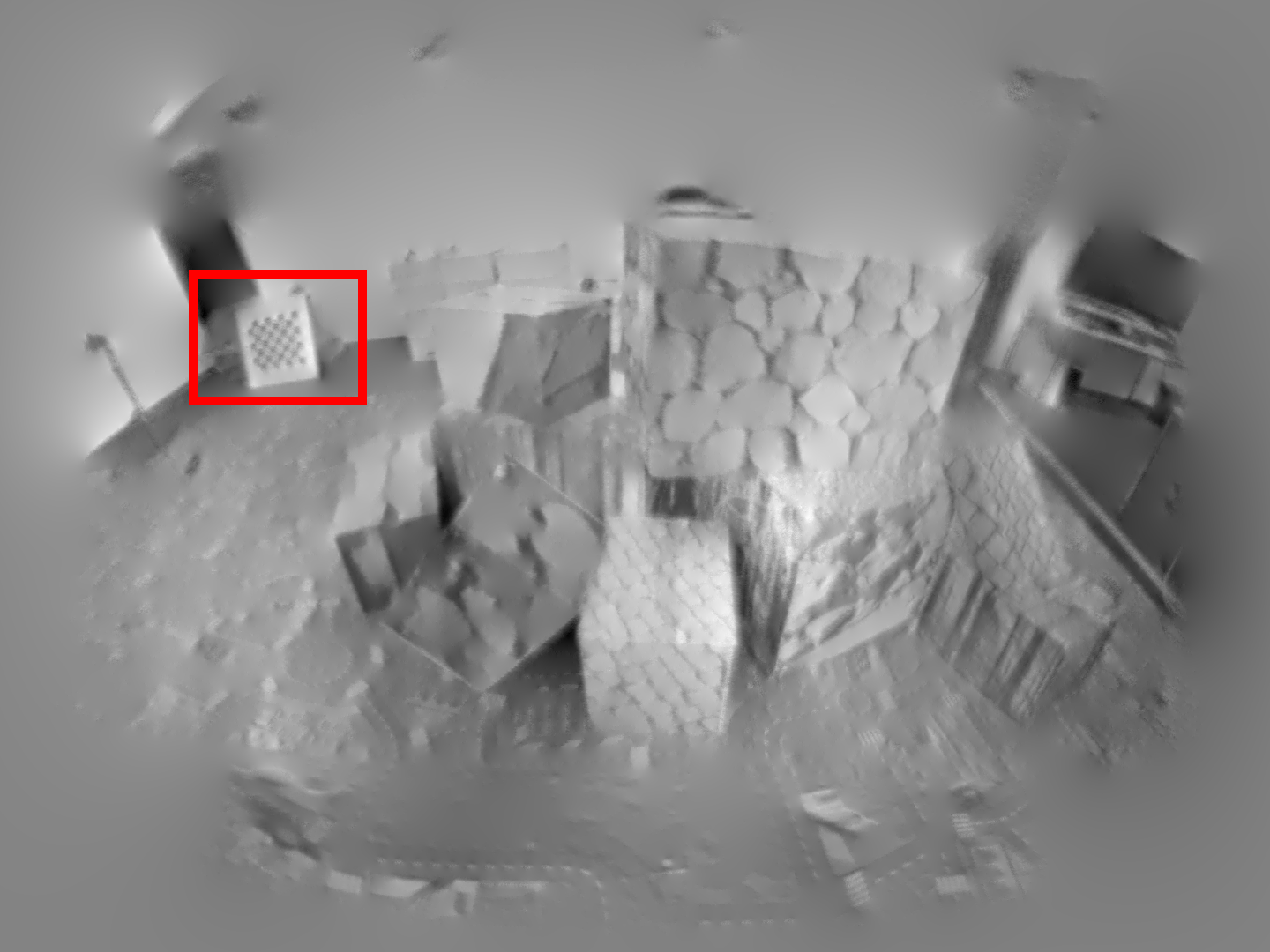}
        &\includegraphics[width=\linewidth]{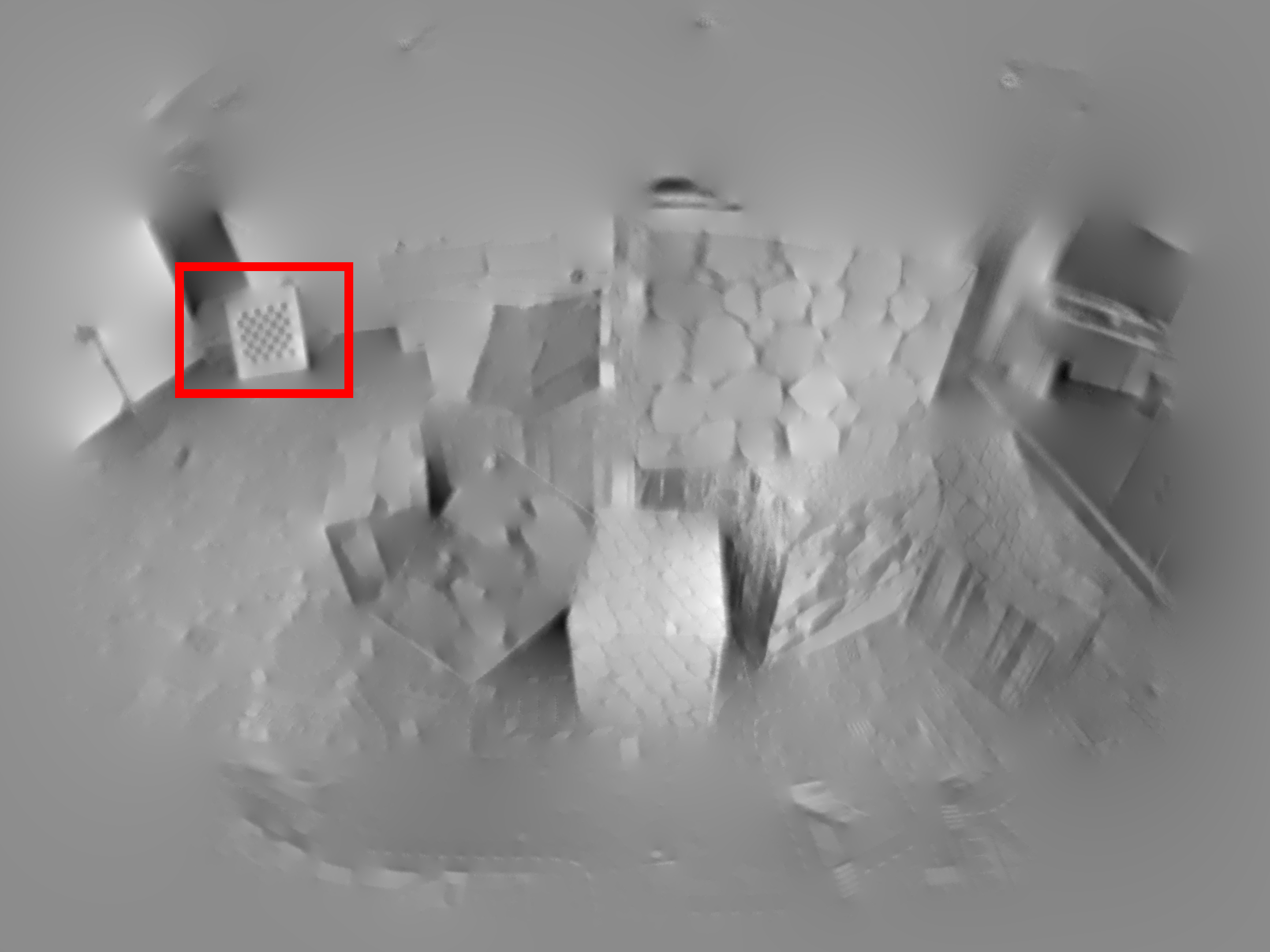}
		\\
		
		\rotatebox{90}{\makecell{zoomed in}}
		&\includegraphics[width=\linewidth]{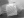}
		&\includegraphics[width=\linewidth]{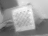}
		&\includegraphics[width=\linewidth]{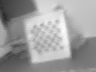}
        &\includegraphics[width=\linewidth]{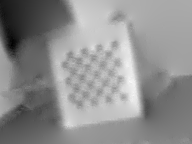}
        &\includegraphics[width=\linewidth]{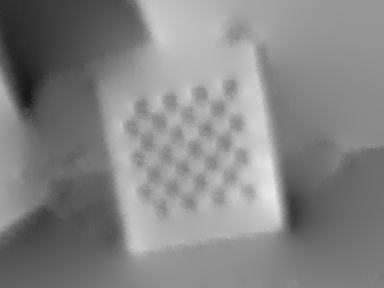}
		\\

        & (a) $512 \times 256$ px
        & (b) $1024 \times 512$ px
        & (c) $2048 \times 1024$ px
        & (d) $4096 \times 2048$ px
        & (e) $8192 \times 4096$ px \\
	\end{tabular}
	}
    \caption{\emph{Grayscale map reconstruction at super-resolution}.
    Results of running \cmaxslam{} at $1024 \times 512$ pixel resolution to estimate the camera trajectory and feed it to the mapping module of \smt{} to obtain panoramic grayscale maps at different resolutions (columns).
    The location of the zoomed-in region is indicated with red rectangles.
    Same events as \cref{fig:iwe_sharpness}.
    }
    \label{fig:super_resolution:grayscale_map}
\end{figure*}
\begin{figure}
     \centering
     \begin{subfigure}[b]{0.48\linewidth}
         \centering
         \includegraphics[width=\linewidth]{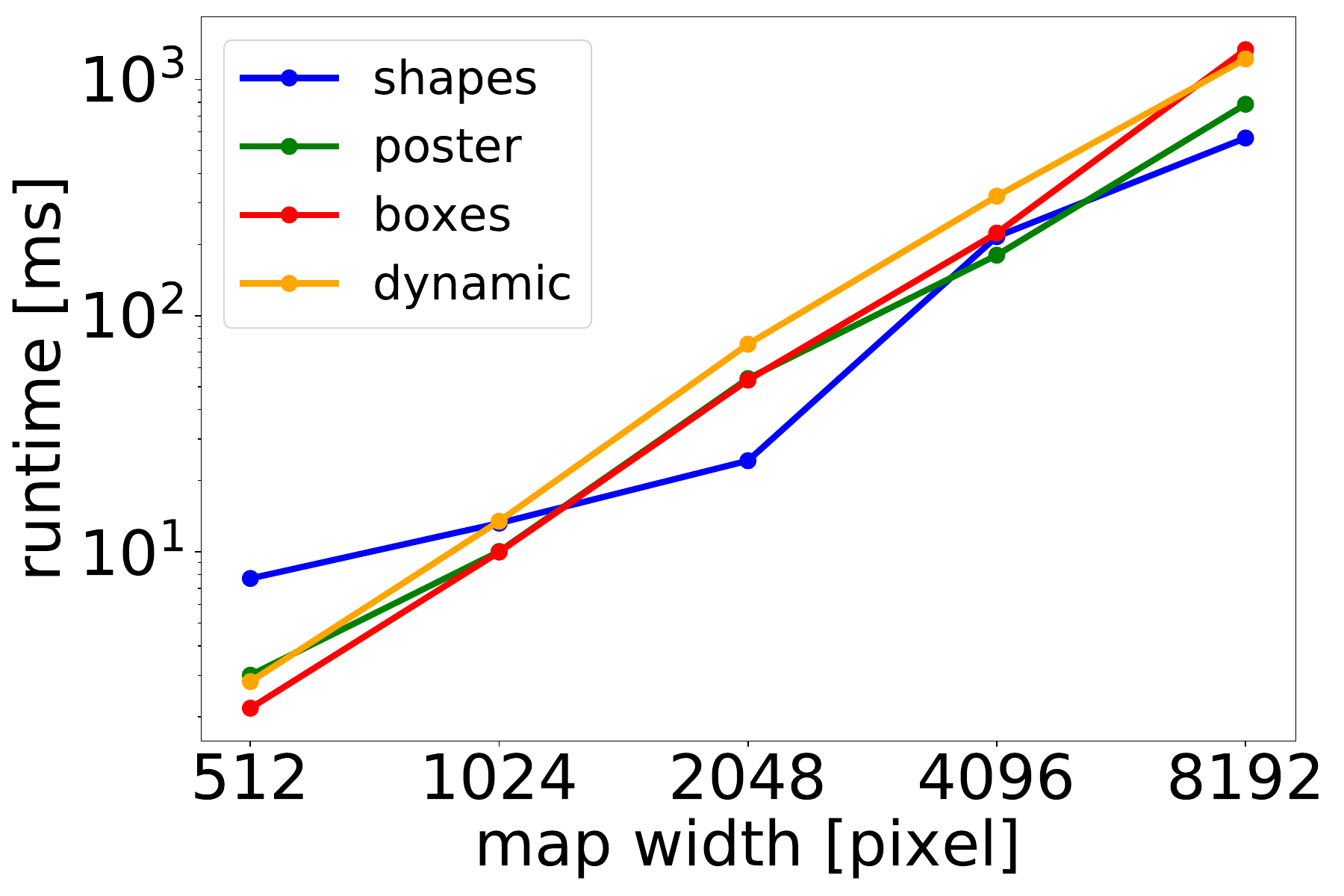}
         \caption{Runtime}
         \label{fig:super_resolution:higher_mosaic:runtime}
     \end{subfigure}
     \begin{subfigure}[b]{0.48\linewidth}
         \centering
         \includegraphics[width=\linewidth]{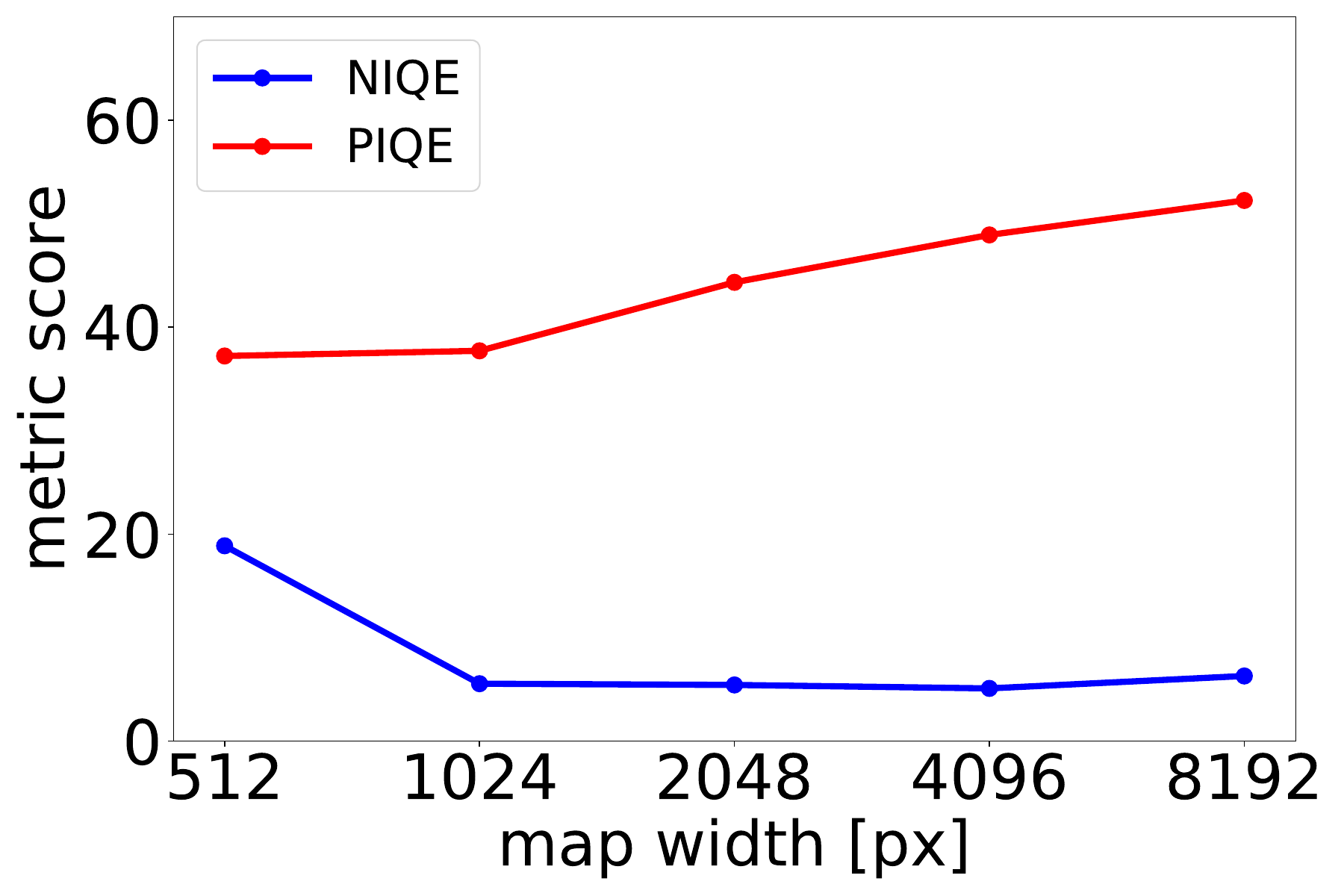}
         \caption{Map quality}
         \label{fig:super_resolution:higher_mosaic:map_quality}
     \end{subfigure}     
        \caption{\emph{Results of grayscale map reconstruction at super resolution}
        (\cref{fig:super_resolution:grayscale_map}).
        (a) Runtime evaluation of Poisson reconstruction at different resolutions.
        (b) Map quality evaluation of grayscale map reconstruction at different resolutions.        
        }
        \label{fig:super_resolution:higher_mosaic}
\end{figure}

In another experiment, we run \cmaxslam{} with a map size of $1024 \times 512$ px, and utilize the estimated trajectory and the events as input to the mapping module of \smt{}.
\Cref{fig:super_resolution:grayscale_map} compares the resulting grayscale panoramas, with resolutions ranging from $512 \times 256$ px to $8192 \times 4096$ px.
Hence, the continuous trajectory and the accurate timing of the events allows us to reach super-resolution.
As the resolution increases, more details of the scene are recovered (e.g., the checkerboard in the second row of \cref{fig:super_resolution:grayscale_map}).
Moving from each column to the next one in \cref{fig:super_resolution:grayscale_map}, 
the number of pixels $\numPixels$ quadruples, 
bilinear voting becomes slower (due to memory access), 
and the cost of Poisson reconstruction via the FFT (whose complexity is $O(\numPixels\log\numPixels)$) also approximately quadruples.
Every time the map size doubles, the runtime of the EKFs that are used in the mapping module of \smt{} approximately increases linearly, 
but that of the Poisson reconstruction increases considerably faster (see \cref{fig:super_resolution:higher_mosaic:runtime}).
We also evaluate the quality of the central parts of the reconstructed grayscale panoramas (first row of \cref{fig:super_resolution:grayscale_map}) using NIQE and PIQE metrics. 
As shown in \cref{fig:super_resolution:higher_mosaic:map_quality}, the panorama with the resolution of $1024 \times 512$ px has the highest quality, closely followed by that of $2048 \times 1024$ px.

\subsection{Sensitivity Analysis}
\label{sec:experiments:sensitive_analysis}

\begin{figure*}
     \centering
     \begin{subfigure}[b]{0.6\linewidth}
         \centering
         \includegraphics[width=\linewidth]{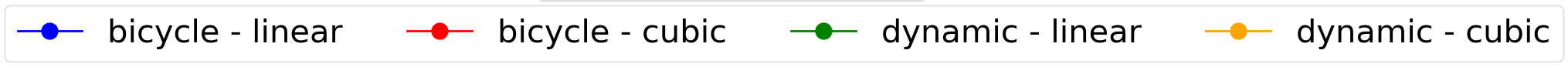}
     \end{subfigure}

    \begin{subfigure}[b]{0.195\linewidth}
         \centering
         \includegraphics[width=\linewidth]{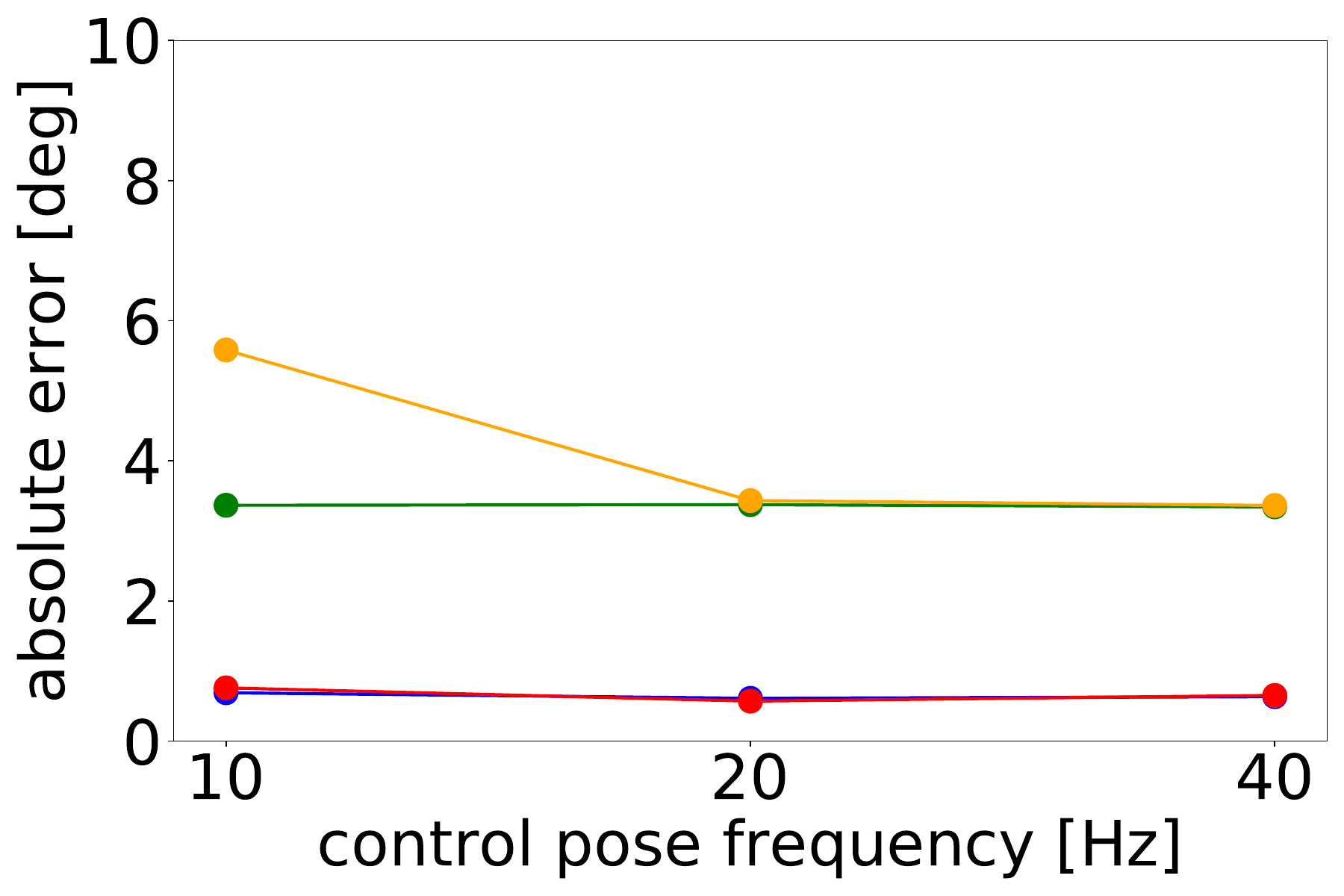}
         \caption{\vspace{-1ex}}
         \label{fig:ablation_study:ctrl_pose_frequency}
     \end{subfigure}
     \begin{subfigure}[b]{0.195\linewidth}
         \centering
         \includegraphics[width=\linewidth]{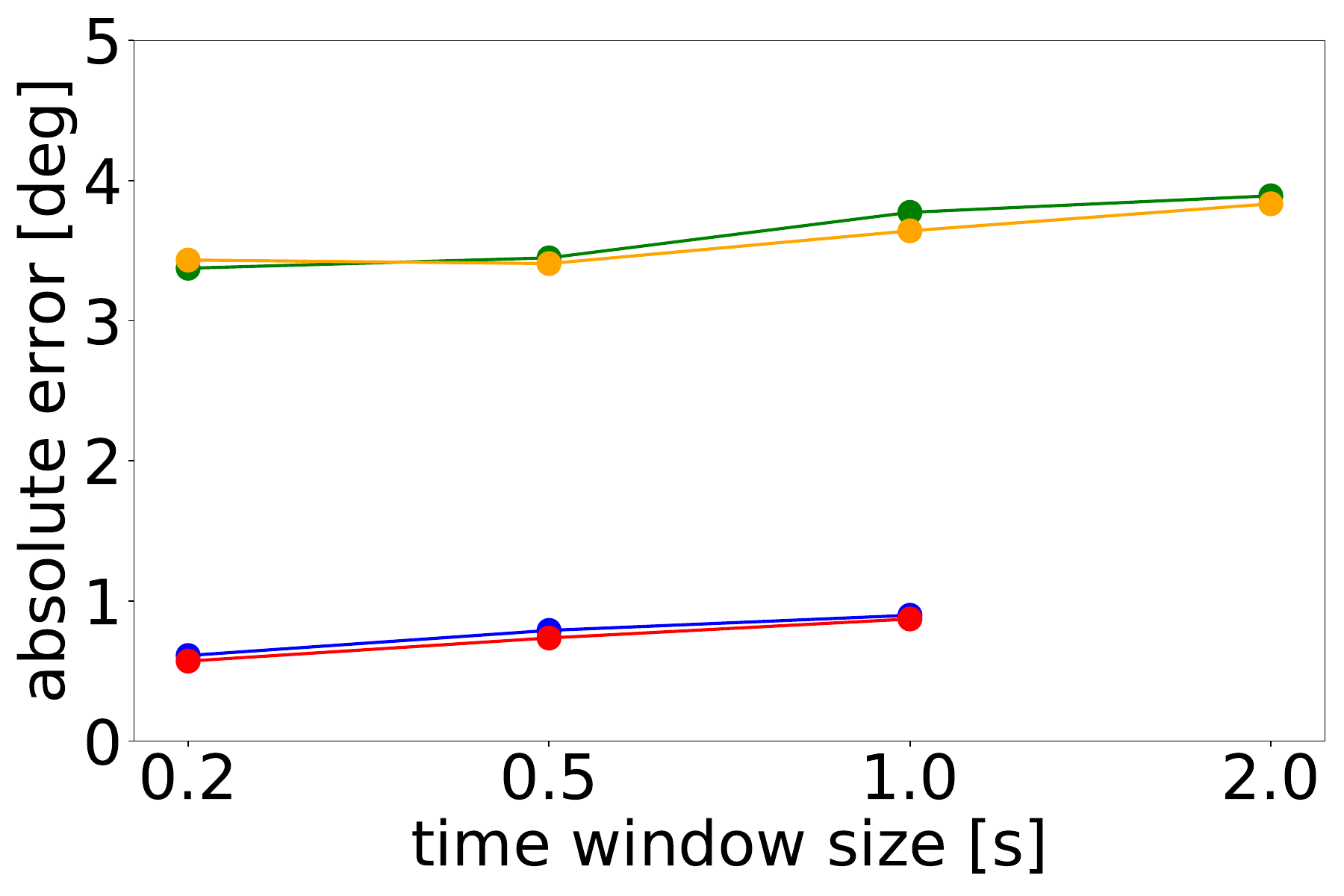}
         \caption{\vspace{-1ex}}
         \label{fig:ablation_study:time_window_size}
     \end{subfigure}
     \begin{subfigure}[b]{0.195\linewidth}
         \centering
         \includegraphics[width=\linewidth]{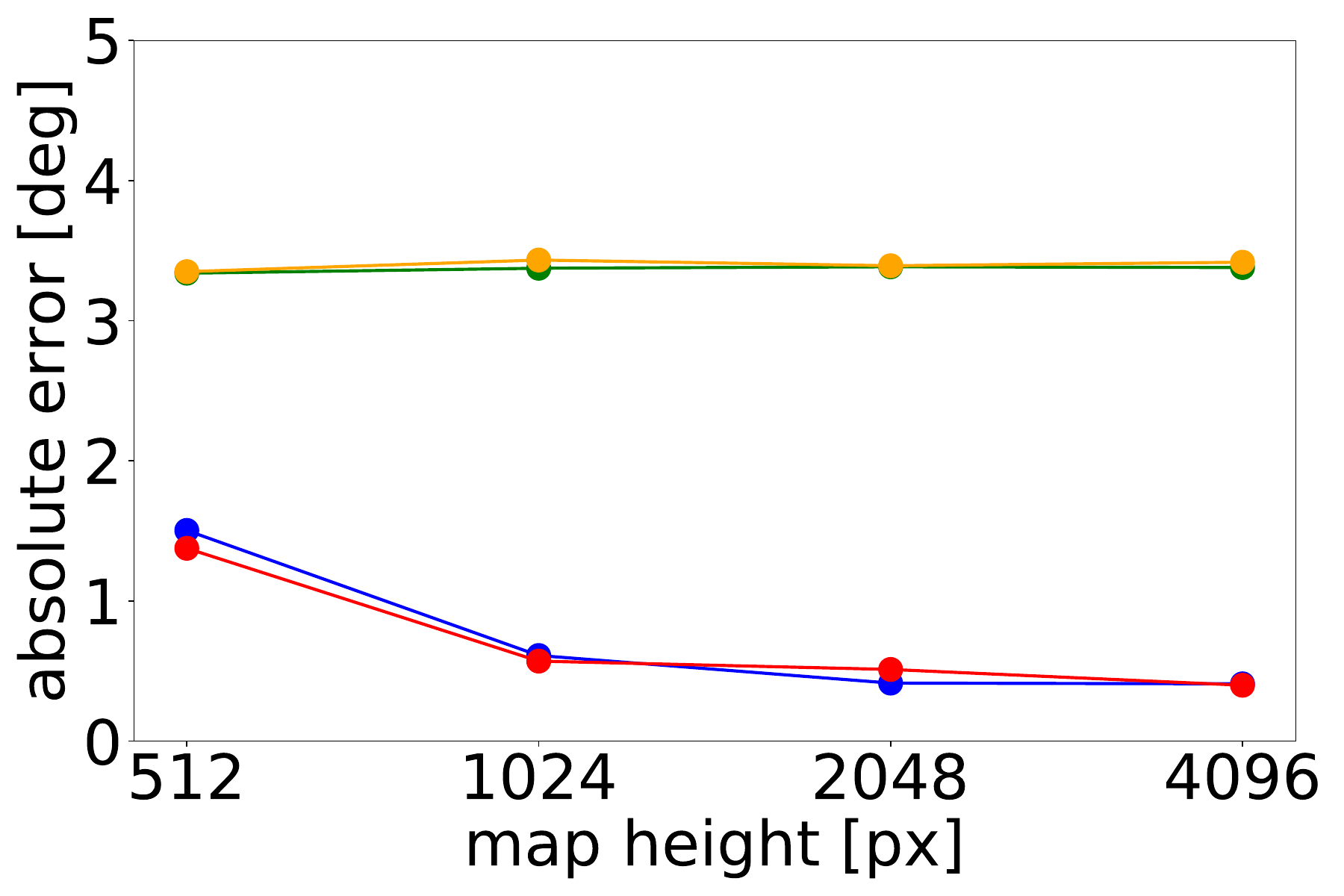}
         \caption{\vspace{-1ex}}
         \label{fig:ablation_study:pano_map_size}
     \end{subfigure}
     \begin{subfigure}[b]{0.195\linewidth}
         \centering
         \includegraphics[width=\linewidth]{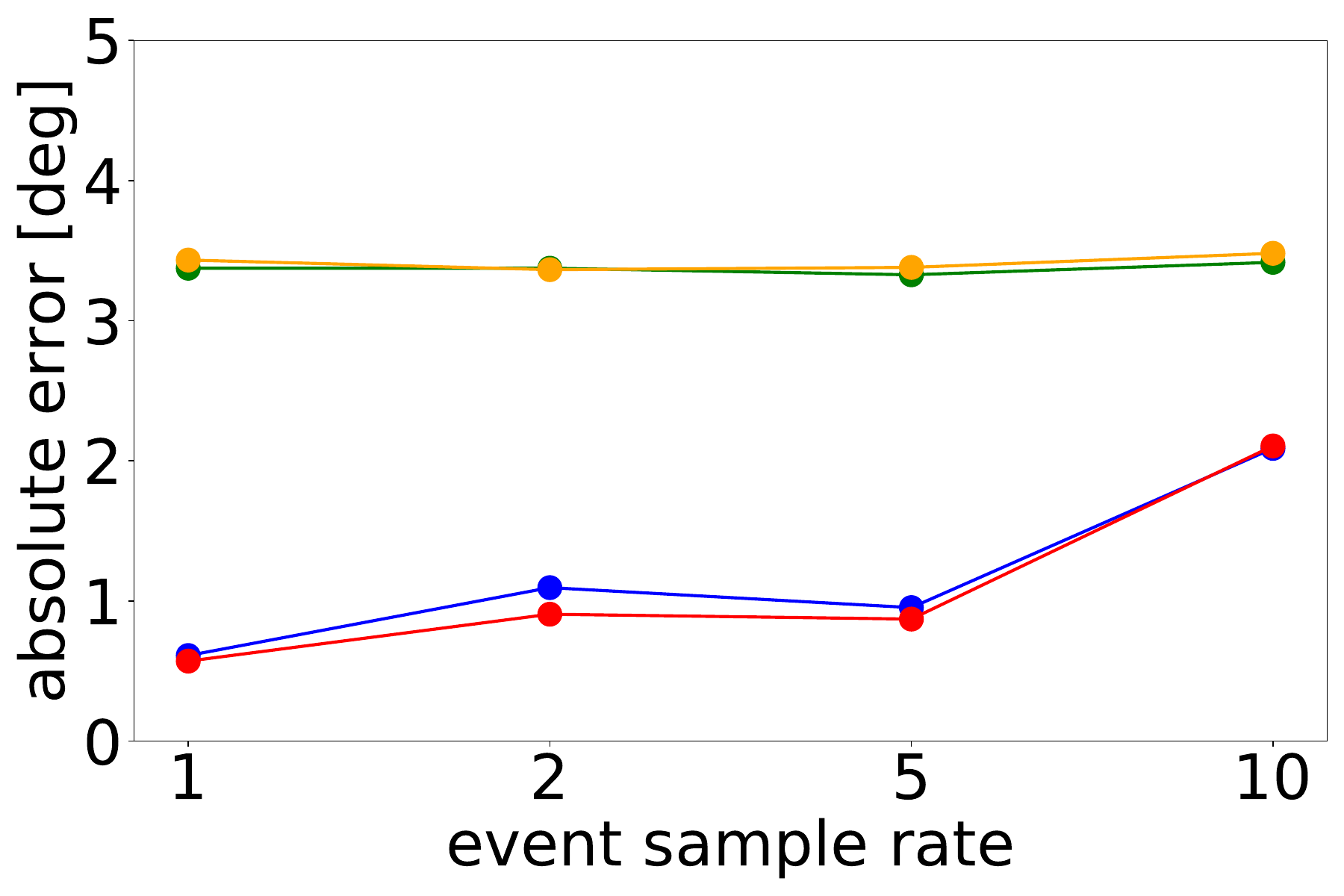}
         \caption{\vspace{-1ex}}
         \label{fig:ablation_study:event_sample_rate_error}
    \end{subfigure}
    \begin{subfigure}[b]{0.195\linewidth}
         \centering
         \includegraphics[width=\linewidth]{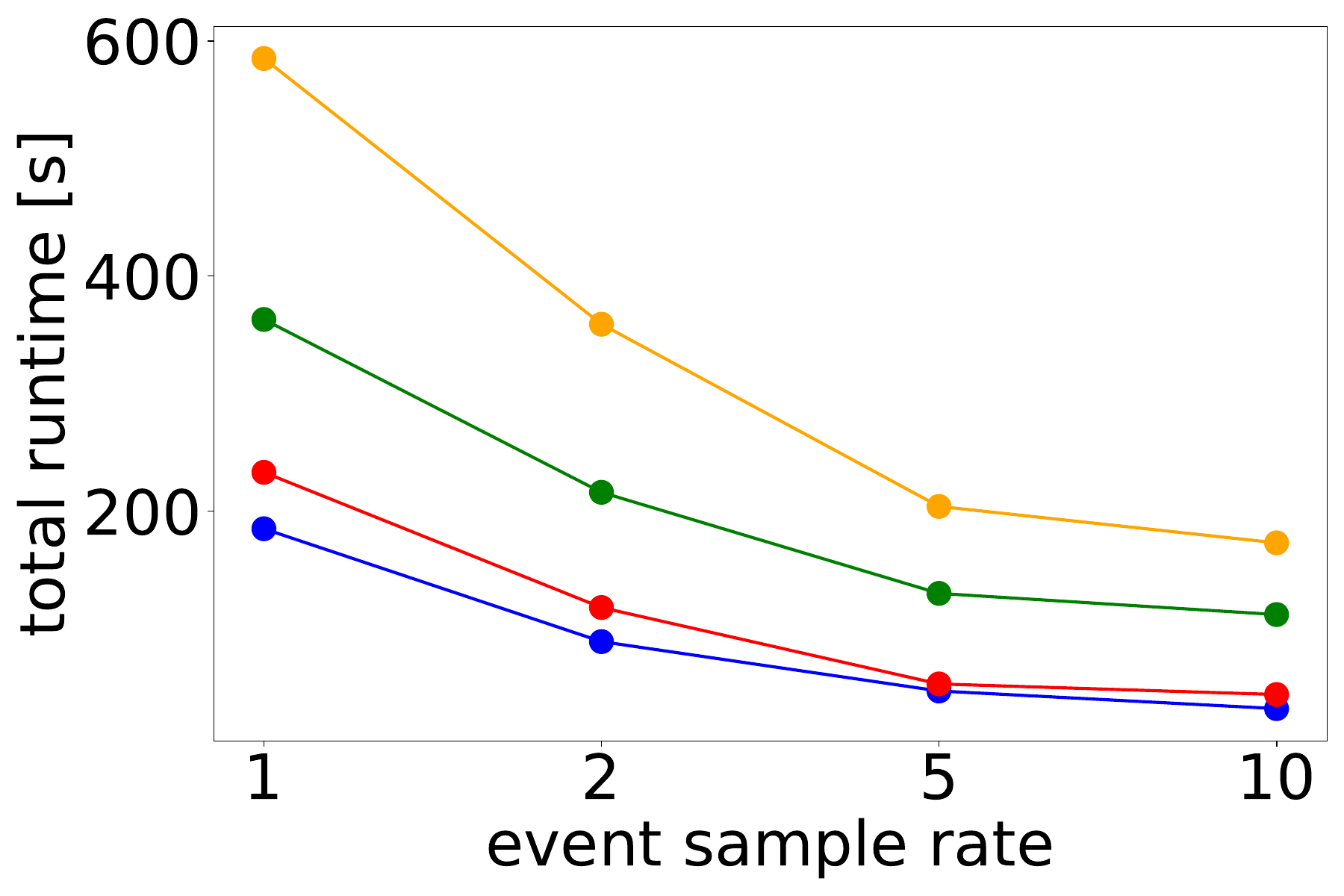}
         \caption{\vspace{-1ex}}
         \label{fig:ablation_study:event_sample_rate_runtime}
    \end{subfigure}

    \caption{\emph{Sensitivity Analysis.}
        From left to right, effect of varying: 
        the frequency of the control poses, 
        the size of the time window, 
        the resolution of the IWE map, 
        and the event sampling rate (on absolute rotation errors and runtime).
        \label{fig:ablation_study}}
\end{figure*}

Finally, we evaluate the sensitivity of our method with respect to several parameters (\cref{fig:ablation_study}).

\subsubsection{Control Pose Frequency}
\label{sec:experiments:sensitive_analysis:ctrl_pose_frequency}
In this set of experiments, we set the time window size to 0.2~s and the resolution of panoramic IWE to $1024 \times 512$ px.
The results are presented in \cref{fig:ablation_study:ctrl_pose_frequency}.
For \bicycle{} (synthetic data), the accuracy of \cmaxslam{} almost does not change as the control pose frequency varies from \mbox{10 Hz} to \mbox{40 Hz}.
For \dynamic{} (real-world data), the error increases obviously when the control pose frequency decreases from \mbox{20 Hz} to \mbox{10 Hz}.
It seems that the control pose frequency is not a major factor to affect the accuracy, once it reaches some value (e.g., \mbox{20 Hz} for \dynamic{}).

\subsubsection{Time (Sliding-)Window Size}
\label{sec:experiments:sensitive_analysis:time_window_size}
In this set of experiments, we set the control pose frequency to \mbox{20 Hz} and the resolution of panoramic IWE to $1024 \times 512$ px.
The stride we use to slide the time window is half of its size.
Overall, the accuracy of \cmaxslam{} slightly decreases as the time window size increases on both \bicycle{} and \dynamic{} (\cref{fig:ablation_study:time_window_size}). 
The window size of 2 s is not tested on \bicycle{} since it is just 5 s long.

\subsubsection{Panoramic IWE Resolution}
\label{sec:experiments:sensitive_analysis:pano_map_size}
In this set of experiments, we set the control pose frequency to \mbox{20 Hz} and the time window size to 0.2~s.
As \cref{fig:ablation_study:pano_map_size} shows, the errors decrease as the map size grows.
However, as described in \cref{sec:experiments:super_resolution}, there should be lower and upper limits of the map size for \cmaxslam{} to work.
Moreover, the computational cost increases rapidly as the map size grows (see \cref{fig:super_resolution:higher_cmax_slam:runtime}).

\subsubsection{Event Sampling Rate}
\label{sec:experiments:sensitive_analysis:event_sample_rate}
In this set of experiments, we set the control pose frequency to \mbox{20 Hz}, the time window size to 0.2~s, and the panoramic map resolution to $1024 \times 512$ px.
The events are systematically downsampled before being fed to the front-end and back-end.
More specifically, an event sampling rate of 5 means both front-end and back-end process one event out of five (20\% of events).
As illustrated in \cref{fig:ablation_study:event_sample_rate_error}, \dynamic{} is not sensitive to the event sampling rate until it grows to 10.
In contrast, the accuracy on \bicycle{} declines gradually as the event sampling rate increases.

In addition, \cref{fig:ablation_study:event_sample_rate_runtime} depicts how the event sampling rate affects the processing time of \cmaxslam{}: 
as expected, the runtime decreases as fewer events are processed, 
but in a non-linear way (processing half of the events does not reduce the runtime by half).
In all cases, the cubic B-spline is always more expensive than the linear one.
With \cref{fig:ablation_study:event_sample_rate_error,fig:ablation_study:event_sample_rate_runtime}, users can set the trade-off for their own applications.

\section{Space Application}
\label{sec:space_application}

The low latency, HDR and low power consumption characteristics of event cameras make them attractive to space applications \cite{Cohen19jas,Afshar20jsen,McHarg22spie}.
For example, \cite{Chin19cvprw,Chin20wacv} apply event cameras to the star tracking problem, which consists of estimating the ego-motion of a rotating camera by tracking stars.
Data may be acquired by an event camera on Earth or on a satellite.

Let us apply the proposed method to address the star tracking problem while simultaneously reconstructing a panoramic star map. 
The experiments show that our method outperforms prior work in terms of accuracy and robustness.

Our method does not convert events into frames nor does it extract correspondences because it is based on CMax \cite{Gallego18cvpr},
which handles data association implicitly (by the warped events that vote on the same pixel). 
Our BA also differs from prior ones used in star tracking, which are feature-based (between 2D-3D point correspondences) after converting events into frames \cite{Chin19cvprw} or fitting line segments and extracting their end points \cite{Chin20wacv}.
Therefore, our method is more adapted to the characteristics of event data than prior work.

\subsection{Experiments}

\subsubsection{Dataset}
\label{sec:space_application:dataset}

We test the proposed method on the star dataset \cite{Chin19cvprw}. 
It has eleven 45~s long sequences recorded with a DAVIS240C event camera ($240\times 180$ px) while observing a star field rotating at a constant angular velocity~of~$4^\circ/$s displayed on a screen. 
Like \cite{Chin19cvprw,Chin20wacv}, experiments are conducted on Seqs.~1 -- 6, which contain 1.4 -- 6.2M events.
Since the format in which the data is provided (in txt files and floating-point undistorted events, using a combined homography-calibration matrix) is not directly compatible with our implementation that uses raw events and known camera calibration matrix, we run the front-end and BA offline.

\subsubsection{Metrics}
\label{sec:space_application:metrics}

Previous works~\cite{Chin19cvprw,Chin20wacv} perform rotation averaging \cite{Hartley13ijcv} by feeding an additional set of absolute poses from a star identification system, which act as anchors for trajectory refinement. 
Such data is not available in the dataset, hence we can only compare the performance in terms of relative rotation estimation.
For a direct comparison, we adopt the benchmark in \cite{Chin19cvprw}: 
measuring the angular distance between relative rotations in a time interval of 400 ms.
We also use the metrics in \cite{Chin19cvprw}: RMSE and standard deviation.

\subsubsection{Results}
\label{sec:space_application:results}

\begin{table*}
\centering
\caption{\emph{Space Application}. 
Errors of relative rotations (RMSE and standard deviation $\sigma$) [$^\circ$].
\label{tab:star_rmse}
}
\adjustbox{max width=\linewidth}{%
\setlength{\tabcolsep}{3pt}
\begin{tabular}{ll*{12}{S[table-format=2.3]}}
\toprule
Sequence & \multicolumn{2}{c}{\text{1}}
         & \multicolumn{2}{c}{\text{2}}
         & \multicolumn{2}{c}{\text{3}}
         & \multicolumn{2}{c}{\text{4}}
         & \multicolumn{2}{c}{\text{5}}
         & \multicolumn{2}{c}{\text{6}}\\
\cmidrule(l{1mm}r{1mm}){2-3}
\cmidrule(l{1mm}r{1mm}){4-5}
\cmidrule(l{1mm}r{1mm}){6-7}
\cmidrule(l{1mm}r{1mm}){8-9}
\cmidrule(l{1mm}r{1mm}){10-11}
\cmidrule(l{1mm}r{1mm}){12-13}
& \text{RMSE} & $\sigma$
& \text{RMSE} & $\sigma$
& \text{RMSE} & $\sigma$
& \text{RMSE} & $\sigma$
& \text{RMSE} & $\sigma$
& \text{RMSE} & $\sigma$\\
\midrule
Chin \emph{et al}. \cite{Chin19cvprw} & 0.951 & 7.098 & 0.731 & 8.880 & 12.461 & 14.467 & 13.435 & 19.612 & 18.248 & 8.141 & 12.799& 17.882 \\[.8ex]
Bagchi \emph{et al}. \cite{Chin20wacv} & 0.850 & \novalue & 0.900 & \novalue & 0.700 & \novalue & 0.970 & \novalue & 0.200 & \novalue & 0.900 & \novalue \\[.8ex]
Ours & 0.487& 0.207 & 0.503& 0.213 & 18.544& 18.161 & 0.968& 0.413 & 0.905& 0.386 & 0.966& 0.413 \\
\bottomrule
\end{tabular}
}
\end{table*}

\begin{figure*}
     \centering
     \begin{subfigure}[b]{0.285\linewidth}
         \centering
         \includegraphics[width=\linewidth]{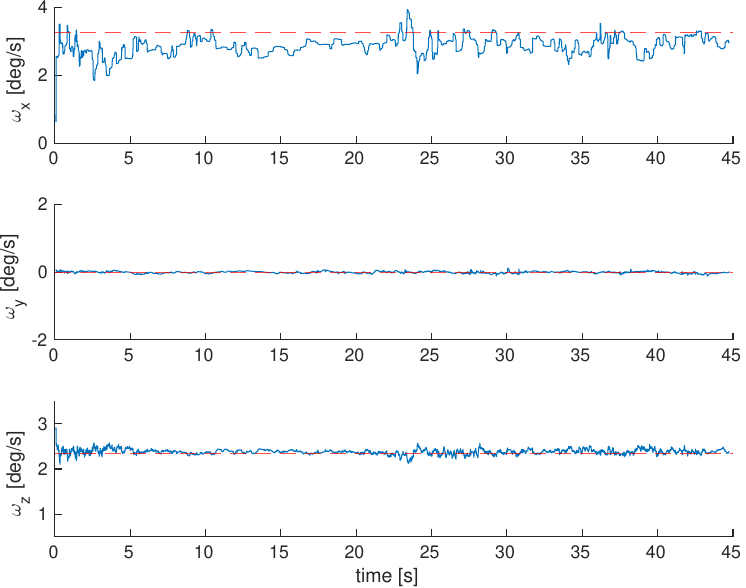}
         \caption{Angular velocity}
         \label{fig:star_tracking:angvel}
     \end{subfigure}
     \hfill
     \begin{subfigure}[b]{0.285\linewidth}
         \centering
         \includegraphics[width=\linewidth]{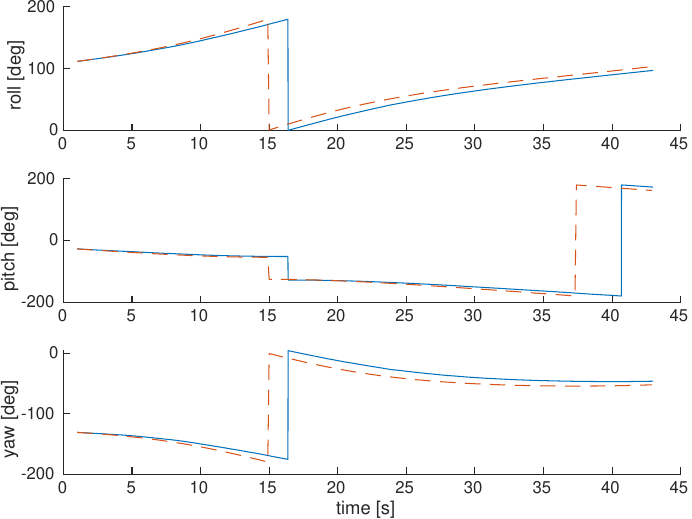}
         \caption{Euler angles}
         \label{fig:star_tracking:poses}
     \end{subfigure}
     \hfill
     \begin{subfigure}[b]{0.148\linewidth}
         \centering
         \includegraphics[width=\linewidth]{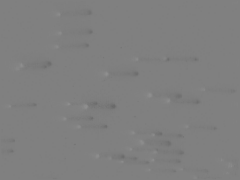}\\[.5ex]

         \includegraphics[width=\linewidth]{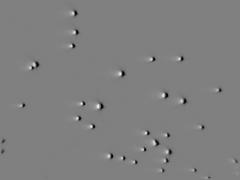}
         \caption{Local IWE}
         \label{fig:star_tracking:local_iwe}
     \end{subfigure}
     \hfill
     \begin{subfigure}[b]{0.18\linewidth}
         \centering
         \gframe{\includegraphics[width=\linewidth]{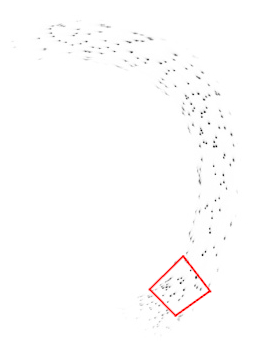}}
         \caption{Global IWE}
         \label{fig:star_tracking:starmap}
     \end{subfigure}
        \caption{\emph{Space Application}.
        Results on sequence 5 of \cite{Chin19cvprw}: 
        (a) estimated angular velocity. 
        (b) Absolute rotation.
        (c) Local IWE before/after CMax (generated by the front-end with the event polarity used), corresponding to the region marked by the red rectangle in (d).
        (d) A zoomed-in part of motion-compensated event-based star map (generated by the back-end with the event polarity not used).
        In (a)--(b) blue solid lines are estimated while red dashed lines are the groundtruth.
        }
        \label{fig:star_tracking}
\end{figure*}

\Cref{tab:star_rmse} compares our approach with the state of the art. 
Our method produces accurate results ($ < 1^\circ$ RMSE): better than the state of the art in Seqs.~1, 2, 4, and comparable in the rest.
Seq.~3 is known to be problematic \cite{Chin20wacv}.
Further, our method is the most consistent, as reported by its smaller standard deviation compared to other methods. 

\Cref{fig:star_tracking} shows the output of our method for one of the sequences.
Since the sequences have no loops, rotation drift is inevitable in \cref{fig:star_tracking:poses} without additional inputs (anchor absolute poses).
Nevertheless, the estimation is good: angular velocity errors are small (\cref{fig:star_tracking:angvel}), and the by-product map is sharp (\cref{fig:star_tracking:local_iwe,fig:star_tracking:starmap}), indicating that the method has successfully estimated the motion that caused the event data.

\section{Limitations}
Event cameras require the presence of contrast in the scene. 
It is difficult to track in textureless regions or with few edges. 
In the absence of data, one could act on the camera parameters, decreasing the value of the contrast threshold $C$ to generate more events, and hopefully be able to sense small edges.

All methods compared assume brightness constancy. 
Events caused by flickering lights or hot pixels do not follow this assumption, and hence could be problematic if they dominated over events produced by moving edges. 
In practice, our method exhibits some robustness to the few flickering lights in the scene, such as residual light from mo-cap systems.

While our system comprises a back-end for refinement, it lacks an explicit module with loop closure capabilities (all tested methods lack it), and therefore inevitably accumulates drift. 
This is most noticeable in sequences where the scene is revisited after a full 360$^\circ$ rotation around the vertical axis. 
To the best of our knowledge, none of the methods on event-based rotation-only ego-motion estimation detect loop closure 
(we even dare to extend this to the event-based 6-DOF ego-motion estimation literature, too). 
This topic is still in its infancy in event-based vision, and hence is left as future work.

Event cameras with VGA and higher spatial resolution may produce a colossal amount of events (e.g., 1Gev/s \cite{Finateu20isscc}), which requires a considerable amount of memory and computing power if all events are to be used. 
More proficient processors, ideally massively parallel like neuromorphic processors, could take on this problem.

\section{Conclusion}
\label{sec:conclusion}
We have presented the first event-based rotation-only bundle adjustment and the first SLAM system comprising both a front-end and a back-end.
Both are principled, based on event alignment (CMax), and the back-end has a continuous-time camera trajectory model.
Two trajectory models have been exemplified (linear and cubic B-splines), with similar accuracy but different complexity (cubic is more demanding). 
Other continuous-time models, such as Gaussian processes, may be used.
To the best of our knowledge, no prior work on the same task has considered past events for trajectory refinement without converting them into frames.

While the problem of CMax is not formulated as an NLLS one (hence powerful second-order algorithms such as Gauss-Newton do not apply), we use first-order nonlinear conjugate gradient efficiently. 
We compute derivatives analytically and search only in the space of trajectories, since the map is naturally determined by the trajectory, the events and the projection model.
This enables potential real-time operation, depending on the event rate and processing capabilities of the platform.

This work provides the most comprehensive benchmark of rotation-only estimation methods with event cameras, evaluating prior works and re-implementing, to the best of our understanding, previous methods whose code is unavailable. 
The experiments demonstrate that our proposal is accurate:
On synthetic data, our method outperforms all baselines.
On real-world data, we propose a more sensible figure of merit for the evaluation of event-based rotation estimation; and the results show that
the front-end gives results on par with the state of the art;
the back-end is closely integrated with the front-end and is able to refine trajectories, as demonstrated by small proxy reprojection errors and sharp maps.

Finally, due to the versatility of our method (the experiments show that it works indoors, outdoors, in natural scenes, and in space data), it could be used on satellites as well as on rovers. 
It could also be used for sky mapping (with wide FOV lenses, since for narrow FOV lenses a simpler linear approximation suffices for event warping).
Hence, our work has a direct impact on event-based HDR sky mapping, situational awareness (SDA) and space robotics.
We hope our work will spark ideas to advance the field of ego-motion estimation (not only in \mbox{3-DOF} but also in higher DOFs).

\section*{Acknowledgment}
The authors would like to thank Mr. Yunfan Yang and Ms. Nan Cai for assistance in recording the data sequences.
\ifarxiv
We thank early efforts from Ms.~Boshu Zhang and Mr.~Haofei~Lu.
\else
\vspace{-.5ex}
\fi 

\ifarxiv
\appendices

\ifarxiv
\else
\section*{SUPPLEMENTARY MATERIAL}
\fi

\section{Proof of Well-posedness (No Event Collapse)}
\label{app:nocollapse}
\def\x{\mathbf{x}}
\def\X{\mathbf{X}}
\def\ba{\mathbf{a}}
\def\bp{\mathbf{p}}
\def\be{\mathbf{e}}

The warp $\Warp$ from the image plane to the panoramic IWE maps a sensor pixel to its location on the panorama. 
One can show that this warp is well-defined, i.e., it does not over-concentrate point trajectories (creating ``sinks''), 
and therefore, no event collapse \cite{Shiba22sensors,Shiba22aisy} occurs. 
This can be shown by computing the determinant of the Jacobian of the warp (which indicates how the area element transforms from the image plane to the panorama) and reasoning that it cannot vanish.

The warp $\x_{k}\stackrel{\Warp}{\mapsto}\bp_{k}$ in \cref{sec:panoramicIWE}
can be described as the composition of three transformations: 
back-projection, 3D rotation and (equi-rectangular) projection:
\begin{equation}
\x_{k}\mapsto\X_{k}\mapsto\X'_{k}\mapsto\bp_{k}.
\end{equation}
In calibrated camera coordinates, $\x_{k}=(x_{k},y_{k})^{\top}$ is
back-projected onto a 3D point at the unit depth plane, $\X_{k}=(x_{k},y_{k},1)^{\top}$,
then it is 3D-rotated into $\X'_{k}=(X',Y',Z')^{\top}=\Rot\X_{k}$
and projected onto the panorama point $\bp_{k}$. 
Applying the chain rule, the Jacobian is given by 
\begin{equation}
\underbrace{\frac{d\bp_{k}}{d\x_{k}}}_{2\times2}
=\underbrace{\frac{d\bp_{k}}{d\X'_{k}}}_{2\times3}
\,\underbrace{\frac{d\X'_{k}}{d\X{}_{k}}}_{3\times3}
\,\underbrace{\frac{d\X{}_{k}}{d\x_{k}}}_{3\times2}.
\label{eq:JacChainRule}
\end{equation}

The derivative of the sensor-pixel back-projection is $\frac{d\X{}_{k}}{d\x_{k}}=(\be_{1},\be_{2})$,
where $\be_{1}=(1,0,0)^{\top}$ and $\be_{2}=(0,1,0)^{\top}$. The
derivative of the 3D rotation is $\frac{d\X'_{k}}{d\X{}_{k}}=\Rot$
(the rotation matrix), and the derivative of the panoramic projection \eqref{eq:equirect_projection} is, 
omitting the translation and scaling from $[-\pi,\pi]\times[-\pi/2,\pi/2]$ to $[0,w]\times[0,h]$, 
equal to
\begin{equation}
\frac{d\bp_{k}}{d\X'_{k}}=\left(\begin{array}{c}
\ba_{1}^{\top}\\
\ba_{2}^{\top}
\end{array}\right),
\end{equation}
with 
\begin{align}
\ba_{1}^{\top} & =\frac{1}{s_{1}Z'^{2}}\left(Z',0,-X'\right)\\
\ba_{2}^{\top} & =\frac{1}{s_{2}r'^{3}}\left(-X'Y',X'^{2}+Z'^{2},-Z'Y'\right)\\
s_{1} & =\sqrt{1+(X'/Z')^{2}}\\
s_{2} & =\sqrt{1-(Y'/r')^{2}}\\
r' & =\|\X_{k}'\|=\sqrt{X'^{2}+Y'^{2}+Z'^{2}}.
\end{align}
Therefore, the Jacobian (\ref{eq:JacChainRule}) becomes
\begin{equation}
\frac{d\bp_{k}}{d\x_{k}}=\left(\begin{array}{c}
\ba_{1}^{\top}\\
\ba_{2}^{\top}
\end{array}\right)\Rot(\be_{1},\be_{2})=\begin{pmatrix}\ba_{1}^{\top}\Rot\be_{1} & \ba_{1}^{\top}\Rot\be_{2}\\
\ba_{2}^{\top}\Rot\be_{1} & \ba_{2}^{\top}\Rot\be_{2}
\end{pmatrix},
\end{equation}
whose determinant is 
\begin{equation}
\det\left(\frac{d\bp_{k}}{d\x_{k}}\right)=(\ba_{1}^{\top}\Rot\be_{1})(\ba_{2}^{\top}\Rot\be_{2})-(\ba_{2}^{\top}\Rot\be_{1})(\ba_{1}^{\top}\Rot\be_{2}).
\end{equation}
After some algebraic manipulations (see \cite[App.~D]{Gallego11thesis}),
\begin{equation}
\begin{split}\det\left(\frac{d\bp_{k}}{d\x_{k}}\right) & =(\Rot\be_{1})^{\top}\ba_{1}\ba_{2}^{\top}\Rot\be_{2}-(\Rot\be_{1})^{\top}\ba_{2}\ba_{1}^{\top}\Rot\be_{2}\\
 & =-(\Rot\be_{1})^{\top}(\ba_{2}\ba_{1}^{\top}-\ba_{1}\ba_{2}^{\top})\Rot\be_{2}\\
 & =-(\Rot\be_{1})^{\top}\left((\ba_{1}\times\ba_{2})\times(\Rot\be_{2})\right)\\
 & =-\det(\Rot\be_{1},\ba_{1}\times\ba_{2},\Rot\be_{2})\\
 & =\det(\Rot\be_{1},\Rot\be_{2},\ba_{1}\times\ba_{2}).
\end{split}
\end{equation}
Computing the cross product 
\begin{equation}
\ba_{1}\times\ba_{2}=\frac{s_{1}}{s_{2}r'^{3}}\left(X',Y',Z'\right)^{\top}=\frac{s_{1}}{s_{2}r'^{3}}\X'_k=\frac{s_{1}}{s_{2}r'^{3}}\Rot\X_k
\end{equation}
and substituting in the Jacobian yields 
\begin{equation}
\label{eq:JacobianDetSimplified}
\begin{split}\det\left(\frac{d\bp_{k}}{d\x_{k}}\right) & =\frac{s_{1}}{s_{2}r'^{3}}\det(\Rot\be_{1},\Rot\be_{2},\Rot\X_k)\\
 & =\frac{s_{1}}{s_{2}r'^{3}}\det(\Rot)\det(\be_{1},\be_{2},\X_k)\\
 & =\frac{s_{1}}{s_{2}r'^{3}}.
\end{split}
\end{equation}

By definition, we have the bounds: $s_{1}\geq1$, $0\leq s_{2}\leq1$
and $1 \leq r' < \infty$ because $r'=\|\X_{k}'\|=\|\Rot\X_{k}\|=\|\X_{k}\|=(1+x_k^2+y_k^2)^{\frac{1}{2}}$. 
The numerator of \eqref{eq:JacobianDetSimplified} is not zero and the denominator cannot grow unbounded because we assume the FOV of the camera is limited so that the perspective back-projection from the sensor to the unit depth plane is well defined.
Since the determinant of the Jacobian of the warp \eqref{eq:JacobianDetSimplified} does not vanish, the area element cannot be arbitrarily scaled down to zero, and therefore event collapse cannot happen.

\section{Implementation Details of Front-end Methods}
\label{appendix:baselines}
The code of \smt{} is not publicly available, so we re-implemented it and improved the following aspects:
\begin{itemize}
    \item In addition to the PF tracker described in \cite{Kim14bmvc}, we implement the EKF tracker presented in \cite{Gallego15arxiv,Kim18phd}, which is faster than the PF one. 
    
    \item We use the measurement model $p(e_k | M)$ in terms of the temporal contrast~\cite{Gallego15arxiv} 
    \begin{equation}
        \label{eq:smt:egm:mapper:contrast}
        z = \pol_k C, \qquad h(\bg) = - \bg \cdot \velflow \Delta t_k,
    \end{equation}
    instead of in terms of the instantaneous event rate \eqref{eq:smt:egm:mapper}.
    The choice is justified by empirical evidence: the former is closer to a Gaussian distribution \cite{Lichtsteiner08ssc} than the latter \cite{Kim18phd}, thus fitting better with the EKF assumptions.
    
    \item In contrast to the per-event processing strategy in \cite{Kim14bmvc}, we process events by batch for speed up, 
    as it is done in the back-end (\cref{sec:method:pipeline:backend}).

    \item Initialization is critical, but \cite{Kim14bmvc} lacks an explicit bootstrapping step. 
    To achieve a stable initialization of the system, we use dead reckoning from IMU data when available (e.g., if using a DAVIS camera \cite{Brandli14ssc}).

    \item Map blur reduction upon revisit: 
    If the camera revisits the same part of the scene, tracking errors will blur the map. 
    To reduce error propagation, and therefore map blur, we introduce a panoramic map updating strategy that limits the number of times that each map pixel $\bp_{m}$ is updated.

    \item Map update outlier rejection:
    The mapping EKFs are based on the linearized EGM, which assumes constancy of the spatial gradient between the two panorama pixels corresponding to two consecutive events (at the same sensor pixel). 
    Therefore, if the two map pixels are further apart than a threshold (e.g., $\Delta \mappointtk > 10$ pixels), the current event is considered an outlier and discarded.

    \item Fast Poisson reconstruction: 
    To reduce the computational burden, we perform Poisson reconstruction on the smallest rectangular region with non-zero gradient 
    (with map pixels visited at least once).
\end{itemize}

Let us provide further details about the parameters of the baseline methods.

\begin{itemize}
    \item \smt{}: We bootstrap \smt{} using IMU data, expect for \playroom{}, where GT is used because there is no available IMU data.
    The resolution of the reconstructed grayscale panoramic map is $1024 \times 512$ px. 
    If a region of the map is visited by the camera more than five times, it will stop being updated to reduce error propagation (e.g., map blurring). 
    As mentioned, for the real data, the precise value of the camera's contrast sensitivity is typically inaccessible, thus we set it to $C = 0.22$. 
    While for synthetic data, $C = 0.2$.
    PF-\smt{} runs with 100 particles. 
    \item \rtpt{}: \cite{Reinbacher17iccp} warps events to a map using a cylindrical projection. 
    For better comparison, the projection is replaced with the equirectangular one adopted in \smt{}. 
    \item \cmaxgae{}: The time interval to compute the angular velocity is set to $dt = 25$ ms, which gives a rate of \mbox{40 Hz} for the estimated angular velocities and poses.
    Note that \cmaxgae{} samples events for speed-up, it processes 25\% of events for all sequences.
    \item \cmaxw{}: We run the front-end of the \cmaxslam{} alone (with the event slicing strategy of \cref{sec:method:pipeline:frontend}), which estimates angular velocities at 100 Hz, and integrates them to provide estimated poses for accuracy evaluation.
\end{itemize}

\IEEEtriggeratref{67} 
\bibliographystyle{IEEEtran}

\else
\bibliographystyle{IEEEtran}

\input{biographies.tex}

\cleardoublepage

\fi 

\end{document}